%% file: overhead_overview_generation_detection.tex
\documentclass{SIP}

\usepackage{multirow}
\usepackage{graphicx}
\usepackage{subfig}
\usepackage{amsmath}
\usepackage{amssymb}
\usepackage{amsxtra}
\usepackage{threeparttable}
\usepackage{glossaries}
\usepackage[output-decimal-marker={,}]{siunitx}
\usepackage{comment}
\usepackage{booktabs}
\usepackage[capitalise, noabbrev]{cleveref}

\input{latex_helpers.tex}

\newcommand{\MB}[1]{\textcolor{magenta}{{#1}}}

\begin{document}

\title{An Overview on the Generation and Detection of Synthetic and Manipulated Satellite Images}

\author[Lydia Abady, \textit{et al}.]{Lydia Abady$^{1}$, Edoardo Daniele Cannas$^{2}$, Paolo Bestagini $^{2}$, Benedetta Tondi $^{1}$, Stefano Tubaro $^{2}$ and Mauro Barni $^{1}$}

\address{\add{1}{Dipartimento di Ingegneria Dell’Informazione e Scienze Matematiche, Universit{\'a} di Siena, Siena, Italy}
\add{2}{Dipartimento di Elettronica, Informazione e Bioingegneria, Politecnico di Milano, Milano, Italy}}

\corres{\name{Lydia Abady, Edoardo Daniele Cannas}
\email{abady@diism.unisi.it, edoardodaniele.cannas@polimi.it}}

\begin{abstract}
%Satellites can be used to acquire a huge variety of images ranging from photographs in the visible domain to complex data acquired through radar system.
Due to the reduction of technological costs and the increase of satellites launches, satellite images are becoming more popular and easier to obtain.
%Due to the diverse and abundant usages of satellite images, the pace of satellites launches is increasing dramatically. A lot of these products are made available publicly for free to increase awareness and allow broader utilization of these imagery by the research community.
Besides serving benevolent purposes, satellite data can also be used for malicious reasons such as misinformation. As a matter of fact, satellite images can be easily manipulated relying on general image editing tools. Moreover, with the surge of \glspl{dnn} that can generate realistic synthetic imagery belonging to various domains, additional threats related to the diffusion of synthetically generated satellite images are emerging. In this paper, we review the \gls{sota} on the generation and manipulation of satellite images. In particular, we focus on both the generation of synthetic satellite imagery from scratch, and the semantic  manipulation of satellite images by means of image-transfer technologies, including the transformation of images obtained from one type of sensor to another one. We also describe forensic detection techniques that have been researched so far to classify and detect synthetic image forgeries. While we focus mostly on forensic techniques explicitly tailored to the detection of AI-generated synthetic contents, we also review some methods designed for general splicing detection, which can in principle also be used to spot AI manipulate images.
\end{abstract}

\maketitle

\thispagestyle{empty}
\let\thefootnote\relax\footnote{This material is based on research sponsored by the Defense Advanced Research Projects Agency (DARPA) and the Air Force Research Laboratory (AFRL) under agreement number FA8750-20-2-1004.
The U.S. Government is authorized to reproduce and distribute reprints for Governmental purposes notwithstanding any copyright notation thereon.
The views and conclusions contained herein are those of the authors and should not be interpreted as necessarily representing the official policies or endorsements, either expressed or implied, of DARPA or AFRL or the U.S. Government.}
\section{Introduction}
\label{sec:intro}

%\MB{My two cents on the title (see the new one I am suggesting). In fact, it is impossible to be 100\% precise given that: i) we do not focus on AI-based manipulations only, ii) we do not (I think) review exhaustively all possible kinds of manipulations (e.g. all those based on photoshop). Given that we focus {\em mostly}  on AI-manipulations, I opted for the title above. Then it is up to the main body of the paper to clarify the exact scope of the overview.}

As technology develops and  deployment costs reduce, satellites  are getting more and more appealing for accomplishing various tasks \cite{kostopoulos2020}. According to the
\gls{unoosa}, the number of launched satellites increased from few hundreds in 2019 to more than a thousand in 2020. These figures continued to grow throughout 2021, and the trend will likely  continue in the next year.

Some classical usages of satellite images include crops monitoring, urban expansion monitoring, meteorological forecasting, land cover mapping, just to mention a few \cite{purnamasayangsukasih2016}. Other less known applications include intelligence or military missions. For instance, satellite images have been used to fend off misinformation campaigns or investigate the truth of events in areas that are too menacing or difficult to reach. More often, among other things, they were used to record military forces deployment \cite{ukraine2022} and damages occurred to infrastructures in conflicts \cite{ukraine2022mariupol}.

Due to their strategic role, satellite images have often been the objective of malicious manipulations \cite{russia} \cite{australia_wildfire} \cite{bbc_ukraine}. As a matter of fact, simple manipulation of satellite data can be obtained with standard image editing tools such as Photoshop and GIMP. More sophisticated kinds of manipulation can be applied exploiting some of the latest deep learning findings.

In the last years, \glspl{dnn} have witnessed rapid improvements in their ability to forge digital contents~\cite{isola_2016}~\cite{zhu_2017}. The performance reached by these tools is such that nowadays the trustworthiness of any type of media we come across can genuinely be questioned. Satellite images are no exception. However, the direct application of common processing tools to the satellite imagery domain  is often not viable, 
%{\em But how far are we from having them also threatening the validity of satellite images?}
due to the different nature of these images with respect to standard images, e.g., their multi-spectral nature and their content, and the different needs stemming from remote sensing applications. This has lead to the development 
of dedicated techniques for the generation and manipulation of satellite imagery. As a consequence, dedicated  detection techniques to reveal  synthetic and manipulated images are also being developed.

%\BTcomm{Here we miss some steps. \\ 1. I would first explicitly say that this is the case also with satellite images and add some reference. \LA{this point was mentioned and referenced in the previous paragraph that talked about military force deployment}\\ 2. Then, I would say something on the different nature of satellite images, in terms of type of data (various bands, wider range for the values taken by the pixels with respect to standard images,....) and informative content, with respect to normal images, and say that this had lead to the development of dedicated techniques for the generation and manipulation  suitable for this particular domain. \LA{I rather not delve into the differences between satellite imagery and camera one even if the differences can be bit depth or number of bands and can impact a bit the architecture but it doesn't mean standard techniques don't work. As a matter of fact, they do work with slight modifications whenever not only RGB is being used}}

In this paper, we overview the most relevant methods developed so far for the generation and manipulation of remote sensing imagery,  with particular attention, but not exclusively, to techniques based on DNNs. We consider both the works focused on the generation of synthetic satellite images from scratch using \glspl{gan}, as well as those aiming at modifying existing satellite images. Data type translations of satellite images and techniques applied to improve the image quality such as colorization and cloud removal, are other examples where image processing techniques and DNN's are used to create synthetic data that do not stem directly from the sensors. Even though these methodologies are usually applied for benevolent purposes, as a matter of fact, their application results in non-genuine products, whose origin should be exposed to users. In the rest of the paper, we will generally refer to images whose content is not the direct result of the observation of the earth surface by an image sensor, as synthetic or manipulated  images. We will also loosely use terms like tampered, fake or forged images, even if the goal of the manipulation is not a malevolent one.

In the second part of the paper, we also overview \gls{sota} forensic techniques, that can be used to assess whether a satellite image is a pristine one or contains synthetically generated parts. The multimedia forensics community has a long and rich experience in the analysis of digital pictures \cite{Piva2013overview}. In recent years, a wide variety of techniques has been proposed to detect editing operations executed either on the whole image \cite{Popescu2005exposing, Kirchner2008fast, Bianchi2011detection, mandelli2018multiple} or locally \cite{Cozzolino2015splicebuster, Bayar2016deep, Bondi2017tampering, cozzolino2020noiseprint}. Moreover, the recent literature has also shown promising results in the detection of synthetically generated content \cite{bonettini2020use, mandelli2020training, alamayreh2021detection, Gragnaniello2022}.
Unfortunately, when it comes to the analysis of satellite images, many of the methods developed for natural images perform poorly due to the different nature of the to-be-analyzed data. For this reason, the multimedia forensics community has started to develop techniques specifically tailored to the analysis of imagery. In this context, we first dig into methods strictly tailored to detect satellite contents generated by Artificial Intelligence (AI) techniques. Then, as these areas are still underdeveloped, we also dig into forgery detection and localization techniques that have not been specifically proposed to spot AI-generated satellite contents, but that can, in principle, be used for such a goal.

The paper is organized as follows: in Section ~\ref{sec:bkgd}, we provide a  description of the satellite images data types that we will consider  throughout the paper, and the most popular datasets of satellite images. We also introduce  the main DNN architectures that were used for image generation and manipulation in Section \ref{sec:generativemodels}. Then, in Section ~\ref{sec:forgeries}, we overview the methods that have been proposed in the literature to generate and modify the content of satellite images. In Section ~\ref{sec:bforgeries} we go beyond forgeries and focus on techniques that are meant to edit satellite images without necessarily altering their semantic content. We follow with the detection techniques in Section ~\ref{sec:detection}. In Section VII, we critically review the state of the art and highlight the open challenges researchers are still being faced with.  Lastly, we conclude our work with some final remarks in Section ~\ref{sec:conclusion}.
%

%\BTcomm{{\bf I added some considerations in the intro especially  pertaining the motivation part.}}
%\PB{No strong suggestions on the intro at the moment. The only (minor) thing that sounds a bit weird is that we first say that satellite images can be edited with GIMP, but then we say they are interesting because of specific forgeries operations. I'll double check if we can add something and slightly rephrase the GIMP part as soon as I'm done with the complete document if you agree.} \BTcomm{OK}

%\section{Background and Taxonomy}
\section{Remote Sensing Imagery}
%TODO add mentioned archs and datatypes if missing
\label{sec:bkgd}
%{ \em Most of the methodologies reviewed in this paper have in common the type of remote sensing data or the same base architecture of \gls{nn} used for the generation of synthetic images. Hence, we provide a definition for each satellite imagery data type considered in the various works analyzed in this paper. In particular, we focus also on the different modality of satellite data commonly used in the majority of remote sensing applications, in order to provide a common ground for forensics practitioners and facilitate the discussion on the forensics analysis of satellite data. After, we provide a more theoretical background on the main \gls{nn} architectures used for the generation of synthetic satellite imagery.}
%\PB{minor comment: shall we show some example EO and SAR pictures during their description?}

In this section, we introduce the satellite imagery data types considered in the various works overviewed in this paper, along with the data sources from which the datasets were collected.
%We also introduce the most important \gls{nn} architectures used for the generation of synthetic images.

\subsection{Data Types}
\label{subsec:rsimod}
The term remote sensing indicates a broad variety of measurements of electromagnetic radiations interacting with the Earth's atmosphere and surface, in order to collect information about an object or phenomenon without being in direct contact with it.
On one hand, this kind of measurements can provide information on the distance between the sensors measuring the radiation and the object interacting with it. On the other hand, by analyzing different quantities related to the measured radiation (e.g., intensity, wavelength, polarization, etc.) they can provide clues about the properties and characteristics of the interacting object.

% \EC{Structure section to write: 1. difference between imaging and non-imaging modalities and between passive (i.e., EO imagery) and active (i.e., SAR) sensor; 2. Subsection on EO imagery (i.e., panchromatic, multi-spectral), with focus on the modality of acquisition (i.e., linear push-broom array, overlap camera array) and the post-processing following); 3. Subsection on SAR, a resume from this article \cite{cannas2022amplitude} might suffice}
Remote sensing data can be acquired by using sensors belonging to two main families: passive sensors (i.e., sensors relying on solar radiations to detect the reflections from the Earth's surface), and active sensors (i.e., sensors providing their own source of energy to execute the measurement) \cite{Toth2016remote}. Examples of data generated by passive sensors are \gls{eo} imagery, while examples of actively generated satellite images are \gls{sar} signals.
Other characteristics differentiate these sensors, the main one being the spatial resolution of the imaged data which is related to the sensor's spectral sensitivity.
%Figure \ref{fig:em_spectrum} provides a graphical representation of the most relevant bands of the electro-magnetic spectrum for remote sensing applications, and

\begin{figure*}[t]
\centering

\subfloat[\gls{sar} image example, vertical-vertical polarization.]{\includegraphics[width=0.5\columnwidth]{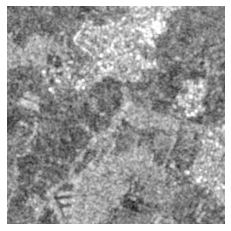}
\label{fig:sar_example}}
\hfil
\subfloat[\gls{eo} image example, RGB.]{\includegraphics[width=0.5\columnwidth]{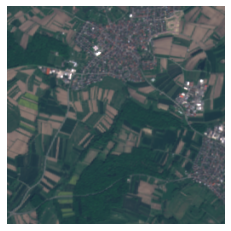}\label{fig:rgb_example}}
\hfil
% \subfloat[Panchromatic image example.]{\includegraphics[width=0.5\columnwidth]{diagrams/modalities_examples/pan_example.png}\label{fig:pan_example}}
% \hfil
\subfloat[Multi-spectral image example, \gls{swir} band.]{\includegraphics[width=0.5\columnwidth]{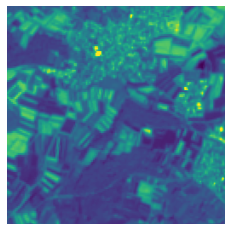}\label{fig:swir_example}}

\caption{Examples of different overhead image modalities from the same scene. From left to right, \gls{sar}, \gls{eo} and multi-spectral.}
\label{fig:modalities_example}
\end{figure*}

In the following, we describe the main satellite modalities analyzed in the literature when it comes to the forensic analysis and synthetic generation of remote sensing imagery, i.e., \gls{eo} and \gls{sar}.
Fig.~\ref{fig:modalities_example} provides some examples of the image modalities considered in this work.
%\PB{Minor suggestion: is there any other strong statement we can make to justify the fact that we focus on EO and SAR? Something like ``they are the most widespread kind of satellite images'', ``they are the most different kinds of satellite images compared to consumer photography'' (of course if this is true).}
% Remote imaging sensors can be mounted on different platforms, such as aircrafts, Unnmanned Aircraft Systems (e.g., drones), and of course satellites.
\begin{comment}
\begin{figure*}
\centering
\includegraphics[width=2\columnwidth]{diagrams/1-s2.0-S0924271615002270-gr3_lrg.jpg}
\caption{Electro-magnetic spectrum with remote sensing imagery most relevant bands, taken from \cite{Toth2016remote} \BTcomm{I think we should change a bit the figure since we can not use figures from other papers. The same comments apply to other figures, if you took them from papers as they are.}.}
\label{fig:em_spectrum}
\end{figure*}
\end{comment}

\subsubsection{Electro-Optical Imagery} with this term we refer to satellite images obtained through passive sensors capturing the solar radiations reflected by the Earth's surface. The first images of this kind were not too dissimilar from natural photographs: they captured light in the visible spectrum thanks to a rigid body camera holding the optics, and a sensor placed directly on the focal plane \cite{Toth2016remote}. Today's \gls{eo} sensors are still optical-based systems, but the technology has progressed to accommodate greater object ground coverage and spatial resolution as required by modern remote sensing systems \cite{Toth2016remote}.

\gls{eo} imagery relies mainly on solid-state chip sensors such as \gls{ccd} and \gls{cmos}, but in different configuration with respect to those  usually employed in consumer photography. Indeed, since the size of such sensors cannot achieve large ground-coverage with a high spatial resolution, they are arranged in linear arrays or area sensors \cite{optic} on satellites and airborne platforms.

Linear arrays acquire samples with a push-broom modality, i.e., as the platforms moves along its trajectory, long ``strips'' of pixels, called pixel carpets, are acquired and then ``stiched'' together to form a single image covering the area of interest. This is the modality adopted, for instance, by  Maxar satellites like WorldView2 \cite{wv2technical}. The second modality, which is employed, for example, by the PlanetLab's Skysat constellation \cite{skysattechnical}, relies on one or two-dimensional sensors that do not acquire images as pixel carpets, but as normal frames, i.e., the scene is imaged at the same moment in time with its pixels having the same rigid relationship with respect to each other. Coverage of areas of interest is then obtained by shooting blocks of overlapping photos. In both cases, strong processing follows to provide final users with more manageable data known as products. Examples of processing are radiometric and sensor correction \cite{Teillet86image}, or orthorectification \cite{ortho}.

Usually, passive sensors are able to capture wavelengths in the visible spectrum (i.e., Blue light with wavelength around $\SI{0.4}{\mu m}$) up to \gls{lwir} (i.e., wavelength up to $\SI{20}{\mu m}$).
These wavelengths can be captured either together, or as different channels or bands in the acquired image through the use of high-quality filters, leading to the formation of different kinds of \gls{eo} products. Panchromatic imagery, for instance, is a monochromatic format of remote sensing data with no spectral information but with high spatial resolution. RGB data
%separates light instead in the red, green and blue spectra, providing more information for discriminating the land-cover content.
collects information in the red, green and blue bands, that are often used, in combination with bands outside the visible spectrum, for discriminating the land-cover content. %\BTcomm{What you mean exactly here? 'discriminate land-cover types'? aren't other bands also useful for that?\LA{It is a combination so amended the sentence}}.
Finally, \gls{msi} can be used to collect spectral information up to the \gls{lwir} bands divided in 8-15 channels, while \gls{hsi} can cover the same spectral range but with a higher spectral resolution. \gls{hsi} can reach up to 100 spectral bands and it is usally employed in more demanding tasks where materials signatures must be accurately identified \cite{Harsanyi1994hyperspectral}.

\subsubsection{Synthetic Aperture Radar (SAR) Imagery} with this term we indicate a remote sensing modality obtained through active sensors, (i.e., imaging radars mounted on moving platforms). As the system travels, it emits sequential high power electromagnetic pulses. These pulses, called chirps, are characterized by constant amplitude and linearly modulated instantaneous frequency. Chirps interact with the Earth's surface, being reflected as back scattered echoes with amplitude and phase changed according to the characteristics of the objects they hit (e.g., permittivity, geometry, roughness, etc.). The platform receives and collects these echoes, and after a series of processing operations required to make the image interpretable (i.e., focusing \cite{Moreira2013}), a 2D matrix of complex values is returned \cite{Moreira2013}. 
%\PB{Let's add a generic SAR description reference (a book?). It's weird to have an entire intro paragraph on SAR without it.}
%
Since the ground coverage of a single echo is often insufficient, \gls{sar} images are typically acquired by collecting and concatenating different measurements together in order to cover the whole area of interest. Other processing steps can then follow in order to project the images on the Earth's surface (e.g., ground-range projection \cite{groundrange},  orthorectification \cite{ortho}, etc.).

\gls{sar} data has gained a lot of popularity due to the fact that, with respect to \gls{eo}, is not affected by daylight, weather and cloud coverage conditions \cite{Tomiyasu1978, Oliver2004}. This makes it suitable for a variety of applications in substitution and/or integrated with \gls{eo} imagery, such as Earth monitoring, change detection, or Earth surface mapping \cite{Moreira2013}.

As with \gls{eo} imagery, also SAR data is distributed in different formats with different characteristics. Each format is know as a product, and may be useful for different applications. For instance, different frequency bands for modulating the chirps can be used, with the most popular being
L (i.e., from 1 GHz to 2 GHz), C (i.e., from 3:75 GHz to
7:5 GHz) and X (i.e., from 7:5 GHz to 12 GHz). Each of them is more suited for different applications, e.g., X for military surveillance, S for medium range meteorological applications, etc \cite{Moreira2013}. Other products can also be obtained by further processing the complex
2D matrix. Sentinel-1 \gls{grd} images, whose pixel values approximate the reflectivity of the ground \cite{Oliver2004}, is an example of an amplitude-based product. 

\subsection{Data Sources}
\label{ssec:rsimagery}
%\PB{minor suggestion: what if we simply change the order of the following datasets? Some of them are portals, some are not, and this looks a bit confusing.}
%\PB{minor suggestion: what if we put EO and/or SAR in parenthesis next to the dataset name to make it clear which kind of data it contains?}

%{\em Due to the rapid development of the Internet as the main communication infrastructure used in our daily life, we have seen a huge increase in the amount of multimedia objects shared among people. This is true also for remote sensing imagery: as a matter of fact}

The huge increase in the amount and variety of multimedia content shared among people is a phenomenon that also affects remote sensing data.
Nowadays, there are many online platforms offering satellite data for free in the form of easy-to-manage products \cite{freesat}. The purpose of this section is to present some of the most commonly used data collections acquired through popular data acquisition missions and available through different online portals.

% TODO: introduce the variety of image types: optical and sar: https://www.intechopen.com/chapters/57384
\subsubsection{\gls{aviris} Data Portal}
\gls{aviris} refers to an optical sensor realized and operated by NASA that gathers spectral radiance of 224 contiguous visible and \gls{nir} spectral bands with wavelengths ranging from 400 to $\SI{2500}{\mu m}$ \cite{aviris_2022}. The data is gathered using four aircrafts platforms. Until now, the area of coverage is North America, Europe, portion of South America, and Argentina. The \gls{aviris} project is focused on studies related to climate change and global environment.
\subsubsection{Copernicus Open Access Hub}
The Copernicus Open Access Hub \cite{copernicus} is the online portal provided by the \gls{esa} to download products generated by one of the Sentinel missions. In particular, Sentinel-1 and Sentinel-2 have gained a lot of popularity. % in the years.

\noindent\wtitle{Sentinel-1} mission provides C-band \gls{sar} imaging \cite{sentinel1_2022}
% operating in four exclusive modes that is able to retrieve data regardless of the time and weather.
with two satellites, Sentinel-1A and Sentinel-1B, with a revisit frequency of 6 days for both and 12 days for a single satellite.
% They operate continuously on \gls{wv} mode over open oceans and it has only a single polarization either \gls{vv} or \gls{hh}. Where as \gls{iw} operates over land and coastal areas with dual polarity while \gls{ew} operates over seas and polar areas with dual polarity. The operating predefined mode is either \gls{iw} or \gls{ew} however in the case of emergency observation requests, it might change into \gls{sm}. The data products are available publicly for users from level-0 \gls{sar}, level-1 \gls{slc}, level-1 \gls{grd}, and level-2 \gls{ocn}.\\
Sentinel-1 images are available through the Copernicus Open Access Hub \cite{copernicus} in different products. For instance, different acquisition modes (i.e., patterns of movements through which the antenna emits electromagnetic pulses) are available. The simplest one is the Stripmap, where pixel carpets are sensed with a fixed antenna pattern, while a more complex variations is the \gls{iw}, where the system emits three chirps steering the antenna in the platform moving direction. Other differences between products are related to the level of processing they undergo. The Open Access Hub offers them in 3 different levels: level 0 consists of unfocused \gls{sar} echo signals; level 1 consists of focused \gls{sar} images, provided either as complex signals as \gls{slc} or as amplitude only signals as \gls{grd} products; additional levels offer even more processing.
%

% \subsubsection{Sentinel 2 \gls{msi}}
\noindent\wtitle{Sentinel-2} mission provides \gls{msi} in 13 bands for land monitoring usage \cite{sentinelmsi_2022}. As for Sentinel-1, the images ase provided by a pair of twin satellites that has a revisit frequency of 5 days for regular coverage areas. Table \ref{tab:sentinel_bands} shows the spatial resolution of all the multi-spectral bands: 4 bands have a resolution of 10m \gls{gsd}, 6 of 20m \gls{gsd}, and 3 of 60m \gls{gsd}. Sentinel-2 \gls{msi} is offered in 5 different products, 3 of which are not publicly available (i.e., level-0, level-1A and level-1B). The orthorectified products level-1C (i.e., \gls{toa}) and level-2A (i.e., \gls{boa}) are freely available to all users.
\begin{table}[htbp]
\begin{center}
\begin{tabular}{|l|l|}
\hline
Spatial Resolution (m) & Band Number               \\ \hline
                       & 2 - Blue                  \\ \cline{2-2}
                       & 3 - Green                 \\ \cline{2-2}
                       & 4 - Red                   \\ \cline{2-2}
\multirow{-4}{*}{10}   & 8 - NIR                   \\ \hline
                       & 5 - Vegetation Red Edge 1 \\ \cline{2-2}
                       & 6 - Vegetation Red Edge 2 \\ \cline{2-2}
                       & 7 - Vegetation Red Edge 3 \\ \cline{2-2}
                       & 8a - Narrow  NIR          \\ \cline{2-2}
                       & 11 -  SWIR 1              \\ \cline{2-2}
\multirow{-6}{*}{20}   & 12 - SWIR 2               \\ \hline
                       & 1 - Coastal Aerosol       \\ \cline{2-2}
                       & 9 - Water vapour          \\ \cline{2-2}
\multirow{-3}{*}{60}   & 10 - SWIR Cirrus          \\ \hline
\end{tabular}
\end{center}
\caption{Spatial Resolution of Sentinel-2 MSI bands. 
%\PB{Do we want to put captions above or below tables?\LA{below}}
}
\label{tab:sentinel_bands}
\end{table}

\subsubsection{Maxar DigitalGlobe Portal}
The DigitalGlobe Discover portal \cite{digitalglobe} is an online platform for downloading satellite imagery produced by Maxar technologies. Maxar counts 7 satellites in its constellation providing \gls{eo} imagery, 4 on orbit (i.e., WorldView1-2-3 and GeoEye1) and 3 decommissioned (i.e., QuickBird, Ikonos and WorldView4), whose images, however, are still available in the portal archive. According to the technical guide \cite{digitalglobeuserguide}, the \gls{gsd} resolution can vary according to the type of imagery produced: for panchromatic products, the ground resolution varies from \SI{50}{cm} to \SI{2}{m}, while for \gls{msi} it ranges from \SI{2}{} to \SI{2.4}{m}.

\subsubsection{U.S. Geological Survey Landsat Data Access} The \gls{usgs} offers a portal \cite{usgeologicalysurvey} to download all the imagery produced by the Landsat program. The Landsat program \cite{landsat} is a joint mission by NASA and \gls{usgs} aiming at monitoring the Earth surface for any remote sensing application, which has been active now for more than 50 years. It comprehends two twin satellites (i.e., Landsat8 and Landsat9) providing \gls{msi} in the visible, near and short-wave infrared spectra, as well as thermal infrared wavelengths, with a ground sampling distance of \SI{30}{m}.

\subsubsection{SEN12MS}
The SEN12MS dataset \cite{schmitt_2019} is a large scale satellite dataset aimed at training deep learning architectures for land-cover related applications (e.g., land-cover classification, segmentation, etc.). It is made up of 180,662 $256\times256$ triplets of Sentinel-1 \gls{sar} image patches, multispectral Sentinel-2 image patches, and MODIS \cite{modis} land cover maps. The images span various locations and seasons and have been processed to provide the information related to the same exact location, i.e., to have similar \gls{gsd} and georeference information. In particular, since Sentinel-2 \gls{msi} is orthorectified, Sentinel-1 data has been orthorectified too, while the MODIS land cover maps have been upsampled to reach a resolution of \SI{10}{m} \gls{gsd} from the original resolution of \SI{500}{m}. \cite{schmitt_2019}\\~\\

%\MB{Perhaps Generative Models could be put in a separate section \LA{They are in separate subsection but I think they should remain in the background section}}

%\MB{It is not mandatory, but the reasons for putting them in separate section are: i) these images are of a completely different kind, ii) section II is very long, splitting it would ease the reading}

\section{Generative Models}
\label{sec:generativemodels}

A \gls{gan}~\cite{goodfellow2014gans} architecture provides a game theoretic framework where two networks, namely a generator and a discriminator, are trained in an adversarial manner. Starting from an input noise sample $z$, the generator synthesizes new samples with a distribution $p_g$ that is similar to the distribution of some source data  $p_x$ on which the discriminator is trained. The discriminator analyzes these samples trying to judge whether they are real or synthesized samples produced by the generator.
% The architecture takes advantage of two adversarial \glspl{cnn}, one network that acts as the generator and the other as the discriminator.
The goal of the  generator is to produce samples that can not be distinguished from the real ones by the discriminator. Formally, the generator aims at minimizing the following loss function, i.e., the adversarial loss,
\begin{equation}
\label{eqn:genloss}
%\min_{\genNet}
\emph{L}_{G}(\disNet,\genNet) = \mathbb{E}_{z\sim{p_z(z)}}[\log(1 - \disNet(\genNet(z)))],
\end{equation}
where $\genNet$ is the generator network function and $\disNet$ is the discriminator network function.
%\BTcomm{You should say what is the input and output of these functions.\LA{it is already implicitly stated above when we say that the generator synthesizes new samples starting from an input noise and following when we say the discriminator analyzes samples}}
% and $z$ is the input noise variable with noise distribution $p_z$.
The discriminator's goal, instead, is to distinguish  between generated samples and real ones. Hence, it tries to minimize the following loss function, where $x$ is the input sample, drawn from  $p_x$: 
%\BTcomm{I rephrased it as a negative loss so that the D wants to minimize instead of maximize it, since it seems that after you use this convention (that is, $\emph{L}_{D}$ is the loss function that D wants to minimize)\LA{ok, although I prefer to keep it as the original paper formulation}}
\begin{equation}
\begin{split}
\label{eqn:disloss}
%\max_{\disNet}
\emph{L}_{D}(\disNet,\genNet) = {-} & \mathbb{E}_{z\sim{p_z(z)}}[\log(1 - \disNet(\genNet(z)))] -\\&
\mathbb{E}_{x\sim{p_{x(x)}}}[\log(\disNet(x))],
\end{split}
\end{equation}
where the first term makes sure that the discriminator recognizes generated samples $\genNet(z)$ as such, and the second term makes sure that it recognizes samples $x$ as original ones belonging to $p_x$.
%
%\BTcomm{You should say about the role of both terms.
%\LA{I don't understand this comment, we say above that the discriminator obj is to distinguish the two classes}}\PB{I tried adding a sentence}
\begin{comment}
Fig. ~\ref{fig:gan_basic} illustrates the architecture of the basic \gls{gan} applied to a satellite image generation task for the case of RGB data image generation.
\begin{figure}[htbp]
\centering
\includegraphics[clip,width=0.9\columnwidth]{diagrams/gan_basic.pdf}
\caption{Basic GAN architecture for RGB satellite image generation.\BTcomm{I think we can remove this figure. Don't say much. Then, it considers the specific RGB case, that is the common image domain, instead of the multi-spectral one.\LA{I prefer to keep this figure as most work does not really tackle multi-spectral domain but RGB of satellite images content}} \BTcomm{It does not say much and is also not very clear. The GAN loop is not highlighted. Also, you mention the latent space that is not mentioned in the text.}\LA{commented out}}
\label{fig:gan_basic}
\end{figure}
\end{comment}
\glspl{gan} have been applied successfully to a variety of different domains, from text \cite{Yu2017seqgan}, to biomedical images \cite{Kazeminia2020gans} and of course natural images. In particular, a great effort have been paid to continuously improve the quality and realism of synthetic natural images ~\cite{karras_2017}~\cite{karras_2018}~\cite{karras_2019}. Due to the impressive results they got and the high realism of the generated images, many variants of the basic \glspl{gan} framework were developed, going beyond image generation from scratch, that are widely used in various computer vision tasks, and also for satellite imagery applications. In the following, we describe the most relevant architectures used in the literature.

One of the most relevant \glspl{gan}-based frameworks is image-to-image translation \cite{isola_2016} \cite{zhu_2017}, wherein the input image is translated from a semantic domain to another. Image-to-image translation is based on \gls{cgan} \cite{Mirza_2014_cgan}, an evolution of the basic \gls{gan} architecture where the training procedure is modified with the addition of a condition on the inputs of either the generator or the discriminator or both. This condition can derive from any kind of additional information.
% like a label $y$ or additional data needed to perform the task at hand.
%
In \cite{isola_2016}, a popular \gls{cgan}, named pix2pix is proposed to improve the quality of the generated images.
To train  the pix2pix architecture, a {\em paired} dataset is used where each input image has its corresponding representation in the target domain, namely a reference image $r$ following a distribution $p_{xr}$. For example, if we want to transfer images from summer to winter, we would need images of the same place belonging to both the summer and winter domains where one will act as the input (summer) to be transferred by the generator and the second is the reference that the network will try to simulate from the input. 
%\MB{Make an example to better clarify the concept}
% In addition to the adversarial loss, t
The authors add a term corresponding to the $L1$ distance between the reference images and the generated images, to the loss of the generator, which is now expressed as:
\begin{equation}
\begin{split}
\label{eqn:pix2pixgenloss}
%\min_{\genNet}
\emph{L}_{G}(\disNet,\genNet) = {} &\mathbb{E}_{x\sim{p_{xr}(x,r)}}[\log(1 - \disNet(x,\genNet(x)))] +\\&
\lambda \mathbb{E}_{xr\sim{p_{xr}(x,r)},}[\left\lVert(r - \genNet(x)))\right\rVert_1],
\end{split}
\end{equation}
where $\lambda$ is a weight parameter balancing the importance of the two loss terms %\MB{Shouldn't $x$ and $r$ be generated dependently by a joint pdf $p_{xr}(x,r)$?\LA{I think they are dependent but in the original referenced paper they don't mention the dependency so I kept similar terms and distributions}}. \MB{Don't care what the original paper does, if $x$ and $r$ are generated dependently then use $p_{xr}(x,r)$. By the way this is on of the main differences between pix2pix and cycleGAN where the it is not possible to generate {\em paired} $x$ and $r$ \LA{ok}} 
 $\emph{L}_{D}$ is the same as in \eqref{eqn:disloss}.
%\BTcomm{And what about $\emph{L}_{D}$?\LA{It was kept the same}}

\begin{figure}[htbp]
\centering
\includegraphics[clip,width=0.9\columnwidth]{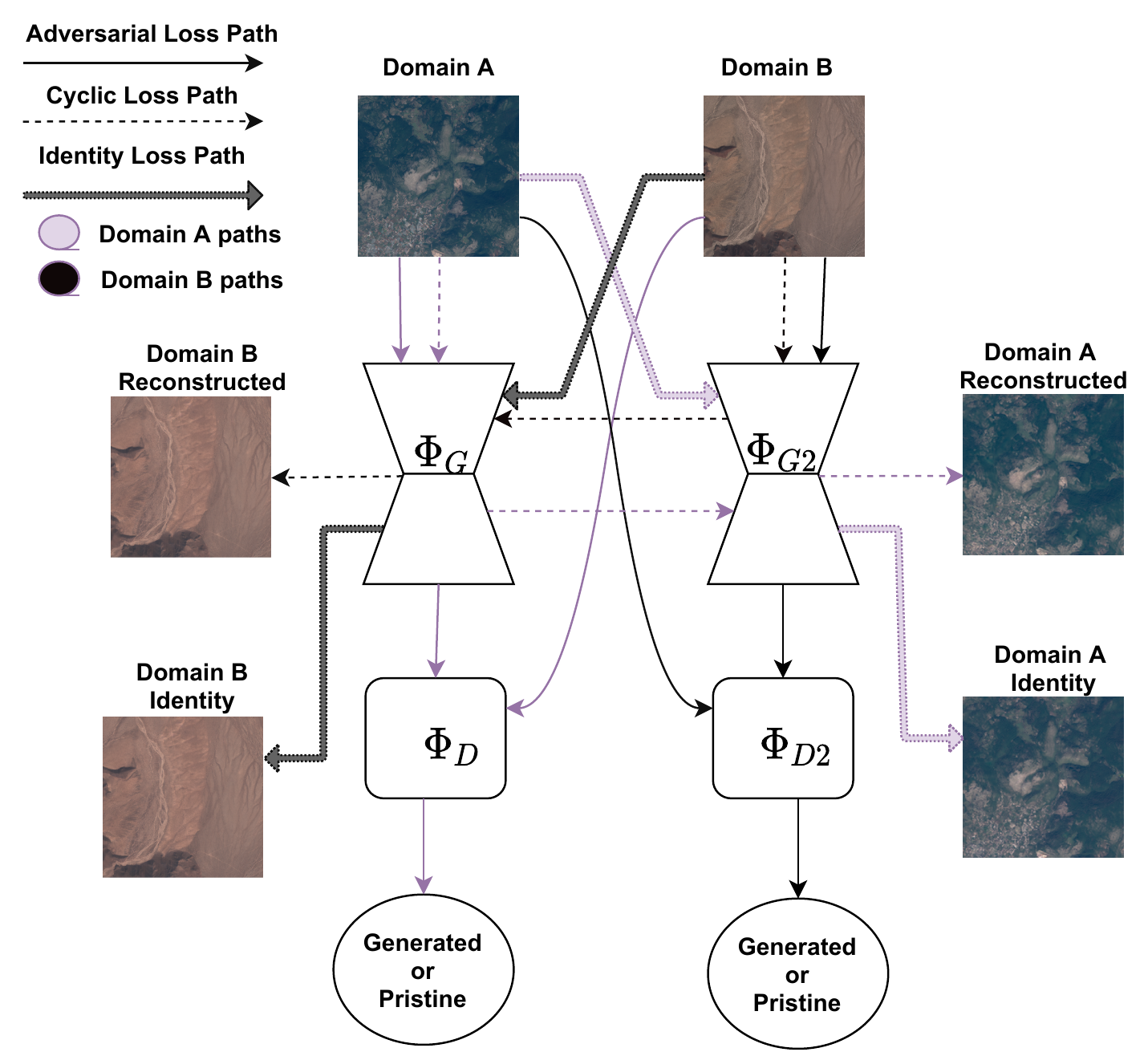}
\caption{cycleGAN Architecture}
\label{fig:cyclegan+_arch}
\end{figure}

Another very popular \gls{cgan} is the \gls{cyclegan} \cite{zhu_2017}, whose general architecture is shown in Fig. \ref{fig:cyclegan+_arch}. It replaces the $L1$ distance loss term of  pix2pix with a so called cycle consistency loss term, computed by resorting to two generators and two discriminators.  The  cycle consistency loss is defined as
\begin{equation}
\begin{split}
\label{eqn:cycleconsisloss}
\emph{L}_{cyc}(\genNet,\sgenNet) = {} &\mathbb{E}_{x\sim{p_x(x)}}[\left\lVert(\sgenNet(\genNet(x))) - x)\right\rVert_1] +\\&
\mathbb{E}_{y\sim{p_y(y)}}[\left\lVert(\genNet(\sgenNet(y))) - y))\right\rVert_1],
\end{split}
\end{equation}
where $x$ (drawn from $p_x$) denotes a  sample from the first domain, given as input to the first generator $\genNet$ and the first discriminator ($\disNet$),
%$\disNet$, 
and $y$ (drawn from $p_y$) is a sample from the second domain, given as input to the second generator $\sgenNet$ and the second discriminator ($\sdisNet$).

An advantage with respect to pix2pix is that with \gls{cyclegan} there is no need for a paired dataset for training (that can be difficult, if not impossible, to collect in many applications).

An additional constraint, known as identity loss, can also be added to the loss of the generator. The goal of the identity loss is 
to ensure that the output of the generator is equal to its input, when a sample $x$ of the first domain  is fed at the input of the second generator $\sgenNet$. The same applies when a sample of the second domain is fed to the first generator $\genNet$. Formally, the identity loss has the following expression:
\begin{equation}
\begin{split}
\label{eqn:idloss}
\emph{L}_{identity}(\genNet,\sgenNet) = {} &\mathbb{E}_{x\sim{p_x(x)}}[\left\lVert(\sgenNet(x)) - x)\right\rVert_1] +\\&
\mathbb{E}_{y\sim{p_y(y)}}[\left\lVert(\genNet(y)) - y))\right\rVert_1].
\end{split}
\end{equation}
%\CHLA{Figure \ref{fig:cyclegan+_arch} shows the cycleGAN architecture.}
Fig. \ref{fig:cyclegan+_arch} shows the cycleGAN architecture.
%\CHLA{The adversarial losses and the discriminator losses are kept as they are.}

A variant of \gls{cyclegan} is  the \gls{nicegan} \cite{chen2020}, where, instead of designing  a dedicated encoder for the generator, the first layers of the discriminators are used as the generator encoding layers. Hence, the generator and the discriminator share some common layers.

Another line of research aims at improving not only the generated sample quality, but also the training process, to mitigate the problems of convergence instability that often affects \glspl{gan}. One of the methods proposed to achieve this goal is the \gls{wgan-gp} \cite{Gulrajani2017}, where the Wasserstein loss formulation, in which the discriminator acts as a critic and increase the distance between the real and fake samples instead of classifying the images as real or fake, is considered (WGAN \cite{arjovsky2017_cwgan}) and a gradient penalty is added to the discriminator's loss to fulfill a Lipschitz constraint. The loss for the generator and the discriminator are defined as
\begin{equation}
\label{eqn:wgangenloss}
%\min_{\genNet}
\emph{L}_{G}(\disNet,\genNet) = -\mathbb{E}_{z\sim{p_z(z)}} [\disNet(\genNet(z))],
\end{equation}
and
\begin{equation}
\begin{split}
\label{eqn:wgandisloss}
%\min_{\disNet}
\emph{L}_{D}(\disNet,\genNet) = {} & \mathbb{E}_{z\sim{p_z(z)}} [\disNet(\genNet(z))] -\\&
\mathbb{E}_{x\sim{p_{x(x)}}}[\disNet(x)] +\\&
\lambda\mathbb{E}_{w\sim{p_w(w)}}[(||\nabla_w\disNet(w)||_2 - 1)^2],
\end{split}
\end{equation}
where $\lambda$ is the gradient penalty  tradeoff and $w$ is a random sample either produced by the generator or taken from the distribution of real samples.

%\BTcomm{it seems that other things changed beside the addition of the Lipschitz constraint, Why don't you have the same terms as in Eq. 1 and 2?? Why there a difference in these terms as well? If there is not an error we should explain/say why. \LA{no, it is basically the same, I used the equations from the relative papers but practically minimizing equation 1 is actually equivelant to minimizing equation 6 and the relation is similar between equation 2 and 7, I can readd same equations terms but I rather keep the related papers' form}
%}

%\MB{You never spoke about the stability issue. Explain}
Another approach that allows to mitigate the instability of \glspl{gan} training, to enhance the quality of the generated images, and also to speed up the training, is the progressive training methodology. The main feature of the \gls{progan} \cite{karras_2017} is the incremental  approach, with the size of the model increasing incrementally during training. The training starts on small resolution data, typically $4\times4$ pixel images, then, during training, additional convolutional layers are added to  both the generator model and the discriminator models to increase the resolution.

In addition to \glspl{gan}, another widely used generation framework builds upon \glspl{vae} \cite{Kingma2014}.
An autoencoder \cite{Hinton1993_autoencoders} $\emph{A}$ is a neural network trained to reconstruct at the output the same data given as input, after processing it with a series of operations that avoid learning the identity function by respecting some constraints (e.g., reducing the dimensionality of the data at some point in the network). The autoencoder is composed by two blocks:
\begin{itemize}
    \item the encoder $\emph{A}_\text{e}$ that maps the input $x$, to a hidden representation $h$ (i.e., $h = \Phi_\text{Enc}(x$)).
    \item the decoder $\emph{A}_\text{d}$ that has a specular architecture compared to the encoder, and that maps the hidden representation to an approximate version of the input $\tilde{x}$ (i.e., $\tilde{x} = \Phi_\text{Dec}(h)$).
\end{itemize}
In case of tensor data, the input can be a RGB image $\Xrgb$ and the hidden representation a vector $\mathbf{h}$, with the encoder and decoder trained together to minimize a reconstruction loss, typically, an L2 loss term, between the input samples and the output (decoded) sample.
%A \gls{vae} \cite{Kingma2014} is an architecture composed by two neural networks, that is, an encoder, that learns the distribution of data, and a decoder, that reconstructs the data given a sample of latent space selected from the distribution.

%\BTcomm{I have removed the text below since it seems that we are going in too much level of details. What you could do is to put a figure of the VAE as you did for the GAN (since you wanted to keep that figure, you could have one also for the VAE)\\\LA{I commented out the gan figure since you suggested to remove it anyway and about the equations, isn't it betterto keep them since we added them in all the needed cases?}}\BTcomm{{\bf I do not think they are necessary. But if you insist. In that case, add both GAN and the VAE, clearly hightighing the difference between the two frameworks.\LA{I think we can remove both as originally suggested since it is not the focus of the overview}}}
%in addition to a regularization term.
%The encoder encodes the input into a Gaussian distribution with mean $E$ and variance $V$ over the latent space instead of just a single point like the standard autoencoder architecture. Then a sampler will sample a latent variable from the Gaussian distribution that can be used as input to the decoder which in turn outputs a reconstruction of the initial input data. The total loss is represented by}
In \glspl{vae}, the input is not only being encoded into a vectorial representation, but the hidden variable is forced to follow a Gaussian distribution $\mathcal{N}(f(x); g(x))$, with mean $f(\cdot)$ and variance $g(\cdot)$ being functions of the input implemented by the encoder network. At the decoding stage, a sample from the hidden representation variable distribution is drawn and used as input to the decoder which in turn outputs a reconstruction of the initial input data.
The main intuition behind this approach is to allow the decoder to generate new data by sampling from the hidden variable distribution. From this perspective, the hidden variable distribution can be assumed to have some desirable properties, e.g., being a Gaussian normal distribution. In this scenario, the total loss iused during training is equal to:
\begin{equation}
\begin{split}
\label{eqn:vaeloss}
\emph{L}(x,\tilde{x})  = {} & \left\lVert x-\tilde{x}\right\rVert^{2}_{2} +\\&
\beta L_{kl}(\mathcal{N}(f(x), g(x)), \mathcal{N}(0, I_d)),
\end{split}
\end{equation}
where the first term represents a ``data fidelity term", i.e., a L2 loss between the input sample $x$ and the estimated sample $\tilde{x}$. The second term, applies a kind of ``regularization", by forcing the network to minimize the Kullback–Leibler divergence $L_{kl}$ between the learned hidden variable distribution and a desired normal $\mathcal{N}(0; I_d)$ distribution, with $I_d$ being the identity matrix of $d$ dimensionality, and $\beta$ a hyper parameter weight.

%\BTcomm{Do we mention VAE in the paper? It seems that we do not name it any more. Is this correct? In that case, we should perhaps remove it from here. Or we could say that VAE could also being used for the generation. Hovever the literature only conisders GAN.... }

%\MB{You still need to say how you used a VAE to generate synthetic images}
After training the \gls{vae}, the decoder is used to generate new images by picking random samples from the learned distribution.

\section{Satellite Forgeries via Deep Neural Networks}
\label{sec:forgeries}

In this section, we overview the most relevant methods for the generation and manipulation of satellite imagery, with particular attention to those based on \glspl{gan} and VAEs.
As mentioned in the introduction, due to the different characteristics of satellite images and to the different needs of remote sensing
applications, a number of dedicated methods have been developed, which are suitedfor this kind of images.

In the following, we classify the various methods based on the type of forgery they aim at.
In particular, we are considering the following types of forgeries: i)  generation from scratch of synthetic satellite images (addressed in Section \ref{ssec:scratch}) and
ii)  modification of the semantic content of pre-existing satellite images (Section \ref{ssec:semantic}).

\begin{figure}[htbp]
\centering
\includegraphics[clip,width=0.9\columnwidth]{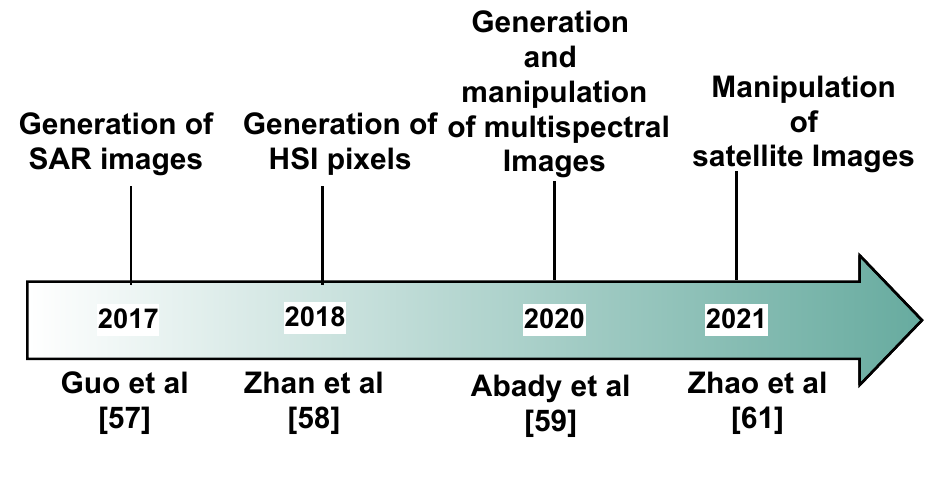}
\caption{{Timeline showing the first satellite generation methods for each datatype}}
\label{fig:forgtimeline}
\end{figure}

Table \ref{tab:forgeries} summarizes all the generation methods
considered in Section \ref{sec:forgeries} of this overview, categorized according to the proposed classification.
Additionally, Figure \ref{fig:forgtimeline} reports a visual representation of the timeline of the seminal works related to satellite image generation for each datatype.
%\BTcomm{AS you see, I am proposing to add a table similar to the one that we have in Section IV, that would be nice. We can have the following columns (just a proposal): 'Reference' 'Year' 'Kind of data' 'Type of forgery' (that should be one of the 4 categories we identified), and in case a more specific 'Goal/Task'.\LA{ok I added a table with the suggested columns}}. 

\begin{table*}[]
\centering
\resizebox{\textwidth}{!}{%
\begin{tabular}{|l|l|l|l|l|}
\hline
\textbf{Reference}                                  & \textbf{Year} & \textbf{Task}                       & \textbf{Data type}                       & \textbf{Description}                                                                                            \\ \hline
\cite{Guo2017}                     & 2017          & Generation from scratch             & SAR                                      & Generate simulated SAR images                                                                                   \\ \hline
\cite{zhan2018_ganforalignment}   & 2018          & Generation from scratch             & \gls{hsi}               & Generate \gls{hsi} to aid later on in classification                                           \\ \hline
\cite{abady2020}                   & 2020          & Generation from scratch             & Multispectral                            & Generate multi-spectral images                                                                                  \\ \hline
\cite{abady2020}                   & 2020          & Semantic Modification               & Multispectral                            & Convert the land cover of multi-spectral images                        \\ \hline
\cite{Ren2021deepfaking}           & 2021          & Semantic Modification               & Multispectral                            & Convert the land cover of multi-spectral images                       \\ \hline
\cite{zhao2021_cycleganrgb}       & 2021          & Semantic Modification               & RGB                                      & Convert the landscape of a source city to that of a target city                        \\ \hline
\end{tabular}
}
\caption{List of techniques discussed in Section \ref{sec:forgeries} and their characteristics.}
%\BTcomm{Great. I like it. If we can short the description of some of them (method no 4,5,6) we can use a larger font that would be better.\LA{ok}}}
\label{tab:forgeries}
\end{table*}

\subsection{Generation of Synthetic Images from Scratch}
\label{ssec:scratch}

In \cite{Guo2017} , the authors use a conventional \gls{gan} architecture to generate synthetic \gls{sar} images.
%to aid in target recognition \CHLA{i.e the capability of detecting a target in an image}0.
The generator is implemented by a deconvolutional network that takes as input observation parameters, that are directly measured from the images, e.g., platform azimuth and target depression angle, and a latent vector characterizing other observation conditions, that is, the target position and environmental factors such as clouds and rain. The discriminator is fed with real samples and generated ones having the same observation parameters.
%Convergence  instability was observed caused by the randomness of the clutter. 
A cluster normalization procedure is implemented to reduce the influence of the clutter, causing convergence instability.
Specifically, segmentation is applied to the images in the training set to separate target and clutter. Then, the images are normalized so that the clutter levels are all the same.
%
%This task requires to know specific observation parameters, such as the platform azimuth and target depression angle. To achieve more realistic results, the authors therefore provided these information to both the generator and the discriminator estimating them from the original samples. The dataset used in their experiments was the \gls{mstar} \cite{mstar_2022}, which offers a \ang{360} target coverage with \ang{15} and \ang{17} depression angles.
% however they cropped patches of size 64 $\times$ 64 from the center of the dataset $128\times128$ patches.
%Unfortunately, due to the randomness of the clutter \CHLA{(the background measurements) \cite{HUFFMAN1992171}} \BTcomm{clutter???\LA{I added here a reference that explains the terms of a target recognition experiment including clutter}} within batches, it was not easy for the \gls{gan} to converge using a standard training procedure. As a matter of fact, the authors noticed that the generator had difficulties in recreating clutters due to the \CHLA{ variability of the clutter} present in the original dataset. The discriminator would then easily tell apart synthesized samples by simply looking at the clutter portion \BTcomm{clutter portion???}  of the samples. To overcome this problem, the authors proposed to use clutter normalization: each training sample is segmented into clutter and target, and then normalized so that the clutter levels of the mini-batches were approximately the same. 
%
Thanks to normalization, the discriminator learns to ignore the clutter and focuses on the target. The generated images are evaluated by applying to them a \gls{cnn} classifier considering 10 selected target categories from the \gls{mstar} dataset. The classification accuracy on the synthetic images is similar to that achieved on pristine images, thus proving the plausibility of the synthetic images.
%They tested the classifier on pristine images and generated images and obtained 95.2\% accuracy and 91.2\% accuracy respectively. 
%However when they trained the classifier on 10 classes of simulated images and tested on pristine images, the classification accuracy was not promising. 
The visual quality of the generated  \gls{sar} images is also assessed and compared with that of images simulated by means of ray tracing  \cite{raytracing_2010}, and that of  real samples, for specific observation parameters.
%in specific a depression angle of \ang{45} and a depression angle of \ang{0}.
The authors show that both simulators (GAN and ray-tracing) are able to predict images close to real ones. 
%\BTcomm{You do not say which architecture is used, how the GAN is trained \LA{basic GAN architecture that is trained on mstar dataset to generate similar images based on depression angle and azimuth}....}\\

%
In \cite{zhan2018_ganforalignment}, the authors propose a semisupervised
algorithm for \gls{hsi} classification exploiting images produced by \gls{gan}s, to overcome the difficulty of gathering a large labeled \gls{hsi} training dataset. 
The authors propose a 1D-\gls{gan}, called hyper spectral \gls{gan} (HSGAN),  inspired by the architecture described in  \cite{radford_2016} (with a difference in the input dimension, set to 1 instead of 2).
The proposed framework enables 
automatic extraction of spectral features needed for HSI classification.
The HSGAN is trained by using unlabeled hyperspectral data, so that it learns to generate hyperspectral samples similar to the pristine samples.
%\MB{MB: this sentence seems incomplete}
Once the GAN is trained, the  discriminator is modified by replacing the last sigmoid layer with a 16 class sigmoid layer and 
then is fine-tuned on a small labeled dataset for hyperspectral classification.
Both HSGAN training and fine-tuning of the discriminator is performed on the Indian Pines dataset,  gathered by \gls{aviris} sensor in June 1992 over the Indian Pines region in northwestern Indiana. 
The size of the original images is 145 $\times$ 145 pixels  with 220 spectral bands. The  noisy and water-absorption bands are filtered out, getting 200 bands, that correspond to the output size of the 1D-\gls{gan}.

% The input of the generator is a 1x100 noise vector and the output is a 1x200 spectral bands vector, while the input of the discriminator is the 1x200 spectral bands vector and returns the probability that the data came from real data rather than the generator. The generator and discriminator were trained on the optimization functions shown previously in Eq. \ref{eqn:genloss} and Eq. \ref{eqn:disloss} respectively. They trained the \gls{gan} for 200 epochs with batch size 128 and \gls{sgd} as the optimizer with \gls{lr} 0.0005 and momentum 0.9.

%
\begin{comment}
\begin{figure}[htbp]
\centering
\includegraphics[clip,width=0.9\columnwidth]{diagrams/p1_zhan_arch.pdf}
\caption{HSGAN Architecture \cite{zhan2018_ganforalignment}}
\label{fig:zhan_arch}
\end{figure}
\end{comment}
%
% Fig. \ref{fig:zhan_real_gan}-(a) shows an example of 128 real spectral lines that the \gls{gan} was trained on. Each line has 200 bands. While Fig. \ref{fig:zhan_real_gan}-(b) displays an example of real spectral waveform. Similarly Fig. \ref{fig:zhan_real_gan}-(c) shows 128 generated spectral lines and Fig. \ref{fig:zhan_real_gan}-(d)  shows a sample generated waveform.
%

\begin{figure}[htb]
\centering

\subfloat[Real Spectral Samples.]{\includegraphics[width=0.4\columnwidth]{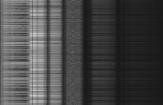}
\label{fig:real_spectral_lines}}
\hfil
\subfloat[Real Spectral Waveform.]{\includegraphics[width=0.4\columnwidth]{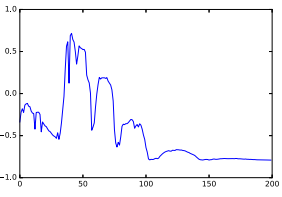}
\label{fig:real_spectral_waveform}}

\subfloat[Generated Spectral Samples.]{\includegraphics[width=0.4\columnwidth]{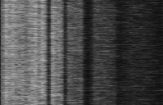}
\label{fig:gen_spectral_lines}}
\hfil
\subfloat[Generated Spectral Waveform.]{\includegraphics[width=0.4\columnwidth]{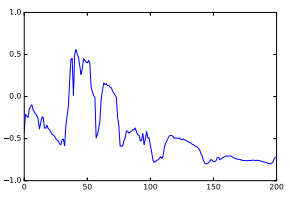}
\label{fig:gen_spectral_waveform}}

\caption{Examples of real and generated \gls{hsi} bands from \cite{zhan2018_ganforalignment}. \ref{fig:real_spectral_lines} and \ref{fig:gen_spectral_lines} show 128 samples of real and generated spectral lines 
%\LA{each line is a sample and each sample have 200 value reach representing a band} \MB{So, a sample is a pixel, correct? If it is correct say it somewhere}, where each line represents a sample with 200 bands (real in the former case, generated  in the latter)
. \ref{fig:real_spectral_waveform} and \ref{fig:gen_spectral_waveform} show a sample of real and generated spectral waveforms respectively, with 200 values each.}
\label{fig:zan_real_gan}
\end{figure}

Fig. \ref{fig:zan_real_gan} shows 128 examples of real and synthetic spectral bands, where each line corresponds to the values assumed by one pixel on the 200 bands. The figure also shows a real and a synthetic waveform.
%\MB{As mentioned in the caption, the meaning of mages shown in the figure is not clear}
The authors have shown experimentally the superior performance of the HSI classifier trained on synthetic images with respect to  state-of-the-art methods. 
They also assessed the impact of the GAN training dataset size on the classification performance, proving that the size of the HSGAN has a noticeable impact on the classification  accuracy; the more data the HSGAN is trained on, the more accurate the classifier is.
%Table  \ref{table:zhan_results} shows the overall accuracy of the classification when the HSGAN is trained on different percentages of the Indiana Pines dataset. Fine-tuning of the discriminator is performed  on  10\% of the labeled  Indiana Pines dataset.
%\BTcomm{I think that the table can be removed since it does not add much. The textual description is enough. I left it for the moment.\LA{ok,removed}}

%
%Table. \ref{table:zhan_results}, shows the results of their experiments.
\begin{comment}

\begin{table}[htbp]
\noindent
\begin{center}
\begin{tabular}
{||ccccc||}
 \hline
 Dataset percentage & $100\%$ & $50\%$ & $25\%$ & $10\%$\\
 \hline
\gls{oa} &83.5&81.2&79.1&78.6\\ \hline
\end{tabular}
\end{center}
\caption{Effect of \gls{gan} training dataset size on the classification accuracy.}
\label{table:zhan_results}
\end{table}
\end{comment}

In  \cite{abady2020}, the authors use a \gls{progan} architecture \cite{karras_2017} to generate 13 bands Sentinel-2 level-1C images of $256\times256$ resolution. For training, they have used all 180k samples of the SEN12MS dataset. Similar to \cite{karras_2017}, a \gls{wgan-gp} loss function is used.

\subsection{Semantic Modifications}
\label{ssec:semantic}

\begin{figure}
\begin{minipage}[b]{1\linewidth}
  \centering
  \centerline{\includegraphics[width = 1\linewidth]{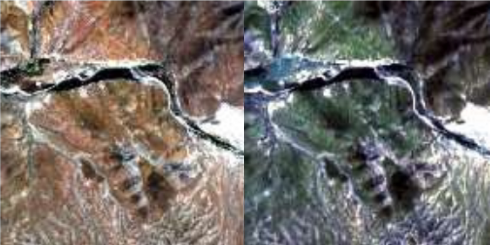}}
  \centerline{(a) Image translation from barren to vegetation domain.}\medskip
\end{minipage}\hfill
\begin{minipage}[b]{1\linewidth}
  \centering
  \centerline{\includegraphics[width = 1\linewidth]{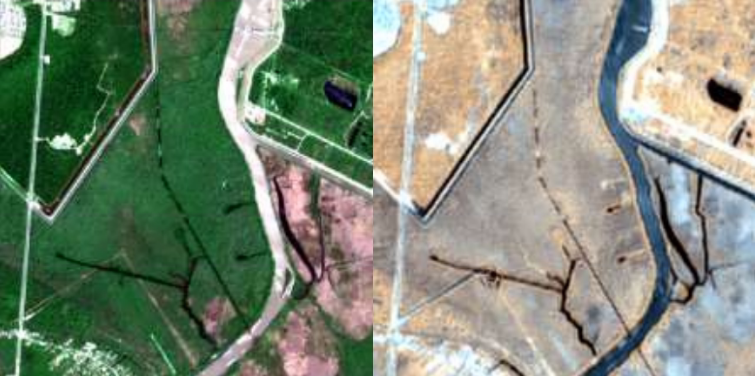}}
  \centerline{(b) Image translation from vegetation to barren domain.}\medskip
\end{minipage}
\caption{Land-cover image translation examples taken from \cite{abady2020}.}
\label{fig:abady_style_real_gan_samples}
\end{figure}

In \cite{abady2020} mentioned before, the authors propose an image-to-image translation solution tailored to the remote sensing scenario. They trained a network for land cover transfer, i.e., a network able to change the content of images to move them from one land-cover class to another. In particular, the paper focuses on transferring a sample from the vegetation class into the barren class and vice versa. They rely on the \gls{nicegan}~\cite{chen2020} architecture to perform  unpaired style transfer\footnote{For the land-cover transfer task, an unpaired dataset has to be used (given an image from a source domain, the corresponding image in the target domain is not available).}. The model was trained on 4 bands, i.e., RGB and NIR (10 meters bands) images of size $480\times480$  cropped from Sentinel-2 level-1C products, with no cloud coverage. For the vegetation domain, 20k images were gathered from Congo, El Salvador, Montenegro, Gabon and Guyana, while for the barren domain 20k images were gathered from Western Sahara. Samples were split into training and testing.  Specifically, 18K images  from each domain are used for training, while the remaining 2K are left for testing.
%Adam Optimizer was used with a \gls{lr} of $10^-3$, $\beta_1$ of 0.5 and $\beta_2$ of 0.99, and the networks were trained for 15 epochs.
Fig. \ref{fig:abady_style_real_gan_samples} shows  an example of real and GAN-transferred images for each transfer direction. The authors also verify that the expected correlation between the bands is preserved in the generated images, by looking at the spectral view of pixels belonging to different land cover classes for both real and  generated images.

A similar task is pursued by Ren et al in \cite{Ren2021deepfaking}. In this work, the authors exploit a \gls{cyclegan} architecture to translate 10 bands of Sentinel-2 level-1C image, namely the 10m and 20m bands,
from drought to vegetation and vice versa. \\~\\

\begin{figure}[htbp]
\centering
\includegraphics[clip,width=0.9\columnwidth]{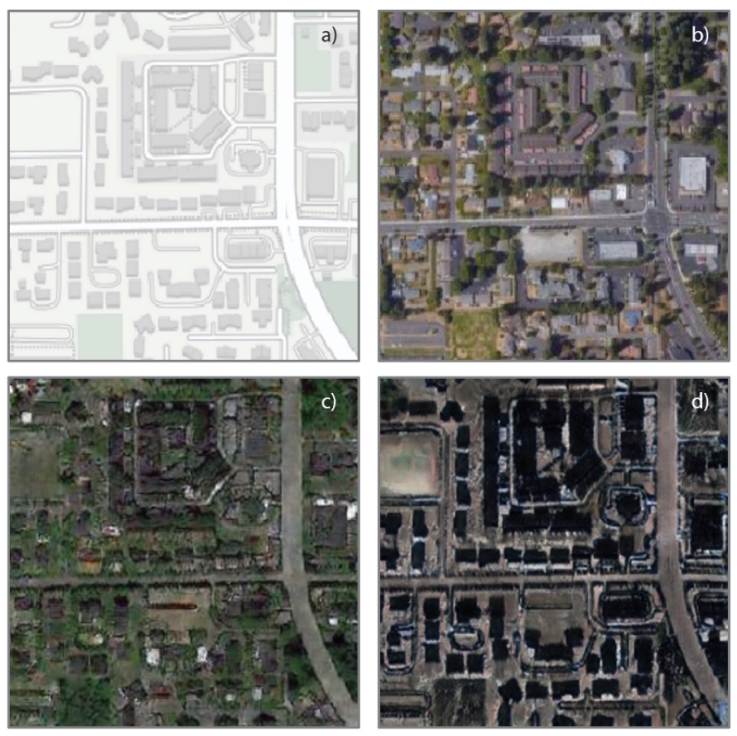}
\caption{An example of generated satellite images  using the basemap image   in (a) from Tacoma, with the landscape features of Seattle (c) and Beijing (d). (b) shows the ground truth satellite image that belongs to the basemap in (a) \cite{zhao2021_cycleganrgb}.}
\label{fig:p4_Zhao_imgsamples}
\end{figure}

In \cite{zhao2021_cycleganrgb}, the \gls{cyclegan} architecture is used for a different semantic modification:
the creation of synthetic images having the urban structure of a given city (i.e., Tacoma in Washington, U.S.) but with the landscape features of another city (i.e., Seattle in Washington, U.S. and Beijing, China).
To achieve this task, they train a \gls{cyclegan} model on a given city to generate an image with the  landscape features of this specific city, starting from an input base-map with the desired city structure.
For the map domain, the authors use the cartoDB basemaps,
%\BTcomm{I would put a ref\LA{the original paper didn't add a reference, I do have an assumption about the correct software but I rather not reference it as it might be wrong}}, 
which provides basic urban structural information.
As for the satellite imagery domain, they rely on Google Earth's satellite imagery. They use the QTile plugin in QGIS open source software \cite{qgis} to collect their datasets for both domains.
% and three cities at zoom level 16.
All datasets have a resolution of $256\times256$. 1196 pairs of images are collected, that is satellite images and their corresponding basemaps,  for Seattle, 1122 pairs for Beijing and 758 pairs for Tacoma.
%\BTcomm{Why you are speaking about pairs? this is unpaired right? \LA{it is unpaired but when they collected the datasets they obtained the satellite images and their respective basemaps hence they called them pairs so I am using same terminology as the paper}}. 
%The models were trained for 200 epochs.
Fig. \ref{fig:p4_Zhao_imgsamples} shows an example of  a satellite image generated from a basemap from Tacoma, showing the landscape features of Seattle and Beijing.

\section{Beyond Forgeries}
\label{sec:bforgeries}

In this section, we overview additional methods for the generation and manipulation of satellite imagery, considering techniques proposed to edit an image content without a necessarily malevolent goal. This is the case, for example, of image enhancement techniques. Despite the fact that these methods are not meant to be harmful or used in a deceptive way, they still undermine image integrity to some extent. For instance, a colorized image obtained synthetically from a gray scale one could be considered as altered from the data integrity point of view.

\begin{figure}[htbp]
\centering
\includegraphics[clip,width=0.99\columnwidth]{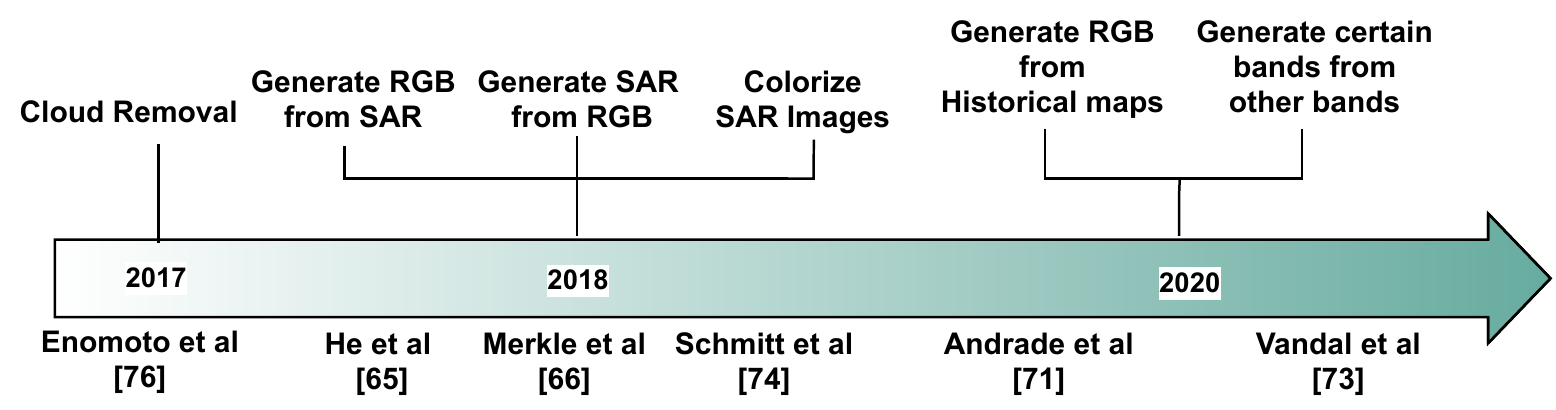}
\caption{Timeline highlighting the main works related to synthetic image generation that can not be categorized as malevolent forgeries.}
\label{fig:enhtimeline}
\end{figure}

In the following, we consider two types of generation and manipulation: i) generation of satellite images of a given type from a different type of data, e.g., the generation of an \gls{eo} image starting from a \gls{sar} image and vice versa (addressed in Section \ref{ssec:type});  and ii) modifications aiming at quality enhancement, such as colorization and cloud removal (see Section \ref{ssec:quality}).
Table \ref{tab:bforgeries} summarizes all the generation methods considered in
%Section \ref{sec:bforgeries} of this overview,
this section, categorized according to the proposed classification. Figure \ref{fig:enhtimeline} shows the timeline of the works described in this section.

\begin{table*}[]
\centering
\resizebox{\textwidth}{!}{%
\begin{tabular}{|l|l|l|l|p{0.4\linewidth}|}
\hline
\textbf{Reference}                                  & \textbf{Year} & \textbf{Task}                       & \textbf{Data type}                       & \textbf{Description}                                                                                            \\ \hline
\cite{he2018_rgb}                 & 2018          & Datatype transfer / Inter-modality  & RGB-\gls{sar}           & Generate optical images from \gls{sar} input or fused optical-\gls{sar} input \\ \hline
\cite{Merkle2018}                  & 2018          & Datatype transfer / Inter-modality  & \gls{sar}-RGB           & Simulate \gls{sar} from optical images                                                         \\ \hline
\cite{enomoto2018gan}              & 2018          & Datatype transfer / Inter-modality  & RGB-\gls{sar}           & Simulate optical images from \gls{sar} images                                                  \\ \hline
\cite{liu2018}                     & 2018          & Datatype transfer / Inter-modality  & RGB-\gls{sar}           & Simulate optical images from \gls{sar} images and vice versa                                   \\ \hline
\cite{reyes2019}                   & 2019          & Datatype transfer / Inter-modality  & RGB-\gls{sar}           & Simulate optical images from \gls{sar} images                                                  \\ \hline
\cite{bermudez2019synthesis}       & 2019          & Datatype transfer / Inter-modality  & RGB-\gls{sar}           & Simulate optical images from \gls{sar} images                                                  \\ \hline
\cite{andrade2020}                 & 2020          & Datatype transfer / Inter-modality  & RGB                                      & Generate optical images from historical maps                                                                    \\ \hline
\cite{yuan2020}                    & 2020          & Datatype transfer / Intra-modality  & Multispectral                            & Generate NIR images from RGB images                                                                                           \\ \hline
\cite{vandal2020}                  & 2020          & Datatype transfer / Intra-modality  & Multispectral                            & Generate certain bands using other bands as input                                                              \\ \hline
\cite{schmitt2018colorizing}       & 2018          & Quality Improvement / Colorization  & RGB-\gls{sar}           & Generate colorized \gls{sar} images from \gls{sar}-optical fused image        \\ \hline
\cite{tasar2019}                   & 2019          & Quality Improvement / Colorization  & RGB                                      & Adapt color distribution of a testing dataset to match that of a classifier training dataset             
\\ \hline
\cite{enomoto2017filmy}            & 2017          & Qaulity Improvement / Cloud Removal & Multispectral                            & Remove clouds from RGB images using NIR band as auxiliary information                                           \\ \hline
\cite{singh2018}                   & 2018          & Quality Improvement / Cloud Removal & RGB                                      & Remove clouds from RGB images                                                                                   \\ \hline
\cite{grohnfeldt2018_conditional} & 2018          & Quality Improvement / Cloud Removal & Multispectral-\gls{sar} & Remove thick clouds from multispectral images                                                                   \\ \hline
\cite{zotov2019}                   & 2019          & Quality Improvement / Cloud Removal & RGB                                      & Remove clouds from RGB images                                                                                   \\ \hline
\cite{ebel2020}                    & 2020          & Quality Improvement / Cloud Removal & RGB-\gls{sar}           & Remove clouds from RGB images                                                                                   \\ \hline
\cite{gao2020cloud}                & 2020          & Quality Improvement / Cloud Removal & RGB-\gls{sar}           & Remove clouds from RGB images                                                                                   \\ \hline
\cite{sarukkai2020cloud}           & 2020          & Quality Improvement / Cloud Removal & Multispectral                            & Remove clouds using temporal data of RGB and NIR bands                                                          \\ \hline
\cite{wen2021}                     & 2021          & Quality Improvement / Cloud Removal & RGB                                      & Remove clouds from RGB images                                                                                   \\ \hline
\end{tabular}
}
\caption{List of methods discussed in Section \ref{sec:bforgeries} and their characteristics.}
%\BTcomm{Great. I like it. If we can short the description of some of them (method no 4,5,6) we can use a larger font that would be better.\LA{ok}}}
\label{tab:bforgeries}
\end{table*}

\subsection{Datatype Transfer}
\label{ssec:type}
One of the main applications of remote sensing data is Earth monitoring and change analysis. For these tasks, both \gls{eo} (optical) imagery and \gls{sar} are usually exploited, considering data captured at different times. Being able to generate one type of data or modality from the other facilitates these tasks since only a type of data would need to be acquired.  We refer to  this kind of image translation with the term  {\em inter-modality} datatype transfer. As described in Section \ref{ssec:modalitytrans}, several methods have been proposed in this category.

Another remote sensing application for datatype transfer comes from land cover classification and object detection. Typically, different spectral bands are used for these tasks. Instead of incurring in the costs of acquiring all the bands,  only some of them are acquired while the others are synthetized automatically. Similarly, we can generate  missing spectral bands relying on  existing spectral bands. We refer to this kind of transfer as {\em intra-modality} datatype  transfer (Section \ref{ssec:intramodalitytrans}).

\subsubsection{Inter-modality Transfer}
\label{ssec:modalitytrans}
the prediction of optical images using \gls{sar} images is first considered in \cite{he2018_rgb},
%

%\MB{This method is given considerably more attention than the others, unbalancing the content of the section.}

Specifically, \cite{he2018_rgb}  addresses the problem of generating optical images that represent a prediction of the foreseen land-cover changes using different combinations of remote sensing data as input.
Two different architectures are proposed. The best performing method resorts to a pix2pix architecture, that adopts a
ResNet-like architecture \cite{He2016resnet}  for the  generator, and a patchGAN \cite{isola_2016} with 5 layers for the discriminator.
The patchGAN classifies the patches of an input image, providing a score matrix at the output; the final score on the whole image is taken by the discriminator by averaging all the outputs.
%to get the final result whether the image is fake or real.

Let T1 be a given acquisition time or period, and T2 the target period (corresponding to a later time - the images are considered of the same period if the collection dates' difference is less than 5 days).
Two combinations of the input samples are considered in \cite{he2018_rgb} and their impact on the networks' ability to predict the optical samples assessed.
Specifically, the authors consider providing as input: i)  only  \gls{sar} images, from both T1 and T2; ii) both \gls{sar} and optical images from T1 and \gls{sar} images from T2.
%
%For both networks, the authors experimented with two input configurations. The first one consisted in providing  only Sentinel-1 \gls{sar} patches \BTcomm{from both T1 and T2?\LA{from either}}, while the second one utilized the previous pairs of \gls{sar} and optical tiles from period T1 in addition to the the \gls{sar} tile of the target period T2. 
% For clarity, the CNN network that was trained only on \gls{sar} images as input will be referred to as \textbf{CNN} while the one that was trained on the previous period paired images and current \gls{sar} image will be referred to as \textbf{MTCNN} and similarly for the trained \gls{cgan} will be \textbf{cGAN} and \textbf{MTcGAN}.
In the following, the two networks trained with the two input combinations will be denoted as \gls{cgan} and MTcGAN, respectively.

The data for training and testing are gathered from the  Copernicus hub \cite{copernicus}. Three different regions are considered for the experiments: Iraq, Jianghan, and Xiangyang. 
For each area, 4 images are downloaded from the Copernicus hub, that is two Sentinel-1 images (i.e., \gls{sar}) and two Sentinel-2 images (i.e., \gls{eo}), from T1 and T2.
\gls{sar} images were pre-processed using \gls{snap}.
%with calibration, despeckling, range doppler terrain correction and two bands of \gls{vv} and \gls{vh} intensities with pixel spacing of 10m. 
For Sentinel-2 products, only  4 bands are  considered, i.e., RGB and NIR, with a \gls{gsd} of \SI{10}{m}. The \gls{sar} and optical images are co-registered by reprojection in order to provide information on the same geographical area. Then the images were divided into $256\times256$ patches and split into training and test sets.

%The quality of the predicted optical images is evaluated by measuring the \gls{psnr}, the \gls{ssim}, and the \gls{msa}  \cite{KRUSE1993}, which is computed between the prediction and the ground truth (the lower the better). 
%
%The \gls{psnr} and \gls{ssim} are computed for each of the 4 bands of the predicted optical image and the reference image from period T2, and then the average value is considered.
%
%Table \ref{table:he_case1results} shows the metrics for the two different input configurations, when data were collected from Iraq, for which the terrain considerably changed between the two selected time periods due to an earthquake, thus making the prediction task challenging. 
%
With the assessment made in \cite{he2018_rgb}, it  is argued that using the optical data %from the previous period 
as additional input is beneficial, leading to an improvement of the networks' capability of predicting the optical samples corresponding to T2. 

Inspired by \cite{he2018_rgb}, considerable research has been dedicated to the generation of optical images from \gls{sar} images and vice versa, for a different or also the same acquisition time. 
%\BTcomm{It seems that most of the approaches that you present in the following just care about datatype translation at the same time instance, correct? so the previous one seem from general.\LA{yes} }
The most relevant approaches are described in the following.

The opposite transfer with respect to that considered in \cite{he2018_rgb}, that is, the generation of \gls{sar} images from optical images, 
is addressed in \cite{Merkle2018}, 
with the goal of improving the matching between optical and \gls{sar} images,
to improve the geolocation accuracy of optical satellite images.
%\CHLA{in order to extract \gls{gcp} for improved geolocation accuracy}. 
%
%The authors first selected suitable matching areas based on predefined conditions to increase the reliability and accuracy of matching points. Then, they 
The authors use a pix2pix architecture to generate \gls{sar} patches from the corresponding optical image using three different losses: the original \gls{gan}  loss, a variation of the adversarial loss adopting the mean square error in place of the log likelihood \cite{mao2017_lsgan}, and finally the conditional Wasserstein loss \cite{arjovsky2017_cwgan}. Training is performed  on 201 $\times$ 201 paired patches from TerraSAR-X (\gls{sar}) and PRISM (optical), gathered all over Europe.

 Finally, they assess the matching between the \gls{sar} images and the generated \gls{sar} images
using three \gls{sota} image-matching techniques: \gls{ncc}, \gls{sift} and \gls{brisk}. All metrics prove that the generated \gls{sar} images were beneficial to improve the matching. 
%\BTcomm{Unclear which is the target to be improved matching. See my previous comment.\LA{improving matching between optical and sar images is a research topic that aims to obtain complementary information by fusing the optical images with the the sar images in this paper they try to extract ground control point in an accurate way for improved geolocation accuracy, i added additional clarifications }} 
%
%

Other  methods have been proposed addressing similar tasks.
In  \cite{enomoto2018gan}, the pix2pix architecture is  used to address the  transfer from SAR images to optical images.
A \gls{cyclegan} architecture, instead of a pix2pix, is used in \cite{reyes2019} for the same task of translating  \gls{sar} into optical images. 
%
\begin{comment}
{\em Similarly, in \wtitle{Exploiting GAN-Based SAR to Optical Image Transcoding for Improved Classification via Deep Learning \cite{ley_2018}}, the same task was carried out to later on pretrain a land cover classifier on the generator's learned features, aiming at improving the classification accuracy when the labeled data is limited.}
\BTcomm{I could not undersatnd the above method. If tou think is relevant and you can explain better we can add. Otherwise I will remove it.\LA{it is relevant, the trained a GAN to gene}}
\end{comment}
%
In \cite{liu2018}, a \gls{cyclegan} architecture is proposed to  perform both translation, from optical to  \gls{sar}  images, and vice versa.
%translate bi-directionally optical to \gls{sar} images.
%
% commenting this out because it is only on arxiv with 35 citations
%In \wtitle{Generating High Quality Visible Images from \gls{sar} Images Using CNNs \cite{Wang2018}}, the authors generated optical images from \gls{sar} images using a cascade of two \gls{cnn} networks, where one was used to despeckle the image and the other was used to colorize it.
%
Finally, in \cite{bermudez2019synthesis}, the authors 
%aimed to mitigate the impact of heavy clouds on optical images. Hence, by using a pix2pix architecture conditioned on current matching of \gls{sar} image and \gls{sar}-optical pairs of the same area, but from a different date, they replaced patches containing heavy clouds with generated optical images.
%
address the generation of optical images from  \gls{sar} images at different times, to mitigate the impact of heavy clouds on optical images (i.e., generating optical images that do not contain clouds).
The same approach introduced in \cite{he2018_rgb} is used, with the difference that a pix2pix archiutecture is considered instead of the \gls{cgan} architecture that is adopted in \cite{he2018_rgb}.

A last application of inter-modality transfer is the generation of optical images starting from maps or auxiliary raster data. In \cite{andrade2020}, the authors generate satellite-like imagery from historical maps using a pix2pix network.
%\BTcomm{I rephrased below. As far as I understood at the end, the historical maps are only used for testing and not for trianing (this was not clear. I am guessing). Check that the rephrased description is ok.\LA{it is ok}}
%
%As a first step, they applied segmentation to the historical maps 
%{\em by considering the context as the localization of the textures. The context was approximated using features extracted from small towns around the region of analysis} 
%\BTcomm{I did not fully understand this (what you mean with 'by considering the context as the localization of the textures'?\LA{no this is important, it cant be removed, they didnt have actually the pair so they approximated how the area would have look like manually so the texture is estimated based on the location and context}) but I thnik it can be removed.\LA{ok commented it out, though it points out the manual segmentation based on assumption for historical maps}}.
They train the pix2pix architecture on satellite images extracted from Google  and the corresponding segmented images, providing  the optical image and the segmented image as  input images from the two domains. %They trained two models, one on the 
%%current location of the historical map that was urbanized
%urbanized areas and the second on rural and natural nearby areas. 
During testing, segmentation is applied to the historical maps, then the generator is fed with the segmented image to get the optical image representing the area (what the area might have looked like).
They train two models, one on urban areas and the second on rural and natural areas. 
They argue that merging the output of the two models based on the label (that is, taking the urban labeled pixels
from the model trained on urban area, and the rural labels from the model trained on rural areas) yields a better representation of the area. 
%\CHLA{The merging was done based on label where the urban labeled pixels were taken from the model trained on urban area while the rural labeled pixels where taken from model trained on rural areas.} 
%\BTcomm{How this merging is performed? is it blending? how can they say that it gives a 'better representation'? I mean, how do they judge the image they get (they do not have the GT)?\LA{since they don't have the ground truth, they assumed based on visual observations }}

\begin{figure}[htb]
\centering
\includegraphics[clip,width=0.9\columnwidth]{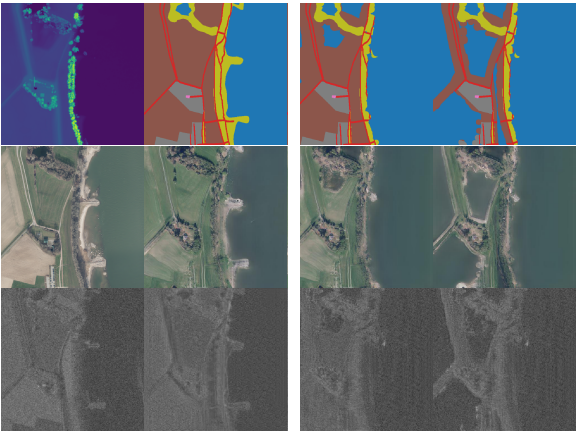}
\caption{An example of generated datatype with semantic modification \cite{baier2022synthesis}. The first column shows a real  DEM, optical and SAR image.  The second column shows the original land cover map with the corresponding generated optical and SAR image. The third and fourth column show the modified land cover with threshold set to height 32m and 33m respectively with the corresponding generated optical and SAR images.}
\label{fig:manipulatedbaier}
\end{figure}

In \cite{baier2022synthesis}, the authors generate optical RGB and \gls{sar} imagery from land  cover segmentation maps and auxiliary raster data, To this purpose, they use a variation of the pix2pix architecture where all the normalization layers are replaced by \gls{spade} layers \cite{spade_2019}.
%, in this way generating images from only semantic maps.
The auxiliary data is used as an input to the generator while the land cover is  passed as an input to all \gls{spade} layers. For training and testing, two datasets are used: i) GeoNRW \cite{geonrw_2020}, containing aerial photographs, terraSAR-X, DEMs and land cover with 10 classes; ii) DFC2020 \cite{DFC_2020}, containing Sentinel-1, Sentinel-2 and land cover data with 10 classes.
For the GeoNRW, SAR images or optical images is generated from land cover maps and \gls{dem} \cite{dem_2021} used as auxiliary data, while for the DFC2020 (that lacks  \gls{dem} data), \gls{sar} and optical imagery are generated using either solely land cover maps or a combination of land cover with optical (for \gls{sar} generation) or \gls{sar} imagery (for optical generation)  as auxiliary input. %\BTcomm{It is unclear how the network and the SPADE layers are fed in the second case, with no raster data \LA{SAR or optical were considered the extra input instead of dem}. You said before that the raster data was fed as input while the land cover was passed as an input to all \gls{spade}. What they do in the second case, when they do not have the raster data?}. 
To judge the effectiveness of the generation in terms of land cover coverage, the authors analyze the land cover segmentation maps obtained by a U-Net trained on real data 
and compare the  land cover map obtained from the generated images with the ground truth map, that is used for the generation.
%to generate the images in terms of \gls{iou}
%
%\CHLA{and tested on the generated images and compared the obtained land cover map with the land cover map that was used to generate the images in terms of \gls{iou}} \BTcomm{to me it is unclear how they judge the quality of the generated images using Unet. What is the GT here?\LA{here they used unet for land cover segmentation on the generated sar or optical data and compared the obtained segmentation mask with the land cover segmentation mask that was used as an input to generate the optical or sar images, I paraphrased to make it more clear }}. 
The results show that the generated images are comparable to the real images in term of land cover maps.
%\BTcomm{I added 'in terms of land cover maps'. It seems this is  what they judge using the UNet and NOT actually the quality of the genereated images (this is what you say above). If this is the case, I suggest rephrasing 'To evaluate the quality of the generated images' above with  'To judge the effectiveness of the generation in terms of land cover coverage' or somthing like that \LA{ok} }. 
Moreover, the authors show that this type of image-to-image translation can also be used to modify the semantic content of an  image.
%by slightly modifying the inputs of the generator. 
For instance, by applying a threshold to a certain height in the \gls{dem} image,  and modifying the pixels of the land cover map with a value in the DEM image below that height and labeling them as water, we can get a modified image with a larger area covered by water. An example of this semantic modification obtained with the transfer network in \cite{baier2022synthesis} is shown in Fig. \ref{fig:manipulatedbaier}.
\subsubsection{Intra-modality Transfer}
\label{ssec:intramodalitytrans}
A different kind of datatype transfer regards the generation of a subset of \gls{eo} bands starting from a different subset of  bands.
In \cite{yuan2020}, the authors resort to a  pix2pix network for generating the \gls{nir} band of Sentinel-2 samples using the RGB bands as input. The model is trained using a subset of SEN12MS dataset. 
%{\em The quality of the generated \gls{nir} band is assessed by computing the \gls{mae}, \gls{mape}, and \gls{ssim} between the generated bands and the ground truth.} 
%\BTcomm{I would remove these considerations on the metric, unless you also report the results and then it make sense introducing them. You do not speak about performance in the other cases. We should do the same.\LA{ok, commented them out}}
% where the best achieved scores were $9.67\times10^{-3}$, 4.73\%, and 93.63\% respectively.

In \cite{vandal2020}, the authors synthesize multiple spectral bands from other bands, using unsupervised image-to-image transfer \cite{liu2017_unsupvaegan}.
%\BTcomm{I had a quick read to this paper and rewritten the following description. Please check.\LA{ok}}
More specifically, they rely on a \gls{vae}-\gls{gan}, that is a combination between a \gls{vae} and a \gls{gan}, proposed in \cite{Larsen_2016}, where a discriminator is used to learn to differentiate between \gls{vae} output and real samples, and is used in an adversarial fashion to improve the reconstruction error of the \gls{vae}.
%\CHLA{which is a combination between a \gls{vae} and \gls{gan} that is proposed in \cite{Larsen_2016} where the decoder is replaced by a generator and a discriminator networks in order to maintain the quality of a \gls{gan} and the diversity of a \gls{vae}.} \BTcomm{You should say more on this \gls{vae}-\gls{gan}, otherwise, it is impossible to understand what the method does. }
To improve the reconstruction, the authors introduce  skip connections in the network and a shared spectral reconstruction
loss, encouraging the decoder to reconstruct identical spectral wavelengths
with similar distributions while still synthesizing dissimilar bands. The shared reconstruction loss exploits the availability of shared spectral bands from different satellites.
Data for training and testing are gathered from 3 geostationary satellites: GOES-16, GOES-17, and Himawari-8, with 16 spectral bands, 15 of which overlap, with similar information content
(GOES-16 and GOES-17 include two visible -blue, red-, four near-infrared  -including cirrus-, and ten thermal infrared 
bands. H8 captures three visible -blue, green, red-, three near-infrared -missing cirrus band -, and the same
ten thermal infrared bands as GOES-16 and GOES-17). 
The authors first evaluate the ability of the network to generate an individual band from the other 15 bands acquired by the same satellite and the full set of bands
from the other two satellites.
This approach is applied to GOES-16, hence  each model takes 15 bands of GOES-16 and 16 of  GOES-17 and Himawari-8.
The improvement in the reconstruction error brought by the modifications introduced by the authors, with respect to the use of a standard VAE, is proven experimentally.
The reconstruction \gls{mae} obtained with the proposed solution is then compared against the use of cross sensor and UNIT \cite{unit_2017} as a baseline 
%\BTcomm{What is such cross sensor and UNIT baseline? is it one or two different approaches? did they perform synthesis ? if yes, they whould also be overviewed \LA{UNIT does perform synthesis but wasn't applied on remote sensing what they mean by using it as a baseline is they used it to synthesize and compared to their proposed architecture that is based on vae gan as for cross sensor it means that they take a band from another sensor as it is without considering the source}},
proving the superior performance of the proposed method in terms of \gls{mae} and precision. 
%\BTcomm{which are the conclusion? is the approach superior in something?\LA{theirs}}
%\BTcomm{what you mean with 'precision'?}
%
They also assess the capability of the network to generate multiple missing bands from satellites with a limited number
of spectral bands (older generation satellites often have fewer channels hence being able to generate additional frequency bands can be very useful).
Models are trained on GOES-16, removing bands one by one until just one band was left, and reconstructing the missing bands.
All 16 GOES-17 and Himawari-8 bands were kept.
They observe that the reconstruction \gls{mae} decreases  monotonically (approximately)
 as more bands are given as inputs.
These results show that few bands (3-4) are needed to synthesize images with an acceptable \gls{mae}.

\subsection{Quality Improvement}
\label{ssec:quality}

Editing remote sensing images using deep learning is not confined to changes in the semantic content or in the datatype. Methods have also been developed to change the properties of the image itself, e.g., performing colorization or reduction of cloud coverage. 

\subsubsection{Colorization}
% Due to the fact that \gls{sar} images can obtain physical qualities of a place without regard to the weather or time conditions, make them valuable \gls{eo} data. But because of the difficulty in interpretability, a lot of research is dedicated to colorizing \gls{sar} images. The most prominent is \gls{sar}-optical fusion. Nonetheless, fusion still requires to have matching optical images collected around the same time as the \gls{sar} image.
Being unaffected by weather and daylight conditions, \gls{sar} images are a valuable asset in many applications. However, with respect to \gls{eo} imagery, \gls{sar} images are  more difficult to  interpret visually, as the frequency range they capture  does not cover the visible part of the spectrum. For this reason, an explored research topic in the remote sensing community is the colorization of \gls{sar} images, i.e., the fusion of information from \gls{sar} and \gls{eo} imagery,  through the use of matching image pairs.

\begin{figure}[htb]
\centering
\includegraphics[clip,width=0.9\columnwidth]{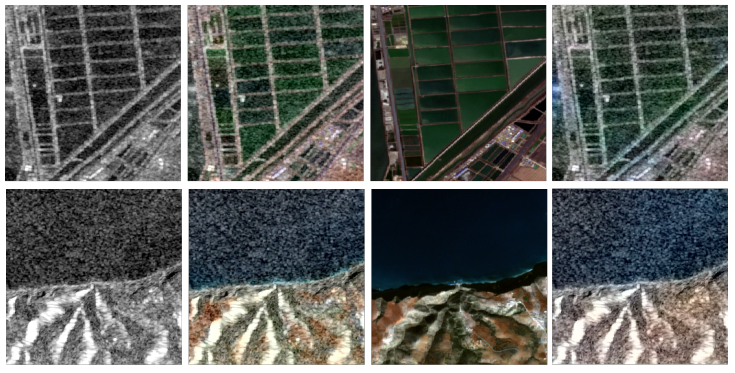}
\caption{Examples of SAR image (left column), target fused optical-sar image (second column), corresponding optical image (third column), and generated colored example (right column) \cite{schmitt2018colorizing}}
\label{fig:schmittcolorizing}
\end{figure}

Recently, 
%as an alternative to this technique, 
the use of \gls{dnn} for colorizing \gls{sar} images has been investigated. In \cite{schmitt2018colorizing}, the authors propose to colorize \gls{sar} images by means of an architecture derived from \cite{deshpande2017_imgclr}, which encompasses a \gls{vae} along with a \gls{mdn} \cite{bishop94_mixturedensity}.
% The first step was to train the \gls{vae} to learn the low dimensional embedding of the colored images. Since colored \gls{sar} images are not available, the authors employed \gls{sar}-optical fusion by color space transform \cite{pauli_77} to obtain colored \gls{sar} images for training.
They first train the \gls{vae} to learn low dimensional embeddings of images obtained by fusing \gls{sar} and optical images through color space transform \cite{pauli_77}.
Then, they train \gls{mdn} to learn 
%a multi-modal conditional distribution describing 
the relationship between the original grey-scale \gls{sar} image and the  low dimensional latent variable embedding of the corresponding \gls{sar}-optical fused image generated by the \gls{vae}. During testing, 
% the trained \gls{mdn} is used, to obtain a sample of the multi-modal colorization hypothesis conditioned by the \gls{sar} image used as input
given a \gls{sar} image as input, the low dimensional latent variable embedding of the \gls{sar}-optical fused image is obtained from the \gls{mdn}, and is then forwarded to the decoder of the \gls{vae} to get the final colorized \gls{sar} image.
Fig. \ref{fig:schmittcolorizing} shows a couple of examples of the \gls{sar} input image, the desired optical-SAR fused image, the optical image and the synthetic colorized \gls{sar} image.

%\MB{MB: Can we show 1 or 2 examples?}.
% They used co-registered Sentinel-1 and Sentinel-2 samples for training and testing.
%

Another application of image colorization considers the correction of \gls{eo} images to mitigate the effect of acquisition conditions (e.g., atmospheric effects).
%
%
%\BTcomm{what does this mean exactly?The acquisition conditions affect what exactly?\LA{the spectral distribution of the image is changed based on the atmosphere and the sun alignment and the time of acquisition, the angle of acquisition causes shadows }}. %
This problem  affects the generalization capability of deep learning tools based on the semantic content, given that
training and testing datasets might present different distributions of the  spectral bands values because of different acquisition conditions. For instance, very often there exists a large difference between spectral bands of satellite images collected from different geographic locations.
%As a matter of fact, training and testing datasets might present different distributions of their spectra values.
%For this reason, in
To overcome the impact of spectral distribution diversity due to acquisition  conditions, in \cite{tasar2019}, the authors propose a new \gls{gan}, named ColorMapGAN, able to generate images semantically identical to the images in the training dataset, but with spectral distribution similar to the test dataset. To preserve the exact semantics, the authors avoided the use of convolutional and pooling layers, that are part of traditional \gls{cnn}-based \gls{gan} architectures.
%were not adopted.
%
Instead, ColorMapGAN simply transforms
the colors of the training images into the colors of the test
images without introducing any structural changes on the objects of
the training images.
%
%\BTcomm{Do you mean that this architecture avoid the use of conv and pool layers? so, which kind of layers is used?\LA{they train it only to apply shift and scale on the channels so the learnable parameters are the shift and scale matrices, I added clarifications}}\CHLA{instead they only train shifting and scaling parameters}.
% Basically, a U-net classifier is trained on a training dataset with ground truth labels then fine tuned on a training dataset adapted to the test dataset using colorMapGAN. They proved that their proposal is superior in comparison to the \gls{sota} domain adaptation for semantic segmentation.
The colorMapGAN is then  used to finetune semantic classifiers by generating samples whose spectral distribution targets that of a desired testing dataset.

\subsubsection{Cloud Removal}
% Besides changing the spectral distribution of an image, another modification is to remove the clouds from an image.
Another common quality improvement to reduce the impact of atmospheric conditions in the acquired images is the removal of clouds from \gls{eo} images.
This problem has been addressed in many works. 

\begin{figure}[htbp]
\centering
\includegraphics[clip,width=0.99\columnwidth]{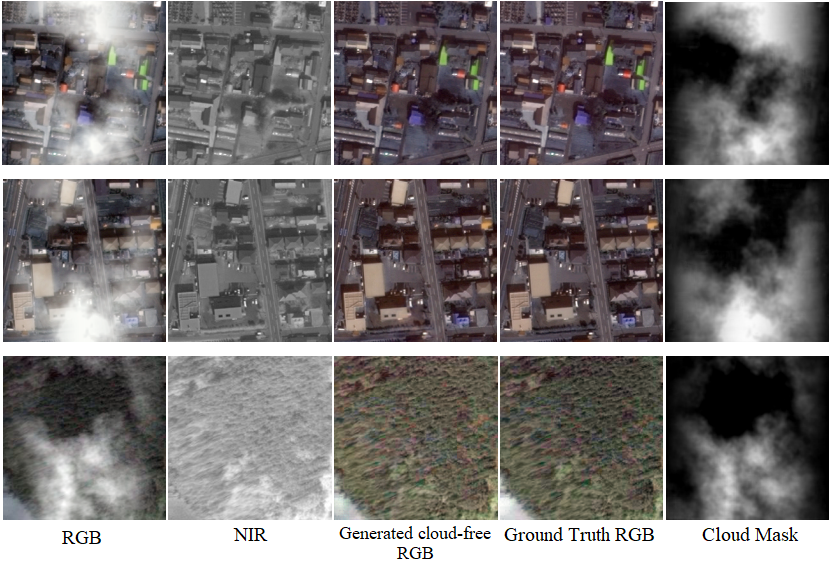}
\caption{Some examples obtained by the cloud removal method in \cite{enomoto2017filmy}. From left to right:  input cloudy RGB images, the corresponding NIR band,  the generated cloud-free RGB images, the ground truth RGB images with no clouds, and finally the generated cloud mask.}
%\BTcomm{In the figure, 'Cloud-free RGB' and 'Ground truth' is not explanatory. The former should be the 'generated cloud-free RGB', while the latter should be the 'GT cloud-free RGB'.}
\label{fig:enomoto2017filmy_sample}
\end{figure}

In \cite{enomoto2017filmy}, the authors  propose to use a pix2pix architecture to remove thin clouds from images. They consider the RGB bands and the NIR band as input, where the NIR band is regarded to as additional (auxiliary) information provided to the network.
The cloud-free RGB image and a binary mask, indicating the cloudy pixels in the original input image, are returned as output.
% They encountered two obstacles concerning the training dataset: 1) bias due to the imbalance of the land cover classes since the majority of the images will belong to water or forest class and 2) gathering paired images under the same conditions for the free of cloud and with cloud domains. They solved the first issue by using \gls{tsne} to cluster the images into different categories based on land cover class then they selected training images uniformly from the classes.
Due to the difficulty of gathering paired images of the same location with and without clouds, the cloud coverage is simulated using Perlin noise \cite{perlin2002_noise} and then merged into the RGB image by alpha blending to get the synthetized cloudy image.  Finally, color correction is applied to both classes of images (i.e., images free of clouds and not) to improve the quality of the images.
The network is trained on images gathered from WorldView2, where the RGB bands and the NIR band are used as a multi-spectral input, and the $L1$ loss is computed between the 4-dim output (consisting of the generated cloud-free RGB images and the cloud mask), and  the corresponding   ground truth (the real RGB image with no clouds and the ground truth mask of cloudy pixels). %\BTcomm{The above text in brown is unclear. It seems that it does not make sense that the NIR is considiered in the L1 loss since we do not have the generated counterpart. I think that this is not the case. Maybe you mean the following  'the $L1$ loss is computed between the 4-dim output, constituted by the generated cloud-free RGB image and the binary mask, and the  ground truth, namely the real RGB image with no clouds and the ground truth mask of cloudy pixels.'. If this is not correct, then please rewrite so that it is clear.\LA{I didn't say they use the NIR because it is translated by the generator and we explained before the generator's out, I will replace it with your text since it is more clear}}
%Hence, the trained model generates %cloud-free images \CHLA{in addition %to} a binary mask \BTcomm{is this an %additional output of the %GAN??\LA{yes}} indicating the cloudy %pixels in the original input image.
Fig. \ref{fig:enomoto2017filmy_sample} shows some examples  of generated cloud-free RGB image and the generated cloud mask.
%with the relevant input RGB and NIR and the ground truth RGB image.
%
%\BTcomm{IN the description you provide, we do not see how the NIR band auxiliary information is used by the method\LA{Added clarifications}. Everything works without the NIR band (even if you do not say how the output binary mask is obtained). Please clarify these things.\LA{should be clarified now}}

% image with the \gls{nir} channel as input.
% enomoto2017filmy_sample

Several methods have been proposed to extend \cite{enomoto2017filmy}.
In \cite{wang2019} the authors propose to use a novel objective function to train the   pix2pix network, to the purpose of improving the quality of the generated cloud-free images. 
In \cite{singh2018}, the authors propose to substitute the pix2pix architecture with a \gls{cyclegan}  to avoid the need for a paired dataset. In this way it is possible to use real cloudy images instead of synthesized cloudy images for training the \gls{gan}. Moreover, only RGB bands are considered,  without the need of  auxiliary information.
%
% conditioned on images with thin clouds.
A similar approach is followed in \cite{zotov2019}.
%while  in \cite{wang2019} the authors added to the standard  pix2pix loss a loss term to improve the quality of the generated cloud-free image.

Later works performing cloud removals with GANs focus on the generation of different spectral bands, using different data types as auxiliary input data.
%Another research direction was the extension of the %number of input bands and the use of different auxiliary %data as input to the network. 
In \cite{grohnfeldt2018_conditional}, the authors resort to the same pix2pix architecture and the  clouds synthesis procedure described in \cite{enomoto2017filmy},  to train a GAN  on 10 bands of Sentinel-2 level-1C images, i.e., the bands with \SI{10}{\metre}  and \SI{20}{\metre}   GSD, and are used the \gls{sar} images as additional input.
Relying on \gls{sar} instead of just NIR as auxiliary information,
they are able to remove also thick clouds and dehaze the image. 
%\BTcomm{So, based on this description, it seems that  in this case cloud removal also works with tick clouds while the previous works on thin clouds. Where this improvement comes from? from the usage of SAR? If yes it must be said explicitly. If the cloud synthesis process is the same in \cite{enomoto2017filmy} as it seems I do not see any other reason for this improvement. \LA{yes that is correct the other only relied on NIR as additional information so this seems to be lacking in removing thick clouds while sar images are known to obtain images even in cloudy conditions, I added above a conclusion regarding that} }. 

Another work where  \gls{sar} images are  used  as auxiliary information  for the generation of cloud-free optical images
is \cite{ebel2020}.
%the authors use \gls{sar} imagery as auxiliary information added to cloudy optical images to generate cloud free optical images however in this work they 
The only difference between this work and \cite{grohnfeldt2018_conditional} 
is that a cycleGAN is adopted, instead of pix2pix, to avoid the need of a paired dataset.
In  \cite{gao2020cloud}, the authors propose to further improve declouding of optical images by exploiting datatype transfer Specifically, they first transfer from SAR to optical images.
Then, a GAN is used to fuse the simulated optical image, the SAR image and the optical image corrupted by clouds.
%, is fused to reconstruct the corrupted area by a generative adversarial network (GAN) with a particular loss function.
%by weighing them with the binary cloud map hence leaving non cloudy pixels without any change and replacing only the cloudy pixels with content transferred from the SAR image.
The simulated optical image provides the reference for the cloudy pixels. Fusion allows to inject accurate spectral information and high-frequency texture in the cloudy area.

\begin{figure}[htbp]
\centering
\includegraphics[clip,width=0.99\columnwidth]{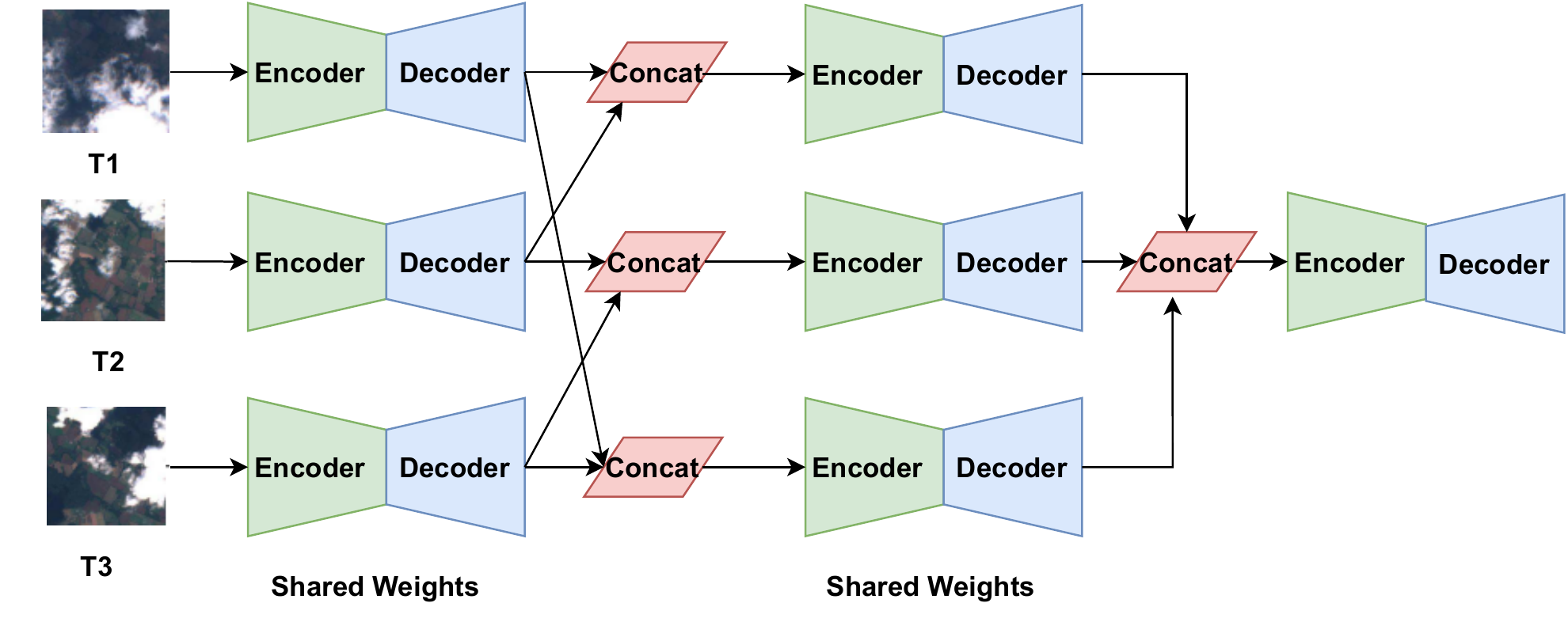}
\caption{Branched Resnet Architecture in \cite{sarukkai2020cloud}.}
%\BTcomm{I am afraid that the font (almost unreadable).}
\label{fig:sarukkai2020cloud_resnet}
\end{figure}

Other works focused on the use of more complicated network architectures and training procedures. For instance, in \cite{sarukkai2020cloud}, the authors propose a \gls{gan} that exploits temporal sequences of cloudy images, namely \gls{stgan}.
% For their experiments the authors built two datasets of RGB and NIR bands of Sentinel-2 images: one of paired cloud and cloud free images; another where samples are in groups of four, i.e., one image is cloud free and three are cloudy. They train the paired single image dataset on pix2pix with either RGB input or RGB and NIR.
Training is performed on a  dataset of RGB and NIR bands extracted from Sentinel-2 examples. 
A temporal sequence of three cloudy images is considered as an input, with  a cloud-free image used as reference.  The  three cloudy images and the cloud-free image are captured from the same location at different  times.
%\BTcomm{It seems that the cloud-free is the GT image that the GAN tries to simulate and not actually an input to the network...Below, you mention 3 inputs and not 4. Perhaps you mention the cloud free here because tof the paired dataset. But this must be clarified. AS it is written now, it seems that we have 4 inputs to the net, but then the part below only mention 3.\LA{yes the cloud free is used as a reference in the loss, I rephrased}}
%
%\BTcomm{This implicitly assumes that we have both the cloud-free and the cloudy versions for each single location. And what about the time period. Should the images be all taken in a short time period? \LA{they specify that they take the first three cloudy images closest in time to the cloud free image within a period of 35 days prior}}.
%
%
The proposed \gls{stgan} rely on a PatchGAN  discriminator. For the generator, two architectures have been considered, a branched ResNet-based and branched U-Net based architecture.
The branched ResNet processes each of the three cloudy image through a separate encoder-decoder. Then, the output features from the three branches are concatenated in pairs, and each pair is fed into another encoder-decoder; the outputs of the second  block of encoders-decoders are concatenated into one vector that is fed as input to a last encoder-decoder block. The architecture of the branched ResNet generator is illustrated in Fig. \ref{fig:sarukkai2020cloud_resnet}.
%
%
%the authors experimented with two architectural variations. The first architecture is a branched ResNet, processing the 3 cloudy images of the input through separate encoder-decoder branches and then concatenating the output features in a pairwise manner. The concatenated outputs constitute another three new inputs that are further processed through a separate encoder-decoder path. Finally, the output of this second encoder-decoder is concatenated to one input to a final encoder-decoder.
%
%
The U-Net, instead, processes each image in the input sequence through a separate encoder, and then all the outputs of the encoders are concatenated and fed to a single decoder. 
The proposed solution is compared against the baseline method in \cite{enomoto2017filmy}.
The results shown in \cite{sarukkai2020cloud} prove that, when using either branched Resnet or branched U-Net, \gls{stgan} provides better quality for the final generated cloud-free images (slightly better results are obtained when using the branched ResNet). 
In addition, it is shown that the cloud-free images generated using STGAN can also help for
land cover classification, as  classification  works better with STGAN generated cloud-free images with respect to the use of original cloudy images, or  cloud-free images generated with the baseline method.

Finally, in \cite{wen2021}, the authors improve the results of \cite{enomoto2017filmy} in terms of 
quality of the generated images considering the YUV color space for the input, instead of the RGB,  treating luminance and chroma components independently.
As a further difference with respect to \cite{enomoto2017filmy}, they resort to a WGAN \cite{arjovsky2017_cwgan}  architecture which is trained in two-steps. A first training is performed on  cloud-free and synthetic cloudy images. Then, the model is fine tuned on pairs of cloud-free and real cloudy images, thus avoiding training from scratch on a limited dataset.
This two-steps training process allows to get better performance in data scarcity conditions, when a limited set of acquired cloud-free and cloudy images for the same locations are available.

%\BTcomm{I think that we should include a figure with some examples of cloud-free images obtained with some of the above methods.  You can do that for the method for which  is more convenient for you (and obviously images are nice).\LA{ok, images are added}}

%\BTcomm{One question: why don't we have the figures mixed with the text? Was our choice to have all of them at the end? In general, I'd prefer having them inside the text (less boring). But if you prefer this way for some reason I will not object.\LA{no, they were mixed, I thought you changed them}}

\section{Forgery Detection and Localization}
\label{sec:detection}

In this section, we consider satellite image forensic methods for the detection of synthetic media and the localization of manipulations. We are interested in both the recognition of \gls{dnn}-generated contents (referred to as synthetic forgeries), being them an entire overhead image or just some regions of it, as well as in the identification of satellite images that have been manipulated with the help of common editing tools (i.e., Photoshop, GIMP, etc.) starting from genuine images.
In fact, while the emergence of applications of AI-tools for editing overhead imagery is worrying the community \cite{maliciousOverheadAI}, malicious modifications created with general purpose softwares like Photoshop are still a non-negligible threat. As a matter of fact, many satellite products are provided in formats easy to use and manipulate (e.g., GeoTIFF, JPEG, etc.). This element, together with the facility of use of editing software suites, allows even non-expert users to create credible forgeries, that can be used for instance to create misinformation campaigns \cite{australia_wildfire}.

To tackle with these menaces, the main idea behind most forensic methods is to exploit the concept of data life-cycle: during the existence of a multimedia object, various non-invertible operations are executed, each of them leaving a peculiar footprint that can be exploited to reconstruct the chain of operations the object has undergone. Forensic methods use this knowledge to expose malicious editing operations or to understand whether an image is authentic or it has been artificially synthentized.

The life-cycle of satellite imagery is characterized by a processing chain completely different from that of natural photographs, including the type of sensors and modalities used for their acquisition \cite{optic}, as well as the compression schemes used to encode them. Also the editing needed for making them manageable by final users (e.g., orthorectification \cite{ortho}, radiometric correction, etc.) includes  operations  that are very specific to the satellite context. All these reasons have pushed the multimedia forensic community to develop techniques specifically tailored to satellite data analysis.

\begin{figure}[htbp]
\centering
\includegraphics[clip,width=0.9\columnwidth]{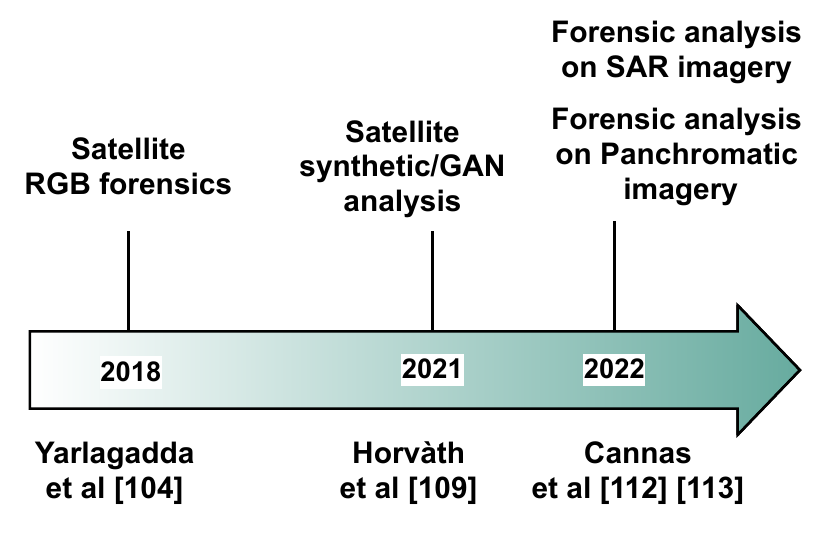}
\caption{{Timeline reporting the first forensic techniques for each datatype}}
\label{fig:dettimeline}
\end{figure}

In the following, we first provide some formal definition common to forensic detection and localization methods. Then, we overview the forensic techniques reported in \cref{tab:detectors}, grouping them into two categories: i) methods explicitly developed for synthetic image detection and localization; ii) methods for forgery detection and localization that, despite they can in principle be used also to spot synthetic content, are targeted to more general overhead imagery manipulations. Figure \ref{fig:dettimeline} highlights the timeline of the initial works related to each datatype.

Table \ref{tab:detectors_performance} provides some information about the datasets used by the various forensic detectors and their performance. Note that papers tend to use ad-hoc datasets, thus making it difficult to directly compare the reported results. The metrics used for detection are those typically used to evaluate binary classifiers: ROC-AUC \cite{fawcett_2006} (i.e., the area under the receiver operating characteristic curve) ; F1-score \cite{taha_2015} (i.e., the harmonic mean between precision and recall); Accuracy (i.e., the degree of closeness of the obtained answers to their actual value). In addition to these metrics, localization performance are evaluated also in terms of: PR-AUC \cite{he_2013} (i.e., the area under the precision-recall curve); Jaccard Index (JI) \cite{manning_2001} (i.e., the ratio between the number of correctly detected forged pixels and the number of pixels in the union set between actual and detected forged pixels).

\begin{table*}[]
\centering
\resizebox{\textwidth}{!}{%
\begin{tabular}{|l|l|l|l|p{0.44\linewidth}|}
\hline
\textbf{Reference}                             & \textbf{Year} & \textbf{Kind of data} & \textbf{Goal}                            & \textbf{Description}                                                                               \\ \hline
\cite{yarlagadda2018satellite}                 & 2018          & Color optical images  & Detect and localize splicings            & One-class GAN autoencoder as feature extractor followede by one-class SVM                          \\ \hline
\cite{bartusiak2019splicing}                   & 2019          & Color optical images  & Detect and localize splicings            & Conditional GAN is trained on two classes. Generator is used to estimate masks                     \\ \hline
\cite{horvath2019anomaly}                      & 2019          & Color optical images  & Detect and localize splicings            & Jointly trained autoencoder and one-class SVM                                                      \\ \hline
\cite{masmontserrat2020generative}             & 2020          & Color optical images  & Localize splicings                       & Use ensemble of PixelCNNs to estimate heatmaps from images                                         \\ \hline
\cite{horvath2020manipulation}                 & 2020          & Color optical images  & Detect and localize splicings            & One-class DBN for detection and localization                                                       \\ \hline
\cite{horvath2021_visiontransformer4detection} & 2021          & Color optical images  & Localize splicings                       & Visual Transformer used to reconstruct image. Binary mask obtained through input-output difference \\ \hline
\cite{horvath2021_attentionunet}               & 2021          & Color optical images  & Detect and localize GAN-based inpainting & Nested U-net trained to estimate heatmaps                                                          \\ \hline
%\cite{cannas2021open}                          & 2021          & Panchromatic images   & Satellite attribution                    & Open-set satellite attribution with CNNs                                                           \\ \hline
\cite{chen2021_geodefakehop}                   & 2021          & Color optical images  & Detect synthetic images                  & Subspace learning for fake image detection                                                         \\ \hline
\cite{zhao2021_cycleganrgb}                    & 2021          & Color optical images  & Detect synthetic images                  & Hand-crafted features and \gls{svm} to detect \gls{gan}-generated images                           \\ \hline
\cite{Ren2021deepfaking}                       & 2021          & Multispectral images  & Detect synthetic images                  & Binary CNN                                                                                         \\ \hline
\cite{cannas2022amplitude}                     & 2022          & SAR amplitude images  & Localize splicings                       & Extracted noise pattern fed to supervised or unsupervised segmentation methods                     \\ \hline
\cite{cannas2022panchromatic}                  & 2022          & Panchromatic images   & Copy-paste localization                  & Ensemble of attribution CNNs                                                                       \\ \hline
\end{tabular}%
}
\caption{List of forensics detectors and their characteristics.}
\label{tab:detectors}
\end{table*}

\begin{table*}[]
\centering
\resizebox{\textwidth}{!}{%
\begin{tabular}{|l|p{0.45\linewidth}|p{0.3\linewidth}|p{0.25\linewidth}|}
\hline
\textbf{Reference}                             & \textbf{Dataset}                                                                                & \textbf{Performance - Detection}                                                                                                                           & \textbf{Performance - Localization}                                                                                                                                    \\ \hline
\cite{yarlagadda2018satellite}                 & Ad-hoc spliced samples created on purpose using Landsat RGB images                                                 & $\text{ROC-AUC} = 89.6\%$                                                                                                                                  & $\text{PR-AUC} = 11.7\%$                                                                                                                                               \\ \hline
\cite{bartusiak2019splicing}                   & Ad-hoc spliced samples created on purpose using Landsat RGB images                                                 & $\text{ROC-AUC} = 100\%$                                                                                                                                   & $\text{PR-AUC} =95.3\%$                                                                                                                                                \\ \hline
\cite{horvath2019anomaly}                      & Ad-hoc spliced samples created on purpose using Sentinel-2 RGB images                                                & $\text{ROC-AUC} = 96.6\%$                                                                                                                                  & $\text{PR-AUC} =21.9\%$                                                                                                                                                \\ \hline
\cite{masmontserrat2020generative}             & Ad-hoc spliced samples created on purpose using Sentinel-2 RGB images                                                & n/a                                                                                                                                                        & $\text{PR-AUC} =59.9\%$                                                                                                                                                \\ \hline
\cite{horvath2020manipulation}                 & Ad-hoc spliced samples created on purpose using Sentinel-2 RGB images                                           & $\text{ROC-AUC} = 77.4\%$                                                                                                                                  & $\text{PR-AUC} =28.4\%$                                                                                                                                                \\ \hline
\cite{horvath2021_visiontransformer4detection} & Ad-hoc spliced samples created on purpose using DigitalGlobe and PlanetScope RGB images                & n/a                                                                                                                                                        & $\text{F1-score} = 0.36$, $\text{JI} = 0.275$                                                                                                               \\ \hline
\cite{horvath2021_attentionunet}               &  

Ad-hoc spliced samples created on purpose  using generated content by several GANs trained on Sentinel-2 RGB images

& \begin{tabular}[c]{@{}l@{}}$\text{ROC-AUC in SD scenario} = 91.8\%$\\ $\text{ROC-AUC in CD scenario} = 66.7\%$\end{tabular} & \begin{tabular}[c]{@{}l@{}}$\text{JI in SD scenario} = 73.5\%$\\ $\text{JI in CD scenario} = 0.21\%$\end{tabular} \\ \hline
\cite{chen2021_geodefakehop}                   & Synthetic samples created using different GANs trained on Google's Earth and CartoDB RGB images & $\text{F1-score} = 100\%$                                                                                                                                                           & n/a                                                                                                                                                                    \\ \hline
\cite{zhao2021_cycleganrgb}                    & Synthetic samples created using different GANs trained on Google's Earth and CartoDB RGB images & $\text{F1-score} = 95.3\%$                                                                                                                                 & n/a                                                                                                                                                                    \\ \hline
\cite{Ren2021deepfaking}                       & Synthetic samples created using Sentinel-2 \gls{msi}                                            & $\text{Accuracy} = 100\%$                                                                                                                                  & n/a                                                                                                                                                                    \\ \hline
\cite{cannas2022amplitude}                     & Ad-hoc spliced samples created on purpose using Sentinel-1 \gls{sar} images                                        & n/a                                                                                                                                                        & $\text{Mean JI} = 0.674$                                                                                                                                         \\ \hline
\cite{cannas2022panchromatic}                  & Ad-hoc spliced samples created on purpose using DigitalGlobe panchromatic images                                                                             & n/a                                                                                                                                                        & $\text{ROC-AUC} = 85.7\%$                                                                                                                                              \\ \hline
\end{tabular}%
}
\caption{List of datasets used  and average performances of the forensic detectors. Results not available in the original papers are denoted with `n/a'. For \cite{horvath2021_attentionunet}, ROC-AUC values are averaged, and SD and CD indicate respectively the same dataset and cross dataset scenarios. JI represents the Jaccard Index.} 
\label{tab:detectors_performance}
\end{table*}

%\BTcomm{\\ GENERAL: The scope of this review of detection methods is even too broad ! I thought that we were overviewing the approaches (data driven or not) detecting AI-based manipulations. However it seems that here you are reviewing ALL the forgery detection methods...This neither agrees with the title :( At least, this should be made very very clear here and already in the introduction. We can also try to motivate our choice or just explain why we made it. I think it is because most of the AI-based forgery methods detection are very recent and the  literature about detection is behind manipulation and so there are not so many methods for the detection of AI-based forgeries, while there are many of them for traditional manipulation (most of them AI-based). And I guess that this is the first survey on this topic...}

%\BTcomm{Given the above: perhaps we can say that  what this overview does is surveying the use of AI for manipulation/generation and detection (If we remove the few statistical-based methods we have now)}

\subsection{Forensic detectors definitions}
\label{subsec:det_and_loc}

% \PBedit{The forensic analysis of synthetically generated images can be carried out with two different goals in mind: i) detection, and ii) localization. Under the detection assumption, the analyst is interested in estimating the likelihood that the image under analysis has been generated synthetically (partly or completely). Under the localization assumption, the analyst is interested in estimating which region of an image has been manipulated through synthetic generation techniques.}
The forensic analysis of overhead images can be carried out with two different goals in mind: i) detection, and ii) localization. Under the detection assumption, the analyst is interested in estimating the likelihood that the image under analysis has been tampered with (partly or completely). With localization, the analyst is interested in estimating which region of an image has been manipulated.

%In particular, we consider as forgery the case of image splicing. This is the local modification of a pixel region of a target satellite image by means of insertion of a portion of an alien image or any extraneous objects, with a possible further local application of editing operations for image adjustment.

\begin{figure}[t]
\centering
\subfloat[Manipulated $\Xrgb$.]{\includegraphics[width=0.4\columnwidth]{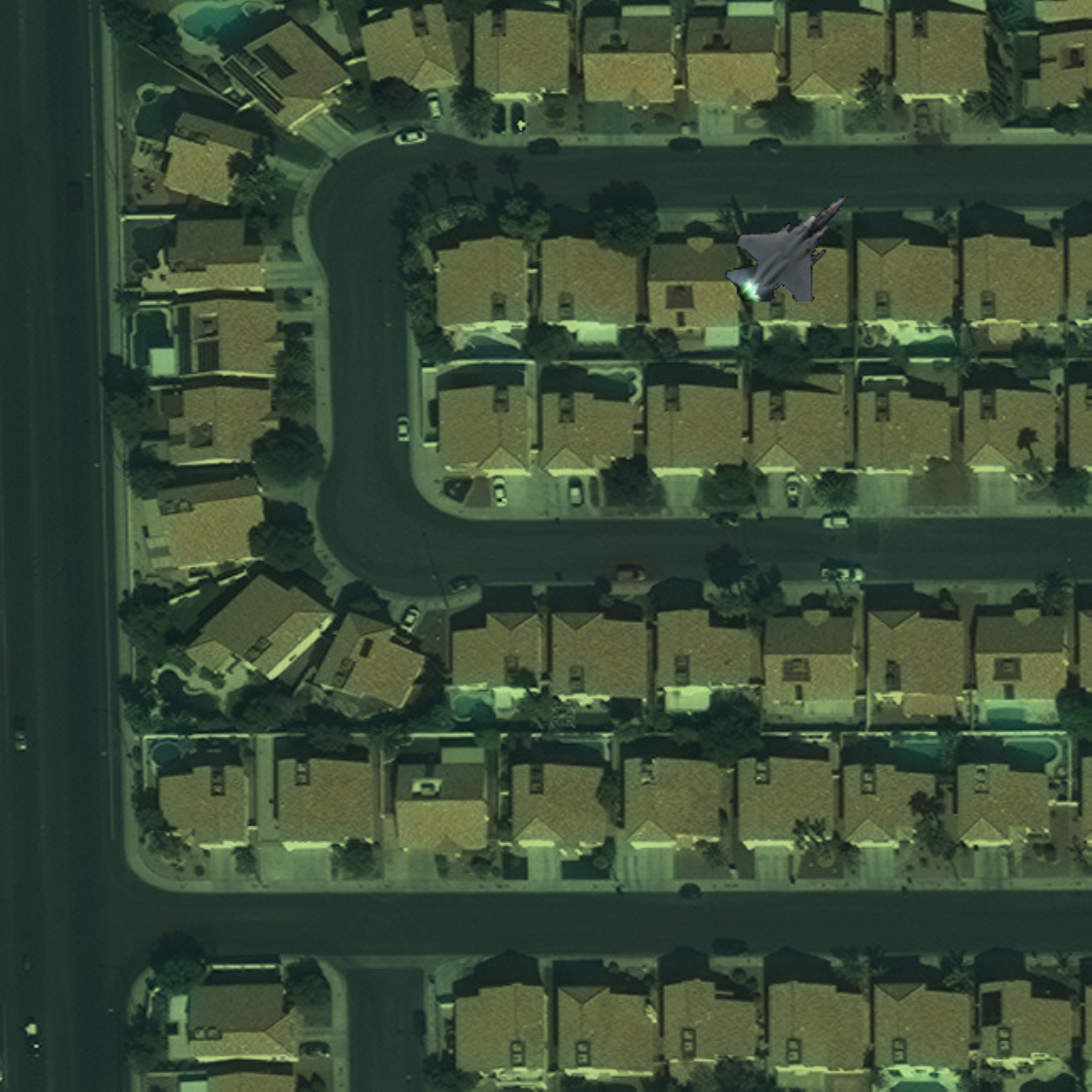}
\label{fig:rgb_spliced_example}}
\hfil
\subfloat[Tampering mask $\M_\text{RGB}$.]{\includegraphics[width=0.4\columnwidth]{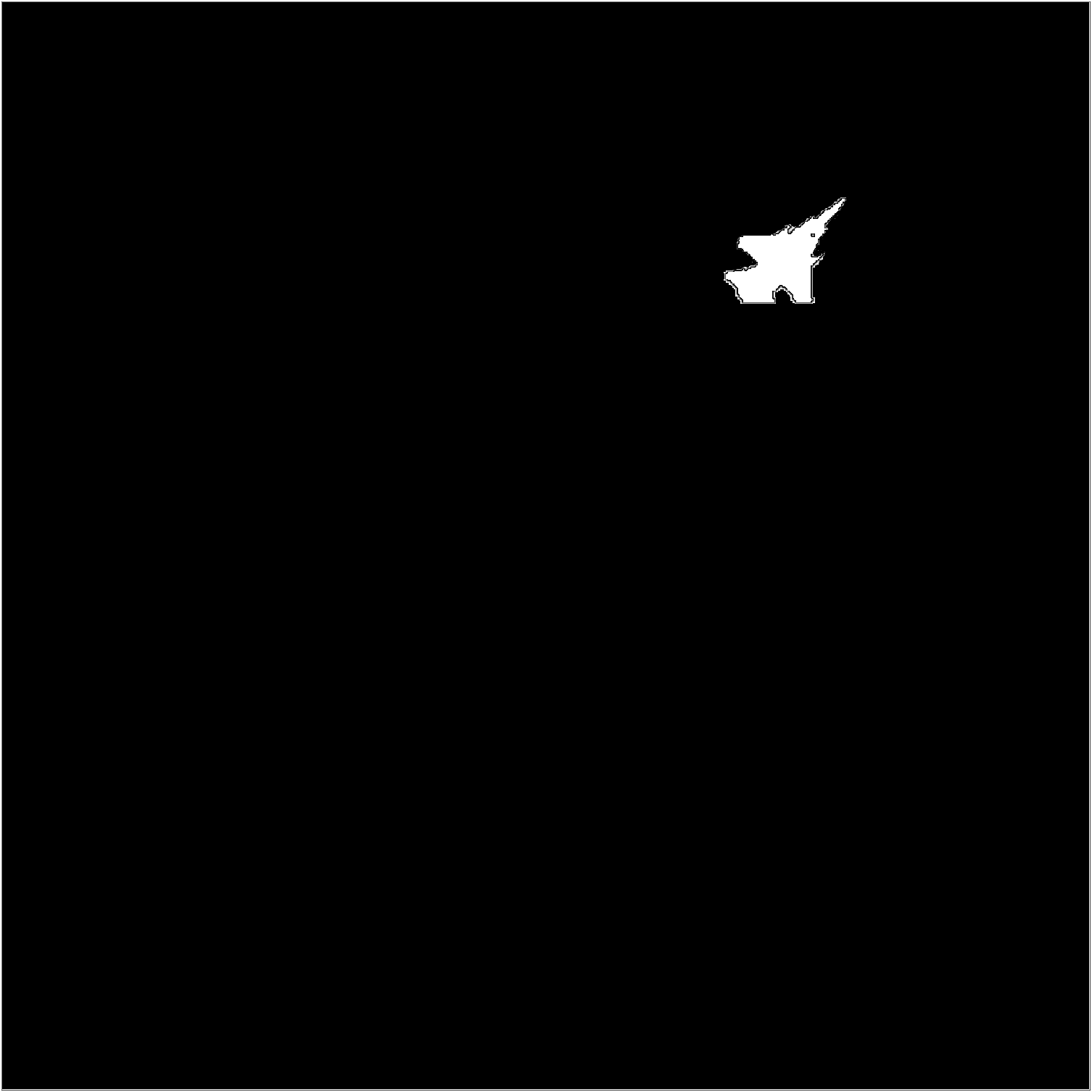}
\label{fig:rgb_tampering_mask_example}}

\subfloat[Manipulated $\Xsar$.]{\includegraphics[width=0.4\columnwidth]{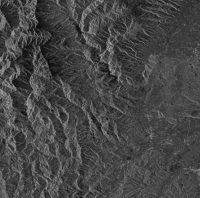}
\label{fig:sar_spliced_example}}
\hfil
\subfloat[Tampering mask $\M_\text{SAR}$.]{\includegraphics[width=0.4\columnwidth]{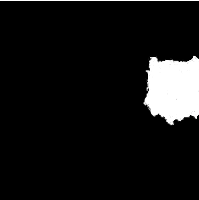}
\label{fig:sar_tampering_mask_example}}

\subfloat[Manipulated $\Xpan$.]{\includegraphics[width=0.4\columnwidth]{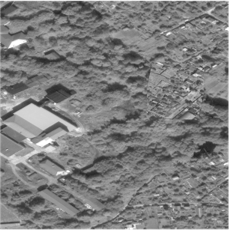}
\label{fig:pan_spliced_example}}
\hfil
\subfloat[Tampering mask $\M_\text{Pan}$.]{\includegraphics[width=0.4\columnwidth]{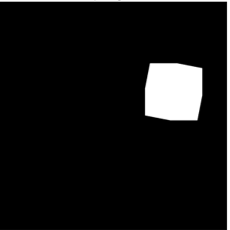}
\label{fig:pan_tampering_mask_example}}

\caption{Examples of manipulated satellite images from different modalities, together with their tampering mask indicating the attacked area. On top row, RGB data from \cite{yarlagadda2018satellite}; in the middle, \gls{sar} from \cite{cannas2022amplitude}; at the bottom, panchromatic from \cite{cannas2022panchromatic}.}
\label{fig:splicing_examples}
\end{figure}

Formally, let us define a generic satellite image with $U \times V$ pixel resolution as $\X$. The coordinates of its pixels can be represented as $(u, v)$, where $u \in [1, \dots, U]$ and $ v \in [1, \dots, V]$.
$U$ and $V$ are the number of rows and columns, respectively. 
% \MB{Why not simply the number of rows and columns?}. 
The pixel-by-pixel integrity of the image can be defined by a tampering mask $\M$, with the same resolution of $\X$, whose pixels take a binary value 1 or 0 depending on whether the corresponding pixel has been manipulated or not. More formally
\begin{equation}
    \label{eq:tamp_mask}
    \mathbf{M}(u, v) = \begin{cases}
    1,\quad \text{if $(u,v)\in\mathcal{S}$}\\
    0,\quad \mathrm{otherwise}
    \end{cases},
\end{equation}
where $\mathcal{S}$ denotes the manipulated region. Given an image $\X$, we can attribute to it a binary label $\y$ of value $0$ if $\X$ is pristine, and $1$ if it contains manipulated content. \cref{fig:splicing_examples} provides an example of manipulations executed on different modalities of satellite data.

\begin{figure}[t]
\centering
\subfloat[Spliced $\Xpan$.]{\includegraphics[width=0.4\columnwidth]{diagrams/PAN_splicing_example.png}
\label{fig:pan_spliced}}
\hfil
\subfloat[Ground truth tampering mask $\M_\text{Pan}$.]{\includegraphics[width=0.4\columnwidth]{diagrams/PAN_splicing_example_mask.png}
\label{fig:pan_tampering_mask}}

\subfloat[Soft tampering mask $\Msoft_{\text{Pan}}$.]{\includegraphics[width=0.4\columnwidth]{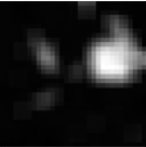}
\label{fig:pan_soft_mask}}
\hfil
\subfloat[Hard tampering mask $\Mbin_\text{Pan}$.]{\includegraphics[width=0.4\columnwidth]{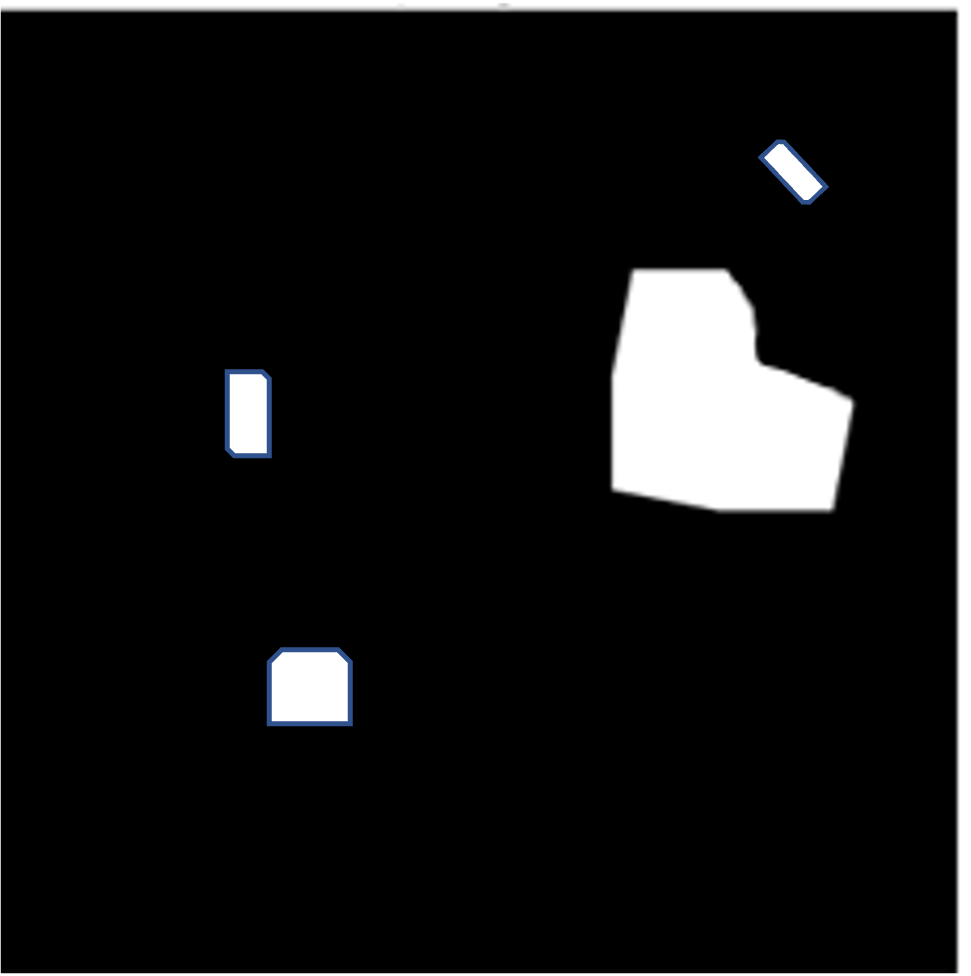}
\label{fig:pan_hard_mask}}

\caption{Examples of soft $\Msoft$ and hard $\Mbin$ tampering masks obtained from a spliced panchromatic sample presented in \cite{cannas2022panchromatic}.}
\label{fig:splicing_examples_soft}
\end{figure}

In this framework, detecting if an image has been manipulated consists in designing a detector that implements a function $d(\X)$ returning a soft forgery score $\ysoft$ (i.e., the likelihood that $\y=1$) or a hard forgery score $\ybin$ (i.e., an estimate of $\y$).
Localizing a forgery consists in either estimating a soft tampering mask $\Msoft$ (i.e., the pixel-by-pixel likelihood that each pixel has been forged) or a hard tampering mask $\Mbin$ (i.e., an estimate of $\M$). An example of a soft and hard mask is reported in \cref{fig:splicing_examples_soft}.

\subsection{Synthetic image detection and localization}
\label{subsec:synthetic_det}
Section~\ref{sec:forgeries} has shown that it is possible to generate high-quality synthetic satellite images. As it happened with other types of artificial data (e.g., synthetic images of faces, animals, etc.), concerns regarding the possible malicious usage of these generators are raising.

This section deals with the detection and localization of forgeries generated through synthetic satellite image generation tools. For this kind of forgeries, the majority of forensics tools rely on data-driven techniques. This is partly motivated not only by the complex processing pipeline characterizing satellite imagery, but also by the fact that the forensic traces contained in forgeries generated through neural networks are not easy to model. A common approach consists therefore in training specialized \glspl{cnn} for detecting and localizing synthetic forgeries.

%In the following we consider works that deal with both tasks, i.e., using the same notation introduced for splicing attacks in Section~\ref{subsec:forgery_det}, either that implement a detection function $d(\X)$ returning a soft $\ysoft$ or hard $\ybin$ estimate of $\y$, or that instead want to localize at a pixel level where the forgery has been executed, i.e., that estimate a soft $\Msoft$ or hard $\Mbin$ tampering mask highlighting a synthetically forged region $\mathcal{F}$.

% \item \cite{horvath2021_attentionunet} manipulation detection using nested Unet
% So far, the research community focused mainly on RGB generated data.
In \cite{horvath2021_attentionunet},
% \wtitle{Nested Attention U-Net: A Splicing Detection Method for Satellite Images \cite{horvath2021_attentionunet}} 
the authors localize RGB forgeries generated by 3 different families of \glspl{gan}: StyleGAN2; ProGAN; \gls{cyclegan}. More specifically, the authors have generated a dataset of Sentinel-2 RGB images splicing them with each one of the analyzed \glspl{gan}.
Their approach is based on a modified version of a \gls{cnn} used for image segmentation called \gls{naun}.
This network takes as input a $\Xrgb$ image and outputs directly a soft mask $\Msoft$ with each pixel value indicating the likelihood that the pixel has been  generated by one of the analyzed \glspl{gan}.

Being a data-driven method, the \gls{naun} needs to be trained directly on synthetically spliced samples. Nevertheless, the authors demonstrate the feasibility of their solution in a cross-dataset scenario, i.e., training the \gls{naun} on images spliced with a specific type of \gls{gan} and testing it on images generated by a \gls{gan} which was not included in the training set.

% \item \cite{zhao2021_cycleganrgb} hand-crafted features and \gls{svm} to detect \gls{gan}-generated images
In \cite{zhao2021_cycleganrgb},
% \wtitle{Deep fake geography? When geospatial data encounter Artifical Intelligence \cite{zhao2021_cycleganrgb}} 
the authors not only present a \gls{gan}-based approach for semantic satellite image translation, but also provide a data-driven approach for detecting the synthetically generated images. As a matter of fact, they rely on \glspl{svm} trained with a number of features (i.e., spatial, histogram-based and spectral features plus a combinations of the three) to detect \gls{cyclegan} generated images.

% \item \cite{chen2021_geodefakehop} network to detect synthetic satellite images
Using the same dataset presented by Zhao et al. \cite{zhao2021_cycleganrgb}, in \cite{chen2021_geodefakehop}
% \wtitle{Geo-DeFakeHop: High-Performance Geographic Fake Image Detection \cite{chen2021_geodefakehop}} 
Chen et al. provide another data-driven method to detect semantically translated RGB satellite images. The core concept is based on the assumption that \glspl{gan} have difficulties in creating high-frequency details in generated samples.
They therefore analyze an input $\Xrgb$ image dividing it into several patches, and perform a frequency analysis using a filter-bank named \gls{saab} transform. Coefficients from each filter are first used to train a XGBoost \cite{chen2016xgboost} classifier learning patch by patch the most discriminant frequency coefficients. Then, coefficients from all patches are considered as features for taking a global decision regarding the authenticity of the analyzed sample.

% \item ``Deepfaking it: experiments in generative, adversarial multispectral remote sensing'' \PB{Cannot download it}
Finally, in \cite{Ren2021deepfaking}
% \wtitle{Deepfaking it: experiments in generative, adversarial multispectral remote sensing \cite{Ren2021deepfaking}} 
Ren et al. rely on a simple \gls{cnn} to detect season transferred multi-spectral images generated through \glspl{cyclegan}. In their study they reach very good detection performances, even though one of their major findings consists in the fact that relying on a simple \gls{cnn} exposes the detection process to adversarial attacks.
However, they have showed that adversarial training can make such a simple approach effective in distinguishing synthetically generated images. They also provide some insights about the features that a \gls{cnn}-based classifier finds more useful in the classification of \gls{gan}-generated multispectral images, thanks to the use of gradient-interpretability techniques such as Integrated Gradients \cite{Sundararajan2017axiomatic}.

\subsection{Forgery detection and localization}
\label{subsec:forgery_det}
% These methods detect and / or localize local forgeries on overhead images.
% They are all data-driven.
% The majority is supervised and returns a heatmap.
% Some only train on pristine data.
% The difficulty behind the modeling of forensic features that characterize the task of sensor attribution, unfortunately affects also the development of solutions for general forgery detection and localization in overhead imagery.
The literature on methods explicitly focusing on the detection of synthetic forgeries is rather limited. However, synthetically generated content can in principle be treated as a specific kind of local or global forgery. It is therefore possible to exploit general purpose satellite image forgery detectors to also expose synthetic forgeries. Indeed, image forgery detection consists in understanding if the image under analysis has been edited (in part or in its totality according to some specific editing definition) or not. Image forgery localization consists in understanding which pixels of the image under analysis have been edited (if any). Editing may be a splicing operation, as well as the insertion of synthetically generated content.

Current image forensics \gls{sota} solutions for forgery detection and localization rely on the use of data-driven solutions \cite{Verdoliva2018deep}. Not surprisingly, this trend also applies to satellite images. As a matter of fact, data-driven techniques enable automatic extraction of meaningful forensics features from corpora of data. In this way, it is possible to devise a good solution also in situations in which data models may be complex or uncertain (e.g., due to the wide variety of possible satellite products).

In this section, we analyze the \gls{sota} data-driven forensics solutions developed for the tasks of forgery detection and localization in satellite imagery.
In the following, we organize the discussion about forgery detection and localization methods based on the modality of satellite imagery analyzed by the considered techniques: i) RGB images; ii) panchromatic images; iii) \gls{sar} images.

\subsubsection{RGB images}

In \cite{yarlagadda2018satellite}
% \wtitle{Satellite image forgery detection and localization using GAN and one-class classifier \cite{yarlagadda2018satellite}} 
Yarlagadda et al. tackle the problem of detecting and localizing general image manipulations on RGB satellite images as an anomaly detection task. More specifically, their approach consists in the training of a \gls{cnn} as an autoencoder.

As introduced in Section~\ref{sec:generativemodels}, autoencoders are a particular kind of \glspl{nn} whose goal is to reconstruct at the output the data provided as input. While this procedure may seem trivial, training of the autoencoder $\emph{A}$ in \cite{yarlagadda2018satellite} is performed in such a way that the hidden representation vector $\mathbf{h}$ possesses some desirable properties. Indeed, the dimension of $\mathbf{h}$ is forced to be considerably smaller than that of the input $\X_\text{RGB}$. Therefore, the autoencoder is forced to learn only the most salient features of the input in order to reconstruct it properly at the output.
%This makes undercomplete autoencoders suitable for data compression tasks \cite{Goodfellow2016}.

For forensics purposes, Yarlagadda et al. train an autoencoder only on pristine satellite RGB images. The expected outcome is that during training the function $\Phi_\text{Enc}(\cdot)$ is optimized to extract information regarding original RGB data. The hidden representation vector $\mathbf{h}$ becomes a low-dimensional representation fitted to pristine images only. Spliced samples can be recognized as their hidden representation constitutes an anomaly with respect to the distribution of vectors extracted from original data.

To this end, the authors propose to train an autoencoder to reconstruct $64\times 64$ patches extracted from RGB satellite data using a L2 loss. They compare two different training strategies: an autoencoder trained as a standalone network, called by Yarlagadda et al. ``without \gls{gan}'' scenario; an autoencoder trained as a \gls{gan} generator starting from a pre-trained autoencoder for initialization, i.e., ``with \gls{gan}'' scenario as named by the authors. In both cases, after training, only the encoding function $\Phi_\text{Enc}(\cdot)$ is retained to provide the hidden representation vectors $\mathbf{h}$.

The extracted hidden vectors are then used to train a 1-class \gls{svm}. The \gls{svm} is trained on vectors $\mathbf{h}$ extracted from pristine images only. The algorithm learns the boundary in the feature space of the $\mathbf{h}$ vectors enclosing the distribution of pristine samples. After convergence, the \gls{svm} implements a detection function $\emph{d}_\text{SVM}(\mathbf{h})$ which outputs a soft value $\ysoft$ representing the likelihood that a vector $\mathbf{h}$ belong to the pristine data distribution.

The proposed framework is flexible enough to handle also localization of manipulations. As a matter of fact, for samples of resolution greater than $64\times 64$ pixels, by dividing the input $\Xrgb$ into patches of the resolution expected by the network and producing a detection score $\ysoft_i$ for each patch $\P^i$, a soft tampering mask $\Msoft$ can be created by assigning the score $\ysoft_i$ to all the pixels belonging to the patch $\P^i$ from which it has been computed.
The authors have tested the performance of the algorithm on spliced RGB data generated starting from images of the Landsat mission. Spliced samples have been realized pasting objects of various shapes (i.e., clouds, airplanes, etc.) and dimensions (i.e, $70\times 70$ pixels, $128\times 128$ pixels, etc.). %\PB{sometimes we report patch size, sometimes we don't. Is that ok, or is it better to be coherent?}
% and the pipeline obtained good localization performances.

Even in \cite{bartusiak2019splicing},
% \wtitle{Splicing Detection And Localization In
% Satellite Imagery Using Conditional GANs \cite{bartusiak2019splicing}} 
Bartusiak et al. work on both manipulation detection and localization. The authors rely on the same dataset used in \cite{yarlagadda2018satellite}, but propose a different approach to analyze RGB data.
In particular, they exploit a two-class \gls{cgan}: starting from a set of spliced $\Xrgb$ images, the \gls{cgan} generator is trained to estimate tampering masks $\Msoft$ as close as possible to the original $\M$, with the discriminator judging their quality. At convergence, only the generator is kept as a splicing localization tool. To accomplish this task, they rely on a combination of the \gls{cgan} loss together with \gls{bce} for evaluating the discriminator performances.

Given an estimated tampering mask $\Msoft$, the authors are also able to produce a single tampering score for the analyzed sample by proceeding as follows: if the input image $\Xrgb$ is pristine, then $\Msoft$ should correspond to a dark image with no bright pixels (i.e., it contains low values close to $0$); conversely, if a splicing attack is localized, $\Msoft$ should highlight the $\mathcal{S}$ region as a bright area (i.e., pixel values close to $1$). Such behavior can be captured by a detection function defined as
\begin{equation}
    \label{eq:bartusiak_splicing_detection}
    \emph{d}(\Msoft) = \frac{\sum_{u=1}^{U} \sum_{v=1}^{V}\Msoft(u, v)}{U\times V} \quad ,
\end{equation}
with $U$ and $V$ indicating the input resolution as in \cref{eq:tamp_mask}. This detection score is proportional to the average pixel values of the forgery mask: the higher the value, the higher the likelihood that the image has been edited.

In \cite{horvath2019anomaly}
% \wtitle{Anomaly-Based Manipulation Detection in Satellite Images \cite{horvath2019anomaly}} 
Horv{á}th et al. rely on an anomaly detection approach for detecting and localizing splicing attacks similarly to Yarlagadda et al. \cite{yarlagadda2018satellite}. In particular, they extend the idea proposed in \cite{yarlagadda2018satellite} adapting the concept of deep one-class classifier presented by Ruff et al. in \cite{Ruff2018deep}. They also rely on the same dataset used in \cite{yarlagadda2018satellite}.
The solution, named \gls{svdd}, consists in using an autoencoder $\emph{A}$ to learn a hidden representation vector $\mathbf{h}$ capturing the most salient features of pristine RGB data, and a \gls{svm} working as an anomaly detector with respect to the distribution of these vectors extracted from pristine data. However, instead of training separately a 1-class \gls{svm} on hidden representation vectors extracted from a training set (as in \cite{yarlagadda2018satellite}), \gls{svdd} trains jointly the autoencoder with the \gls{svm} obtaining better latent representations. Moreover, the authors exploit the so-called kernel trick to learn a non-liner boundary of the pristine class hidden representation distribution.

With respect to \cite{Ruff2018deep}, Horv{á}th et al. also experiment with different optimizers and activation functions.
The authors extract patches of different resolutions (i.e., $32\times 32$, $64\times 64$, etc.) from RGB satellite data, and generate estimate tampering masks $\Msoft$ using the same approach of Yarlagadda et al. \cite{yarlagadda2018satellite}. With respect to the original solution, they are also able to perform detection on samples bigger than the patch resolution on which \gls{svdd} has been trained. In particular, given an estimate mask $\Msoft$, they compute the detection score as
\begin{equation}
    \label{eq:satsvdd_splicing_detection}
    \emph{d}(\Msoft) = \frac{\max(\Msoft)-\mu(\Msoft)}{\sqrt{\frac{\sigma^2(\Msoft)}{\max(\Msoft)}}} \quad ,
\end{equation}
where $\mu$ and $\sigma^2$ denote, respectively, the mean and standard deviation computed on $\Msoft$ pixels, and $\max(\cdot)$ the maximum value operator. Such measure accounts for the fact that spliced samples should present an high anomaly detection value in $\Msoft$, which is high when compared to the average pixel value of $\Msoft$, and also high compared to the standard deviation normalized by its maximum value.

In \cite{horvath2020manipulation},
% \wtitle{Manipulation Detection in Satellite Images Using Deep Belief Networks \cite{horvath2020manipulation}} 
Horv{á}th et al. exploit \glspl{dbn} \cite{Hinton2006reducing} in a similar manner to the way autoencoders are used in \cite{yarlagadda2018satellite, horvath2019anomaly}. \glspl{dbn} are a particular type of neural networks composed by the stacking of several \glspl{rbm} \cite{Ackley1985learning}, a particular type of perceptron where the relation between the input (i.e., visible units) and the output (i.e., hidden units) is expressed in terms of conditional probability distributions.

The authors construct a \gls{dbn} by stacking two \glspl{rbm}, with the goal of training the network to reconstruct RGB patches similarly to an autoencoder. The first \gls{rbm} takes an input patch and outputs an hidden representation vector with reduced dimensionality with respect to the input patch. The second \gls{rbm} takes this hidden representation vector as input and outputs a hidden representation of the same resolution of the input image.

For forensics purposes, Horv{á}th et al. exploit this characteristic and train the \gls{dbn} only on pristine satellite data, spotting spliced RGB images as anomalies in the reconstruction process executed by the second \gls{rbm}. The expected outcome is that the network learns to retain in the hidden units only the most relevant statistical properties of original RGB data. When tested on forged data, the reconstruction is not perfect, and therefore by measuring the reconstruction error it is possible to discriminate spliced images.
% By enforcing the first \gls{rbm} to produce an hidden representation of dimensionality lower than the input patch, and training the complete \gls{dbn} to reconstruct image patches with a low $\emph{L}_{\ell2}$ loss, the network learns the most salient statistical properties of the distribution of pristine patches. When presented with spliced samples at test time, it is possible to recognize them as such by simply looking at the error in the reconstruction process.

With respect to the methods proposed in \cite{yarlagadda2018satellite} and \cite{horvath2019anomaly}, the authors still analyze an input $\Xrgb$ dividing it into different patches $\P^i$ and reconstruct them one by one, but the estimate mask $\Msoft$ is created by simply assigning the L2 loss computed between original and reconstructed patches. A detection score for the entire sample is instead computed by using  \cref{eq:satsvdd_splicing_detection}.

In \cite{masmontserrat2020generative}
% \wtitle{Generative Autoregressive Ensembles for Satellite Imagery Manipulation Detection \cite{masmontserrat2020generative}} 
Mas Montserrat et al. rely again on a sort of anomaly detection approach to perform splicing localization. Instead of exploiting autoencoders, they build their solution on generative autoregressive models, in particular on PixelCNNs \cite{Oord2016pixel} and Gated PixelCNNs \cite{Oord2016conditional}.

These neural networks are designed as generative models, i.e., they model the conditional distribution of pixel values given the values of their neighbors. More specifically, they provide the conditional likelihood for the value of a pixel in an image, using as conditional variables the values of the surrounding pixels, and as distribution the one learned during training. Considering all pixels as i.i.d. variables, the probability of an entire image $\Xrgb$ can be computed. As a matter of fact, given a pixel $x_i$ in a single channel image $\X$, with $i, \dots, N$ being the total number of pixel in it, these models compute the conditional probability of the image pixel values as:
\begin{equation}
    p(\X) = \prod_{i=1}^{N} p(x_i | x_1, \dots, x_{i-1}).
\end{equation}
%In this way, these models are able to generate new images also given as input a test image, for instance by raster scanning (i.e., going row by row, left to right) each pixel in a probe sample. \PB{This is not very clear to me, but PixelCNN has never been clear to me.}

The idea behind this pipeline is that training the generative autoregressive models only on pristine data, they are able to learn the conditional distribution of pixels of pristine RGB data. When presented with a spliced sample, the trained models will output a very low likelihood when processing spliced pixels, thus allowing the creation of an estimate tampering mask $\Msoft$.
Therefore, in order to localize splicing manipulations the authors train PixelCNNs and Gated PixelCNNs only on pristine images. At the deployment stage, they then test the solution on spliced samples from the dataset presented in \cite{yarlagadda2018satellite}. To have a more accurate and robust prediction, the authors run some experiments with model ensembling, i.e., averaging the predictions from multiple networks with different parameters and trained on different pixel scan ordering.
%\PB{I am a bit confused on how this generator outputs a mask. Maybe a pipeline figure would help.}

% \gls{cnn} for RGB satellite image splicing detection
% \cite{fouad2020detection} \PB{Very bad journal, very badly written. Shall we include it? A lot of details are missing...} \EC{Agree on this one too.}

% \cite{horvath2021_visiontransformer4detection} manipulation detection in satellite images using Vision Transformer
In \cite{horvath2021_visiontransformer4detection}
% \wtitle{Manipulation Detection in Satellite Images Using Vision Transformer \cite{horvath2021_visiontransformer4detection}} 
another anomaly detection approach based on autoencoders is proposed, relying on VisionTransformers \cite{Kolesnikov2021transformers} instead of \glspl{cnn}. VisionTransformers avoid to process RGB image patches $\P^i$ independently. Rather, the whole architecture is designed to jointly compare all pixels among different patches, denoted as tokens. %\PB{We should better explain the token idea which is not so well-known by people not using transformers.}

\begin{figure}[htbp]
\centering
\includegraphics[width=1\columnwidth]{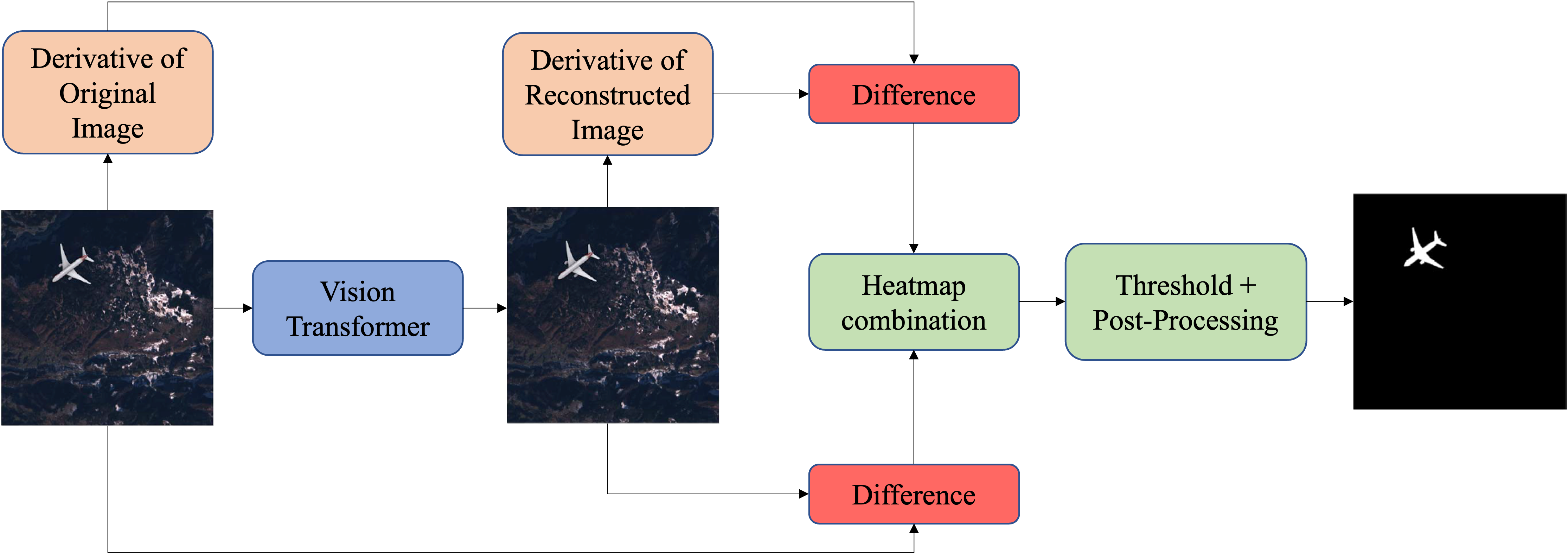}
\caption{Vision Transformer pipeline proposed in \cite{horvath2021_visiontransformer4detection}.}
\label{fig:vt_pipeline}
\end{figure}

The authors train the VisionTransformer to reconstruct pristine RGB data similarly to an autoencoder, using as loss function a smoothed version of the L1 loss.
%\PB{As for the PixelCNN I don't have the input-output pipeline very clear in my mind at this point. Maybe a figure would help.}
At deployment time, the authors do not simply consider the loss between the reconstructed and input sample, but proceeded in the following way. Defining as $\Xrgb$ the original input, and as $\Xhrgb$ the VisionTransformer reconstructed output:
\begin{enumerate}
    \item apply a $3\times 3$ convolution with a Laplacian filter $\emph{l}(\cdot)$ to both $\Xrgb$ and $\Xhrgb$. The reason behind this is to use the Laplacian filter as an edge detector, enhancing the high-frequency components of the reconstructing samples, as autoencoders notoriously suffer in the generation of such artifacts;
    \item compute an estimate tampering mask as
    $$\Msoft = \frac{\bigl(\Xrgb - \Xhrgb\bigr) +  \bigl(\emph{l}(\Xrgb) - \emph{l}(\Xhrgb)\bigr)}{2};$$
    \item impose a threshold on $\Msoft$ to generate a binary splicing localization mask $\Mbin$;
    \item apply a series of morphological operators to clean $\Mbin$ removing false positives.
\end{enumerate}
\cref{fig:vt_pipeline} provides a visual representation of the complete pipeline. The authors test their solution on a dataset of spliced samples generated starting from Sentinel-2 images and Maxar's WorldView-3 satellite images, considering spliced objects of different sizes extracted from images captured by PlanetScope satellites.

% \cite{cozzolino2020noiseprint} Noiseprint on satellite RGB data
% \cite{cozzolino2020noiseprint}

\subsubsection{Panchromatic}
% \item \cite{cannas2022panchromatic} Copy-paste localization for panchromatic images based on ensemble of \glspl{cnn} trained for image attribution task.
Panchromatic imagery is a valuable asset for a variety of remote sensing applications. It combines the spectral information of the RGB bands, and as such it appears as a grayscale format of images. Its main feature consists in the broader wavelength range it represents. This translates in a greater spatial resolutions. For this reason, a common use for this modality of data is pan-sharpening, i.e., the combination with RGB and multi-spectral data to increase the resolution of the latter.

Due to their importance, in \cite{cannas2022panchromatic}
% \wtitle{Panchromatic imagery copy-paste localization through data-driven sensor attribution \cite{cannas2022panchromatic}} 
Cannas et al. investigate the problem of splicing localization for panchromatic data. This work builds upon the concept of image source attribution, which is a task widely studied in the forensic literature for natural images \cite{Kirchner2015forensic}. More specifically, \cite{cannas2021open} shows that it is possible to understand which satellite has been used to generate a panchromatic image by means of a carefully trained ensemble of \glspl{cnn}.

Building upon \cite{cannas2021open}, \cite{cannas2022panchromatic} exploits \glspl{cnn} to spot copy-paste attacks, i.e., splicing attacks where the the spliced region $\mathcal{S}$ comes from a sample captured by a different sensor. The main idea is that it is possible to extract local traces that link an image region to the satellite used for its acquisition. If traces of multiple satellites are found, then the image contains a spliced region.

To implement this idea, the authors first train an ensemble of $\emph{N}$ different \glspl{cnn} to perform sensor attribution on pristine images, i.e., assign an input $\Xpan$ to one of $M$ available satellites in a patch-by-patch fashion. They exploit the same data collection procedure followed in \cite{cannas2021open}, training each network in the ensemble on a different subset of satellites in order to make the ensemble more robust to changes in data distributions.

At test time, the $n$-th \gls{cnn} of the ensemble extracts from each patch $\P^i$ a feature vector
\begin{equation}
    \mathbf{f}^i_n = [f_n^1, \; f_n^2, \; \dots, \; f_n^M],    
\end{equation}
where the $m$-th element is the likelihood for that patch to belong to the $m$-th satellite.

To spot copy-paste attacks, the authors propose to capture the deviation of sensor feature vectors extracted from each image patch with respect to the average sensor feature vector characterizing the image. Deviations from this global descriptor likely indicate that one of the vectors (i.e., one of the patches) comes from a satellite that is different from the one that captured the rest of the image, indicating a copy-paste attack.

\begin{figure}[htbp]
\centering
\includegraphics[width=1\columnwidth]{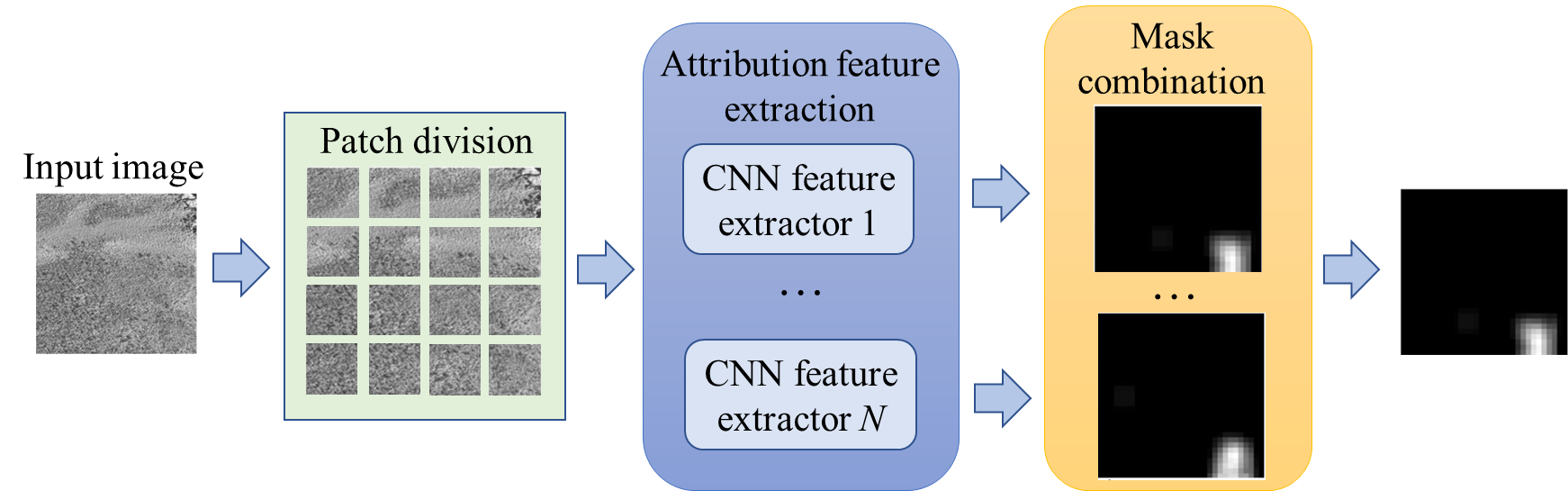}
\caption{Panchromatic copy-paste localization pipeline proposed in \cite{cannas2022panchromatic}.}
\label{fig:pan_pipeline}
\end{figure}

Formally, this is done for each \gls{cnn} by computing the average sensor feature of the image under analysis as the arithmetic mean of all $\mathbf{f}^i_n$ vectors over $i$ (i.e., over all patches).
The deviation is computed as the L2 distance between the average feature and each vector $\mathbf{f}^i_n, \, \forall i$. By assigning to each patch the deviation from the average feature, the authors build a tampering mask $\Msoft_n$. Repeating the procedure for all networks in the ensemble, Cannas et al. generate different estimate tampering masks, which are aggregated into an estimate tampering mask $\Msoft$ through a weighted averaging procedure. 
\cref{fig:pan_pipeline} presents a graphical representation summarizing the whole pipeline.

\subsubsection{SAR images}
% \item \cite{cannas2022amplitude} \gls{sar} amplitude image splicing localization through denoising \gls{cnn} (Noiseprint-like) followed by supervised or unsupervised technique for heatmap estimation.
%{\em Due to its characteristic of providing high resolution imagery independently on the weather conditions and daylight, \gls{sar} imagery has been widely exploited for remote sensing applications.}
%\BTcomm{I would remove the part in italic. We already  said this  in Section II. This is a repetition. }
Even if \gls{sar} images are complex radar signals, amplitude-based products gained a lot of popularity recently. This is also thanks to the many platforms, such as the Copernicus Open Acces Hub, which offer them in easy to manage formats. Unfortunately, such products (e.g., those of the Sentinel-1 mission)  are also easy to manipulate as they can be edited by relying on any common image processing suite (e.g., Photoshop).

\begin{figure}[t]
\centering

\subfloat[Spliced $\Xsar$.]{\includegraphics[width=0.3\columnwidth]{diagrams/SAR_splicing_example.png}
\label{fig:sar_spliced}}
\hfil
\subfloat[Tampering mask $\M_\text{SAR}$.]{\includegraphics[width=0.3\columnwidth]{diagrams/SAR_splicing_example_mask.png}\label{fig:sar_gt}}
\hfil
\subfloat[Extracted fingerprint.]{\includegraphics[width=0.3\columnwidth]{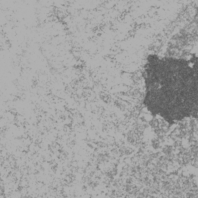}\label{fig:sar_finger}}

\caption{Examples of \gls{sar} forged image \protect\subref{fig:sar_spliced}, ground truth tampering mask \protect\subref{fig:sar_gt}, and fingerprint \protect\subref{fig:sar_finger} extracted by the method proposed in \cite{cannas2022amplitude}.}
\label{fig:cannas_fingerprint_example}
\end{figure}

For this reason, in \cite{cannas2022amplitude}
% \wtitle{Amplitude SAR Imagery Splicing Localization \cite{cannas2022amplitude}} 
the authors analyze the problem of splicing localization in amplitude SAR imagery. The basic idea behind their approach is that each \gls{sar} product is characterized by peculiar processing traces due to the complex pipeline needed to generate it.
Hence, the authors rely on a \gls{cnn} purposely trained to extract processing traces relative to the overall generation pipeline of amplitude \gls{sar} products. These traces are expressed as a fingerprint of the same resolution of the analyzed image, the fingerprint image, which highlights, with different patterns, regions that underwent different processing operations. In this way, spliced regions $\mathcal{S}$ are highlighted as spots with a different texture. Fig. ~\ref{fig:cannas_fingerprint_example} provides an example of a fingerprint computed as in \cite{cannas2022amplitude}.

After extraction, the fingerprint is analyzed to produce a binary tampering mask $\Mbin$ localizing precisely the spliced pixels. The authors experiment with both unsupervised and supervised techniques that turn the fingerprint into a binary mask, testing their solutions on a dataset of spliced samples created starting from \gls{grd} images downloaded from the Copernicus Open Access Hub and using different kinds of editing to conceal the spliced area $\mathcal{S}$.

% \BTcomm{The impression I have is that the detection is (by far) behind generation, and  detectors for some types of forgery (e.g. datatype translation) have not being developed yet. If this is the case we should perhaps stress this somewhere and in the concluding remarks.}
% \PB{I tried adding this remark to the conclusions.}

\section{Discussion and future challenges}
\label{sec:disc}
% MAIN POINTS:
% 1) the multimedia forensics community has recognized the strategic role of overhead imagery, and started to focus on this kind of data.
% 2) despite the great availability of tools for the forensic analysis of natural imagery, specific tools are needed for satellite data. This is due to the fact that overhead imagery presents a different nature wrt standard digital photographs (e.g., SAR data, multi-spectral imagery) and for those modality of overhead close to natural imagery (i.e., EO like RGB) the processing pipelines behind the generation of such satellite products are very different. This implies that standard pictures footprints are not guaranteed to provide optimal performance.
% 3) Due to complex processing pipeline characterizing satellite data, deep learning tools have been for now the preferred tool for their forensic analysis. Deep learning instruments as a matter of fact, on top of being the SOTA for the analysis of natural images as well, uplift the burden of modeling specific forensic footprints for the data at hand, which might be a very complex operation.
% 4) As the multimedia forensics community deepens its knowledge in the overhead field, we might expect the appearance of methods that exploit forensic traces characterizing the lifecycle of satellite imagery, as well as the exploitation of clues specifically related to the scenes capture in overhead images.
In this section, we recap the possibilities available today to generate and manipulate satellite images, and the techniques investigators and users can adopt to detect non-genuine satellite images, highlighting their main limitations in order to pave the way to future work.

\subsection{Synthetic overhead imagery generation}

In Section~\ref{sec:forgeries}, we have overviewed different approaches to create satellite images from scratch or to manipulate their semantic content using generative models.
Additionally, in Section~\ref{sec:bforgeries} we have shown how it is possible to further edit satellite images through techniques that allow to enhance their quality and change their image type. In light of this, it is easy to understand how these techniques could be used by malicious users, allowing them to create high-quality synthetic forgeries of various types. At the same time, generation techniques are far from perfect. All methods developed so far have some limitations that offer new opportunities for future research.

One of the limitation common to all the techniques developed so far is that they do not work on full resolution remote sensing data. They rather consider small image patches up to a resolution of 512$\times$512 pixels. This is due to the inability of the current generative models to store into the memory of GPUs full satellite images, which are usually characterized by resolutions that can reach the gigapixel. This means that a high-resolution synthetic image can only be obtained by stitching together multiple smaller patches. This is sub-optimal from a visual point of view, and leaves traces that can in principle be exploited by forensics detectors.

Another aspect that is often overlooked is related to the fact that some kinds of satellite images have peculiar characteristics that make them very different from natural photographs. These characteristics are not always properly taken into account. For instance, multi-spectral and hyper-spectral data are characterized by correlated multiple bands, whereas most generation techniques usually generate  single bands independently. \gls{sar} images are complex in nature (i.e., a phase term is also present in addition to the amplitude image) but generation techniques developed so far, focused mainly on the image amplitude. Many kinds of satellite images come with 10 to 16 bit of depths, but generation techniques typically return 8 bit data. All of these aspects should be taken into account by future generators in order to produce realistic satellite images.

Another important characteristic of satellite images is that they are accompanied by abundant metadata. For instance, metadata can provide information related to the kind and position of the satellite used for acquisition, the kind of processing images have undergone to facilitate visual inspection, positional data allowing to geo-locate the image on the planet's surface, as well as other custom data spanning from land-cover information to digital elevation models providing altitude details. These pieces of information are often completely stripped out by synthetic image generators that only work considering raster images in the pixel domain. On the contrary, it would be advisable to take metadata into account during the generation process in order to further increase the quality and plausibility of the synthetic images. Moreover, it would be interesting to allow generators to be driven by metadata themselves (e.g., to generate an image that is coherent with a set of imposed metadata in terms of geo-location, pixel elevation, land-cover, etc.).

If some of these limitations will be naturally solved in the (near) future thanks to technological improvements (e.g., GPU memory cost will decrease allowing to generate higher resolution data), others pose research challenges that require the development of new ad-hoc methods.\\

\subsection{Forensic analysis of satellite images}

Section ~\ref{sec:detection} illustrated how the multimedia forensics community has started paying attention to overhead imagery. Despite the few works published so far, researchers have recognized the strategic role of this kind of data. Therefore a significant growth in the number of contributions is expected in the near future.

One aspect that is stressed in several works is that the nature of satellite images may strongly differ from that of natural photographs. This is evident for \gls{sar} data and \gls{msi}, but it also holds for \gls{eo} products, whose complex processing pipelines have little in common with the life cycle of standard digital pictures. This drives the need for specific forensic tools tailored to overhead data. 

These tools nowadays often rely on deep learning techniques. This is possible thanks to the increasing availability of \gls{eo} products on the Web. As a consequence, many of the proposed forensic tools ended up relying on data-driven technology, causing two possible issues. On one hand, common \gls{cnn} techniques developed for image analysis are tailored to work with 8 bit images with standard photograph resolutions. These tools have to be adapted to work with 16 bit data at higher resolution, which may cause both numerical instability and memory issues. On the other hand \gls{gan}-based methods often look like black boxes that offer little to no interpretation to analysts. This is a major concern as forensic results must be supported by strong evidence in courts of trials. Therefore, we foresee as a future research line the development and implementation of explainable forensic tools, which allow a forensic analyst to better understand why a certain decision has been taken.

Another relevant aspect that current detectors have not exploited yet is the presence of multiple image modalities. Satellite images often come bundled with additional images. For instance, it is customary to work with multiple bands if we think at \glspl{msi}. Moreover, it is common to have pairs of \gls{sar} and \gls{eo} observations of the same region. This could be exploited by forensic detectors to further verify image integrity by means of a joint analysis of different modalities.

Similarly, we expect that metadata could become an additional asset in the hands of forensic analysts. Indeed, as satellite images often come with a great amount of additional side information, this could be exploited to verify if an image is authentic or it has been modified. This aspect could be of paramount importance not only for methods linking images to devices, but also for local forgery analysis (e.g., verify the compatibility of shadows with the sun position and acquisition time).

Finally, we believe that model-based solutions could still be viable techniques in the detection of satellite image forgeries. As this kind of images undergo a set of very specific operations at acquisition time (e.g., multiple acquisition stitching, image projection from the earth surface to a planar bi-dimensional image, multiple band fusion, etc.) an informed analyst could spot interesting forensic traces that may be not visible upon visual inspection. 
%This kind of detectors 
Detectors based on such traces may only come from a close collaboration between the multimedia forensic community and the remote sensing one.

\section{Conclusion}
We have reviewed the different facets characterizing the generation of synthetic satellite images. In our analysis, we have divided \gls{sota} works depending on the envisaged application. In doing so, we have seen that several techniques enable image generation from scratch, while others aim at satellite modality translation and image quality improvement. The vast majority of these techniques has been developed to support a wide range of remote sensing applications. Indeed, due to the diffusion of deep learning tools also in the satellite field, the need for precise and annotated data has become more pressing than before. Despite the numerous sources of remote sensing data available on the Web \cite{freesat}, the use of synthetically generated images can be of great help.

On the negative side, the malicious use of synthetically generated satellite images can lead to severe consequences. As a matter of fact, the increased availability of remote sensing data, while paving the way to numerous research and application opportunities, also exposes this kind of information to manipulations and attacks \cite{russia}.
For this reason, we have also overviewed the forensic tools available to contrast the inappropriate usage of synthetic satellite images. From this analysis, it is evident that satellite imagery represents a complete new challenge to forensic analysts. In particular, detection and localization of synthetic forgeries have been seldom studied in the literature so far. Luckily, different general-purpose forgery detection and localization methods tailored to both \gls{eo} and \gls{sar} imagery have been proposed in the recent literature. We envision that these techniques can be adapted to the problem of exposing synthetic manipulations too.
%and the tools at disposal, even if they have been developed for multimedia objects presenting some similarities with satellite data (e.g., digital natural images), are not guaranteed to offer optimal performances. This is due to the sensibly different life-cycle characterizing satellite images. The forensics community is therefore putting a great effort in adapting existing tasks and concepts to this new scenario (e.g., satellite image attribution, forgery detection and localization), but also in developing solutions tailored to different kinds of remote sensing signals (e.g., \gls{eo} imagery, \gls{sar}, etc.) and to newly emerging threats (e.g., new techniques for synthesizing satellite images).

We are confident that our work will help shedding light on the opportunities that are opening in the near future, as well as on the challenges practitioners must face with regarding the generation and the analysis of synthetic satellite images.
In particular, while data generation techniques seem to have reached quite an advanced status, forensic analysis of this kind of imagery still falls behind. As the forensic research community is still making its first steps in the remote sensing field area, the increasing role that satellite images are acquiring in an ever growing number of applications shows the urgent need for the development of appropriate forensic tools.

In this context, future work could be dedicated to the extraction of forensics footprints characterizing the life-cycle of satellite imagery, or to the development of methods that take into account the physical properties of the depicted scene (i.e., physical interpretable methods). Moreover, since many forensics methods are currently based on deep learning tools, other research lines may touch the field of artificial intelligence interpretability and the fusion of traces extracted from different forensics methods. %\MB{Should we also mention the study of the vulnerability of DL-based techniques to adversarial examples?}
\label{sec:conclusion}

\bibliographystyle{IEEEtran}
\bibliography{references}

\end{document}

%% file: latex_helpers.tex
\newacronym{sota}{SOTA}{State of the Art}
\newacronym{nn}{NN}{Neural Network}
\newacronym{dnn}{DNN}{Deep Neural Network}
\newacronym{ssl}{SSL}{Semi-Supervised Learning}
\newacronym{aviris}{AVIRIS}{Airborne Visible/Infrared Imaging Spectrometer}
\newacronym{hsi}{HSI}{Hyper-Spectral Imagery}
\newacronym{cnn}{CNN}{Convolutional Neural Network}
\newacronym{gan}{GAN}{Generative Adversarial Network}
\newacronym{cgan}{cGAN}{Conditional Generative Adversarial Network}
\newacronym{nir}{NIR}{Near Infrared}
\newacronym{sgd}{SGD}{Stochastic Gradient Descent}
\newacronym{lr}{lr}{Learning Rate}
\newacronym{oa}{OA}{Overall Accuracy}
\newacronym{msi}{MSI}{Multi-Spectral Imagery}
\newacronym{gsd}{GSD}{Ground Sampling Distance}
\newacronym{toa}{TOA}{Top-of-Atmosphere}
\newacronym{boa}{BOA}{Bottom-of-Atmosphere}
\newacronym{wgan-gp}{WGAN-GP}{Wasserstein GAN  with Gradient Penalty loss}
\newacronym{msg}{MSG}{Multi-Scale Gradients}
\newacronym{rmsprop}{RMSProp}{Root Mean Squared Propagation}
\newacronym{nicegan}{NICE-GAN}{No Independent Component for Encoding GAN}
\newacronym{snap}{SNAP}{Sentinel Application Platform}
\newacronym{vv}{VV}{vertical transmit and vertical receive}
\newacronym{hh}{HH}{horizontal transmit and horizontal receive}
\newacronym{slc}{SLC}{Simple Look Complex}
\newacronym{grd}{GRD}{Ground Range Detected}
\newacronym{mstar}{MSTAR}{Moving and Stationary Target Acquisition and Recognition}
\newacronym{ocn}{OCN}{Ocean}
\newacronym{vh}{VH}{vertical transmit and horizontal receive}
\newacronym{hw}{HV}{horizontal transmit and vertical receive}
\newacronym{psnr}{PSNR}{peak signal-to-noise ratio}
\newacronym{ssim}{SSIM}{structural similarity}
\newacronym{mae}{MAE}{mean absolute error}
\newacronym{mape}{MAPE}{mean absolute percentage error}
\newacronym{msa}{MSA}{mean spectral angle}
\newacronym{svm}{SVM}{Support Vector Machine}
\newacronym{sar}{SAR}{Synthetic Aperture Radar}
\newacronym{wv}{WV}{Wave}
\newacronym{iw}{IW}{Interferometric Wide swath}
\newacronym{ew}{EW}{Extra Wide swath}
\newacronym{sm}{SM}{Stripmap}
\newacronym{knn}{KNN}{K-Nearest Neighbors}
\newacronym{km}{KM}{K-Means}
\newacronym{gmm}{GMM}{Gaussian Mixture Models}
\newacronym{dbn}{DBN}{Deep Belief Network}
\newacronym{rf}{RF}{Random Forest}
\newacronym{bce}{BCE}{Binary Cross Entropy}
\newacronym{prnu}{PRNU}{Photo Response Non-Uniformity}
\newacronym{ccd}{CCD}{Charge-Coupled Devices}
\newacronym{cmos}{CMOS}{Complementary Metal-Oxide Semiconductor}
\newacronym{svdd}{SatSVDD}{Satellite Support Vector Data Descriptor}
\newacronym{rbm}{RBM}{Restricted Boltzmann Machine}
\newacronym{naun}{NAU-N}{Nested Attention U-Net}
\newacronym{eo}{EO}{Electro-Optical}
\newacronym{lwir}{LWIR}{Long-Wave Infra-Red}
\newacronym{esa}{ESA}{European Space Agency}
\newacronym{usgs}{USGS}{United States Geological Survey}
\newacronym{unoosa}{UNOOSA}{United Nations - Office for Outer Space Affairs}
\newacronym{gcp}{GCP}{Ground Control Points}
\newacronym{ncc}{NCC}{Normalized cross-correlation}
\newacronym{cc}{CC}{correlation coefficient}
\newacronym{sam}{SAM}{spectral angle mapping}
\newacronym{rmse}{RMSE}{root mean squared error}
\newacronym{sift}{SIFT}{Scale Invariant Feature Transform}
\newacronym{brisk}{BRISK}{Binary Robust Invariant Scalable Key}
\newacronym{progan}{ProGAN}{Progressive Generative Adversarial Network}
\newacronym{vae}{VAE}{Variational Autoencoder}
\newacronym{dem}{DEM}{Digital Elevation Models}
\newacronym{spade}{SPADE}{spatially adaptive normalization}
\newacronym{iou}{IoU}{intersection over union}
\newacronym{nwp}{NWP}{numerical weather prediction}
\newacronym{mdn}{MDN}{mixture density network}
\newacronym{tsne}{t-SNE}{t-distributed stochastic neighbor embedding}
\newacronym{cyclegan}{CycleGAN}{Cycle Generative Adversarial Network}
\newacronym{stgan}{STGAN}{Spatio-Temporal Generative Adversarial Network}
\newacronym{swir}{SWIR}{Short-Wave InfraRed}
\newacronym{saab}{SAAB}{Subspace Approximation with Adjusted Bias}

\def\X{\mathbf{X}}
\def\Xrgb{\X_\text{RGB}}
\def\Xhrgb{\hat{\X}_\text{RGB}}
\def\Xsar{\X_\text{SAR}}
\def\Xpan{\X_\text{Pan}}
\def\P{\mathbf{P}}
\def\M{\mathbf{M}}
\def\Msoft{\tilde{\M}}
\def\Mbin{\hat{\M}}
\def\y{y}
\def\ysoft{\tilde{\y}}
\def\ybin{\hat{\y}}
\def\genNet{ \Phi_\text{G}}
\def\disNet{ \Phi_\text{D}}
\def\sgenNet{ \Phi_\text{G2}}
\def\sdisNet{ \Phi_\text{D2}}

\definecolor{purple}{RGB}{165,0,255}
\definecolor{Darkcyan}{RGB}{0,139,139}
\newcommand{\LA}[1]{\textcolor{purple}{LA: {#1}}}
\newcommand{\CHLA}[1]{\textcolor{olive}{ {#1}}}
\newcommand{\PB}[1]{\textcolor{Darkcyan}{PB: {#1}}}

\newcommand{\wtitle}[1]{\textbf{\textit{#1}}}

%% file: overhead_overview_generation_detection.bbl
% Generated by IEEEtran.bst, version: 1.14 (2015/08/26)
\begin{thebibliography}{100}
\providecommand{\url}[1]{#1}
\csname url@samestyle\endcsname
\providecommand{\newblock}{\relax}
\providecommand{\bibinfo}[2]{#2}
\providecommand{\BIBentrySTDinterwordspacing}{\spaceskip=0pt\relax}
\providecommand{\BIBentryALTinterwordstretchfactor}{4}
\providecommand{\BIBentryALTinterwordspacing}{\spaceskip=\fontdimen2\font plus
\BIBentryALTinterwordstretchfactor\fontdimen3\font minus
  \fontdimen4\font\relax}
\providecommand{\BIBforeignlanguage}[2]{{%
\expandafter\ifx\csname l@#1\endcsname\relax
\typeout{** WARNING: IEEEtran.bst: No hyphenation pattern has been}%
\typeout{** loaded for the language `#1'. Using the pattern for}%
\typeout{** the default language instead.}%
\else
\language=\csname l@#1\endcsname
\fi
#2}}
\providecommand{\BIBdecl}{\relax}
\BIBdecl

\bibitem{kostopoulos2020}
\BIBentryALTinterwordspacing
V.~Lappas and V.~Kostopoulos, ``A survey on small satellite technologies and
  space missions for geodetic applications,'' in \emph{Satellites Missions and
  Technologies for Geosciences}, V.~Demyanov and J.~Becedas, Eds.\hskip 1em
  plus 0.5em minus 0.4em\relax Rijeka: IntechOpen, 2020, ch.~8. [Online].
  Available: \url{https://doi.org/10.5772/intechopen.92625}
\BIBentrySTDinterwordspacing

\bibitem{purnamasayangsukasih2016}
\BIBentryALTinterwordspacing
P.~R. Purnamasayangsukasih, K.~Norizah, A.~A.~M. Ismail, and I.~Shamsudin, ``A
  review of uses of satellite imagery in monitoring mangrove forests,''
  \emph{{IOP} Conference Series: Earth and Environmental Science}, vol.~37, p.
  012034, jun 2016. [Online]. Available:
  \url{https://doi.org/10.1088/1755-1315/37/1/012034}
\BIBentrySTDinterwordspacing

\bibitem{ukraine2022}
CNN, \emph{New satellite images show buildup of Russian military around
  Ukraine}, March 2022 (accessed March 11, 2022),
  \emph{https://edition.cnn.com/2022/02/02/europe/russia-troops-ukraine-buildup-satellite-images-intl/index.html}.

\bibitem{ukraine2022mariupol}
T.~Guardian, \emph{Mariupol bombing: before and after satellite images show
  destruction in Ukraine city}, March 2022 (accessed March 11, 2022),
  \emph{https://www.theguardian.com/world/2022/mar/10/mariupol-bombing-ukraine-before-and-after-satellite-images-map-russian-attack-residential-maternity-childrens-hospital}.

\bibitem{russia}
Mashable, \emph{Satellite images show clearly that Russia faked its MH17
  report}, May 2015 (accessed March 11, 2022),
  \emph{http://mashable.com/2015/05/31/russia-fake-mh17-report}.

\bibitem{australia_wildfire}
G.~Rannard, \emph{Australia fires: Misleading maps and pictures go viral},
  January 2020 (accessed June, 2022),
  \emph{https://www.bbc.com/news/blogs-trending-51020564}.

\bibitem{bbc_ukraine}
J.~Wakefield, \emph{Ukraine crisis: Satellite data firm asks for war images},
  January 2020 (accessed June, 2022).

\bibitem{isola_2016}
P.~Isola, J.~Zhu, T.~Zhou, and A.~A. Efros, ``Image-to-image translation with
  conditional adversarial networks,'' in \emph{2017 {IEEE} Conference on
  Computer Vision and Pattern Recognition, {CVPR} 2017, Honolulu, HI, USA, July
  21-26, 2017}.\hskip 1em plus 0.5em minus 0.4em\relax {IEEE} Computer Society,
  2017, pp. 5967--5976.

\bibitem{zhu_2017}
J.~Zhu, T.~Park, P.~Isola, and A.~Efross, ``Unpaired image-to-image translation
  using cycle-consistent adversarial networks,'' \emph{IEEE International
  Conference on Computer Vision (ICCV)}, 2017.

\bibitem{Piva2013overview}
A.~Piva, ``An overview on image forensics,'' \emph{ISRN Signal Processing},
  vol. 2013, p.~22, November 2013.

\bibitem{Popescu2005exposing}
A.~C. Popescu and H.~Farid, ``Exposing digital forgeries by detecting traces of
  resampling,'' \emph{IEEE Transactions on signal processing}, vol.~53, no.~2,
  pp. 758--767, 2005.

\bibitem{Kirchner2008fast}
M.~Kirchner, ``{Fast and reliable resampling detection by spectral analysis of
  fixed linear predictor residue},'' in \emph{ACM workshop on Multimedia and
  Security (MM\&Sec)}, 2008.

\bibitem{Bianchi2011detection}
T.~Bianchi and A.~Piva, ``Detection of nonaligned double jpeg compression based
  on integer periodicity maps,'' \emph{IEEE transactions on Information
  Forensics and Security}, vol.~7, no.~2, pp. 842--848, 2011.

\bibitem{mandelli2018multiple}
S.~Mandelli, N.~Bonettini, P.~Bestagini, V.~Lipari, and S.~Tubaro, ``Multiple
  jpeg compression detection through task-driven non-negative matrix
  factorization,'' in \emph{2018 IEEE International Conference on Acoustics,
  Speech and Signal Processing (ICASSP)}.\hskip 1em plus 0.5em minus
  0.4em\relax IEEE, 2018, pp. 2106--2110.

\bibitem{Cozzolino2015splicebuster}
D.~Cozzolino, G.~Poggi, and L.~Verdoliva, ``Splicebuster: A new blind image
  splicing detector,'' in \emph{IEEE International Workshop on Information
  Forensics and Security (WIFS)}, 2015.

\bibitem{Bayar2016deep}
B.~Bayar and M.~C. Stamm, ``A deep learning approach to universal image
  manipulation detection using a new convolutional layer,'' \emph{Proceedings
  of the ACM Workshop on Information Hiding and Multimedia Security}, pp.
  5--10, June 2016, {Vigo, Spain}.

\bibitem{Bondi2017tampering}
\BIBentryALTinterwordspacing
L.~Bondi, S.~Lameri, D.~G\"{u}era, P.~Bestagini, E.~J. Delp, and S.~Tubaro,
  ``Tampering detection and localization through clustering of camera-based
  {CNN} features,'' \emph{Proceedings of the IEEE Conference on Computer Vision
  and Pattern Recognition Workshops}, pp. 1855--1864, July 2017, {Honolulu,
  HI}. [Online]. Available: \url{dx.doi.org/10.1109/CVPRW.2017.232}
\BIBentrySTDinterwordspacing

\bibitem{cozzolino2020noiseprint}
D.~Cozzolino and L.~Verdoliva, ``Noiseprint: a {CNN}-based camera model
  fingerprint,'' \emph{IEEE Transactions on Information Forensics and Security
  (TIFS)}, vol.~15, pp. 144--159, 2020.

\bibitem{bonettini2020use}
N.~Bonettini, P.~Bestagini, S.~Milani, and S.~Tubaro, ``On the use of
  {Benford}'s law to detect {GAN}-generated images,'' in \emph{International
  Conference on Pattern Recognition (ICPR)}, 2021.

\bibitem{mandelli2020training}
S.~Mandelli, N.~Bonettini, P.~Bestagini, and S.~Tubaro, ``Training cnns in
  presence of jpeg compression: Multimedia forensics vs computer vision,'' in
  \emph{IEEE International Workshop on Information Forensics and Security
  (WIFS)}, 2020, pp. 1--6.

\bibitem{alamayreh2021detection}
O.~Alamayreh and M.~Barni, ``Detection of gan-synthesized street videos,'' in
  \emph{European Signal Processing Conference (EUSIPCO)}, 2021, pp. 811--815.

\bibitem{Gragnaniello2022}
D.~Gragnaniello, F.~Marra, and L.~Verdoliva, \emph{Detection of AI-Generated
  Synthetic Faces}.\hskip 1em plus 0.5em minus 0.4em\relax Springer
  International Publishing, 2022, pp. 191--212.

\bibitem{Toth2016remote}
C.~Toth and G.~Jóźków, ``Remote sensing platforms and sensors: A survey,''
  \emph{ISPRS Journal of Photogrammetry and Remote Sensing}, 2016.

\bibitem{optic}
P.~University, \emph{Optical Sensors overview}, November 2020 (accessed
  November 20, 2020), \emph{https://www.e-education.psu.edu/geog480/node/444}.

\bibitem{wv2technical}
M.~Technologies, \emph{Radiometric Use of WorldView-2 Imagery - Technical
  Note}, November 2010 (accessed March 11, 2022).

\bibitem{skysattechnical}
P.~L. PBC, \emph{Planet Imagery Products Specifications}, March 2022 (accessed
  March 13, 2022).

\bibitem{Teillet86image}
\BIBentryALTinterwordspacing
P.~M. Teillet, ``Image correction for radiometric effects in remote sensing,''
  \emph{International Journal of Remote Sensing}, 1986. [Online]. Available:
  \url{https://doi.org/10.1080/01431168608948958}
\BIBentrySTDinterwordspacing

\bibitem{ortho}
W.~Foundation, \emph{Orthorectification explanation}, November 2020 (accessed
  November 20, 2020), \emph{https://en.wikipedia.org/wiki/Orthophoto}.

\bibitem{Harsanyi1994hyperspectral}
J.~Harsanyi and C.-I. Chang, ``Hyperspectral image classification and
  dimensionality reduction: an orthogonal subspace projection approach,''
  \emph{IEEE Transactions on Geoscience and Remote Sensing}, 1994.

\bibitem{Moreira2013}
A.~Moreira, P.~Prats-Iraola, M.~Younis, G.~Krieger, I.~Hajnsek, and K.~P.
  Papathanassiou, ``A tutorial on synthetic aperture radar,'' \emph{IEEE
  Geoscience and Remote Sensing Magazine}, 2013.

\bibitem{groundrange}
E.~S. Agency, \emph{Radar Course 2}, June 2021 (accessed June 26, 2021).

\bibitem{Tomiyasu1978}
K.~Tomiyasu, ``Tutorial review of synthetic-aperture radar (sar) with
  applications to imaging of the ocean surface,'' \emph{Proceedings of the
  IEEE}, 1978.

\bibitem{Oliver2004}
C.~Oliver and S.~Quegan, \emph{Understanding Synthetic Aperture Radar Images},
  2004.

\bibitem{freesat}
G.~Geography, \emph{15 Free Satellite Imagery Data Sources}, August 2017
  (accessed January 1st, 2022),
  \emph{http://gisgeography.com/free-satellite-imagery-data-list}.

\bibitem{aviris_2022}
``Aviris,'' \url{https://aviris.jpl.nasa.gov/}, accessed: February 2022.

\bibitem{copernicus}
``Copernicus hub,''
  \url{https://sentinels.copernicus.eu/web/sentinel/user-guides/sentinel-2-msi}.

\bibitem{sentinel1_2022}
``Sentinel 1,''
  \url{https://sentinels.copernicus.eu/web/sentinel/missions/sentinel-1}, note
  = {Accessed: February 2022}.

\bibitem{sentinelmsi_2022}
``Sentinel msi,''
  \url{https://sentinels.copernicus.eu/web/sentinel/user-guides/sentinel-2-msi},
  accessed: February 2022.

\bibitem{digitalglobe}
M.~Technologies, \emph{Maxar DigitalGlobe Discover Portal}, accessed March 14,
  2022, \emph{https://discover.digitalglobe.com}.

\bibitem{digitalglobeuserguide}
DigitalGlobe, \emph{DigitalGlobe Core Imagery Products Guide}, November 2010
  (accessed March 14, 2022).

\bibitem{usgeologicalysurvey}
L.~mission, \emph{U.S. Geological Survey - Landsat Data Access}, March 2022
  (accessed March 14, 2022),
  \emph{https://www.usgs.gov/landsat-missions/landsat-data-access}.

\bibitem{landsat}
NASA, \emph{Landsat Science}, March 2022 (accessed March 14, 2022),
  \emph{https://landsat.gsfc.nasa.gov/}.

\bibitem{schmitt_2019}
M.~Schmitt, L.~H. Hughes, C.~Qiu, and X.~X. Zhu, ``Sen12ms -- a curated dataset
  of georeferenced multi-spectral sentinel-1/2 imagery for deep learning and
  data fusion,'' in \emph{ISPRS Annals of the Photogrammetry, Remote Sensing
  and Spatial Information Sciences}, vol. IV-2/W7, 2019, pp. 153--160.

\bibitem{modis}
NASA, \emph{Moderate Resolution Imagery Spectroradiometer (MODIS)}, June 2022
  (accessed June 22, 2022), \emph{https://modis.gsfc.nasa.gov/about/}.

\bibitem{goodfellow2014gans}
I.~Goodfellow, J.~Pouget-Abadie, M.~Mirza, B.~Xu, D.~Warde-Farley, S.~Ozair,
  A.~Courville, and Y.~Bengio, ``Generative adversarial nets,'' in
  \emph{Advances in Neural Information Processing Systems}, Z.~Ghahramani,
  M.~Welling, C.~Cortes, N.~Lawrence, and K.~Q. Weinberger, Eds.,
  vol.~27.\hskip 1em plus 0.5em minus 0.4em\relax Curran Associates, Inc.,
  2014.

\bibitem{Yu2017seqgan}
L.~Yu, W.~Zhang, J.~Wang, and Y.~Yu, ``Seqgan: Sequence generative adversarial
  nets with policy gradient,'' in \emph{Proceedings of the Thirty-First AAAI
  Conference on Artificial Intelligence}, 2017.

\bibitem{Kazeminia2020gans}
S.~Kazeminia, C.~Baur, A.~Kuijper, B.~{van Ginneken}, N.~Navab, S.~Albarqouni,
  and A.~Mukhopadhyay, ``Gans for medical image analysis,'' \emph{Artificial
  Intelligence in Medicine}, 2020.

\bibitem{karras_2017}
T.~Karras, T.~Aila, S.~Laine, and J.~Lehtinen, ``Progressive growing of gans
  for improved quality, stability, and variation,'' in \emph{6th International
  Conference on Learning Representations, {ICLR} 2018, Vancouver, BC, Canada,
  April 30 - May 3, 2018, Conference Track Proceedings}.\hskip 1em plus 0.5em
  minus 0.4em\relax OpenReview.net, 2018.

\bibitem{karras_2018}
T.~Karras, S.~Laine, and T.~Aila, ``A style-based generator architecture for
  generative adversarial networks,'' in \emph{{IEEE} Conference on Computer
  Vision and Pattern Recognition, {CVPR} 2019, Long Beach, CA, USA, June 16-20,
  2019}.\hskip 1em plus 0.5em minus 0.4em\relax Computer Vision Foundation /
  {IEEE}, 2019, pp. 4401--4410.

\bibitem{karras_2019}
T.~Karras, S.~Laine, M.~Aittala, J.~Hellsten, J.~Lehtinen, and T.~Aila,
  ``Analyzing and improving the image quality of stylegan,'' in \emph{2020
  {IEEE/CVF} Conference on Computer Vision and Pattern Recognition, {CVPR}
  2020, Seattle, WA, USA, June 13-19, 2020}.\hskip 1em plus 0.5em minus
  0.4em\relax Computer Vision Foundation / {IEEE}, 2020, pp. 8107--8116.

\bibitem{Mirza_2014_cgan}
\BIBentryALTinterwordspacing
M.~Mirza and S.~Osindero, ``Conditional generative adversarial nets,''
  \emph{CoRR}, vol. abs/1411.1784, 2014. [Online]. Available:
  \url{http://arxiv.org/abs/1411.1784}
\BIBentrySTDinterwordspacing

\bibitem{chen2020}
R.~Chen, W.~Huang, B.~Huang, F.~Sun, and B.~Fang, ``Reusing discriminators for
  encoding: Towards unsupervised image-to-image translation,'' in \emph{2020
  {IEEE/CVF} Conference on Computer Vision and Pattern Recognition, {CVPR}
  2020, Seattle, WA, USA, June 13-19, 2020}.\hskip 1em plus 0.5em minus
  0.4em\relax Computer Vision Foundation / {IEEE}, 2020, pp. 8165--8174.

\bibitem{Gulrajani2017}
I.~Gulrajani, F.~Ahmed, M.~Arjovsky, V.~Dumoulin, and A.~C. Courville,
  ``Improved training of wasserstein gans,'' in \emph{Advances in Neural
  Information Processing Systems 30: Annual Conference on Neural Information
  Processing Systems 2017, December 4-9, 2017, Long Beach, CA, {USA}},
  I.~Guyon, U.~von Luxburg, S.~Bengio, H.~M. Wallach, R.~Fergus, S.~V.~N.
  Vishwanathan, and R.~Garnett, Eds., 2017, pp. 5767--5777.

\bibitem{arjovsky2017_cwgan}
M.~Arjovsky, S.~Chintala, and L.~Bottou, ``Wasserstein generative adversarial
  networks,'' in \emph{Proceedings of the 34th International Conference on
  Machine Learning, {ICML} 2017, Sydney, NSW, Australia, 6-11 August 2017},
  ser. Proceedings of Machine Learning Research, D.~Precup and Y.~W. Teh, Eds.,
  vol.~70.\hskip 1em plus 0.5em minus 0.4em\relax {PMLR}, 2017, pp. 214--223.

\bibitem{Kingma2014}
\BIBentryALTinterwordspacing
D.~P. Kingma and M.~Welling, ``Auto-encoding variational bayes,'' in \emph{2nd
  International Conference on Learning Representations, {ICLR} 2014, Banff, AB,
  Canada, April 14-16, 2014, Conference Track Proceedings}, Y.~Bengio and
  Y.~LeCun, Eds., 2014. [Online]. Available:
  \url{http://arxiv.org/abs/1312.6114}
\BIBentrySTDinterwordspacing

\bibitem{Hinton1993_autoencoders}
G.~E. Hinton and R.~Zemel, ``Autoencoders, minimum description length and
  helmholtz free energy,'' in \emph{Advances in Neural Information Processing
  Systems (NIPS)}, 1993.

\bibitem{Guo2017}
J.~Guo, B.~Lei, C.~Ding, and Y.~Zhang, ``Synthetic aperture radar image
  synthesis by using generative adversarial nets,'' \emph{{IEEE} Geosci.
  Remote. Sens. Lett.}, vol.~14, no.~7, pp. 1111--1115, 2017.

\bibitem{zhan2018_ganforalignment}
Y.~Zhan, D.~Hu, Y.~Wang, and X.~Yu, ``Semisupervised hyperspectral image
  classification based on generative adversarial networks,'' \emph{IEEE
  Geoscience and Remote Sensing Letters}, vol.~15, no.~2, pp. 212--216, 2018.

\bibitem{abady2020}
\BIBentryALTinterwordspacing
L.~Abady, M.~Barni, A.~Garzelli, and B.~Tondi, ``{GAN generation of synthetic
  multispectral satellite images},'' in \emph{Image and Signal Processing for
  Remote Sensing XXVI}, L.~Bruzzone, F.~Bovolo, and E.~Santi, Eds., vol. 11533,
  International Society for Optics and Photonics.\hskip 1em plus 0.5em minus
  0.4em\relax SPIE, 2020, pp. 122 -- 133. [Online]. Available:
  \url{https://doi.org/10.1117/12.2575765}
\BIBentrySTDinterwordspacing

\bibitem{Ren2021deepfaking}
C.~X. Ren, A.~Ziemann, J.~Theiler, and J.~Moore, ``{Deepfaking it: experiments
  in generative, adversarial multispectral remote sensing},'' in
  \emph{Algorithms, Technologies, and Applications for Multispectral and
  Hyperspectral Imaging XXVII}, 2021.

\bibitem{zhao2021_cycleganrgb}
\BIBentryALTinterwordspacing
B.~Zhao, S.~Zhang, C.~Xu, Y.~Sun, and C.~Deng, ``Deep fake geography? when
  geospatial data encounter artificial intelligence,'' \emph{Cartography and
  Geographic Information Science}, vol.~48, no.~4, pp. 338--352, 2021.
  [Online]. Available: \url{https://doi.org/10.1080/15230406.2021.1910075}
\BIBentrySTDinterwordspacing

\bibitem{raytracing_2010}
S.~Auer, S.~Hinz, and R.~Bamler, ``Ray-tracing simulation techniques for
  understanding high-resolution sar images,'' \emph{IEEE Transactions on
  Geoscience and Remote Sensing}, vol.~48, no.~3, pp. 1445--1456, 2010.

\bibitem{radford_2016}
A.~Radford, L.~Metz, and S.~Chintala, ``Unsupervised representation learning
  with deep convolutional generative adversarial networks,'' in \emph{4th
  International Conference on Learning Representations, {ICLR} 2016, San Juan,
  Puerto Rico, May 2-4, 2016, Conference Track Proceedings}, Y.~Bengio and
  Y.~LeCun, Eds., 2016.

\bibitem{qgis}
``Qgis,'' \url{https://www.qgis.org/}, accessed: May 2022.

\bibitem{he2018_rgb}
\BIBentryALTinterwordspacing
W.~He and N.~Yokoya, ``Multi-temporal sentinel-1 and -2 data fusion for optical
  image simulation,'' \emph{{ISPRS} Int. J. Geo Inf.}, vol.~7, no.~10, p. 389,
  2018. [Online]. Available: \url{https://doi.org/10.3390/ijgi7100389}
\BIBentrySTDinterwordspacing

\bibitem{Merkle2018}
N.~Merkle, S.~Auer, R.~M{\"{u}}ller, and P.~Reinartz, ``Exploring the potential
  of conditional adversarial networks for optical and {SAR} image matching,''
  \emph{{IEEE} J. Sel. Top. Appl. Earth Obs. Remote. Sens.}, vol.~11, no.~6,
  pp. 1811--1820, 2018.

\bibitem{enomoto2018gan}
K.~Enomoto, K.~Sakurada, W.~Wang, N.~Kawaguchi, M.~Matsuoka, and R.~Nakamura,
  ``Image translation between sar and optical imagery with generative
  adversarial nets,'' in \emph{IGARSS 2018 - 2018 IEEE International Geoscience
  and Remote Sensing Symposium}, 2018, pp. 1752--1755.

\bibitem{liu2018}
L.~Liu and B.~Lei, ``Can sar images and optical images transfer with each
  other?'' in \emph{IGARSS 2018 - 2018 IEEE International Geoscience and Remote
  Sensing Symposium}, 2018, pp. 7019--7022.

\bibitem{reyes2019}
\BIBentryALTinterwordspacing
M.~Fuentes~Reyes, S.~Auer, N.~Merkle, C.~Henry, and M.~Schmitt,
  ``Sar-to-optical image translation based on conditional generative
  adversarial networks-optimization, opportunities and limits,'' \emph{Remote
  Sensing}, vol.~11, no.~17, 2019. [Online]. Available:
  \url{https://www.mdpi.com/2072-4292/11/17/2067}
\BIBentrySTDinterwordspacing

\bibitem{bermudez2019synthesis}
J.~D. Bermudez, P.~N. Happ, R.~Q. Feitosa, and D.~A. Oliveira, ``Synthesis of
  multispectral optical images from sar/optical multitemporal data using
  conditional generative adversarial networks,'' \emph{IEEE Geoscience and
  Remote Sensing Letters}, vol.~16, no.~8, pp. 1220--1224, 2019.

\bibitem{andrade2020}
H.~J.~A. Andrade and B.~J.~T. Fernandes, ``Synthesis of satellite-like urban
  images from historical maps using conditional gan,'' \emph{IEEE Geoscience
  and Remote Sensing Letters}, pp. 1--4, 2020.

\bibitem{yuan2020}
\BIBentryALTinterwordspacing
X.~Yuan, J.~Tian, and P.~Reinartz, ``Generating artificial near infrared
  spectral band from rgb image using conditional generative adversarial
  network,'' \emph{ISPRS Annals of the Photogrammetry, Remote Sensing and
  Spatial Information Sciences}, vol. V-3-2020, pp. 279--285, 2020. [Online].
  Available:
  \url{https://www.isprs-ann-photogramm-remote-sens-spatial-inf-sci.net/V-3-2020/279/2020/}
\BIBentrySTDinterwordspacing

\bibitem{vandal2020}
T.~Vandal, D.~McDuff, W.~Wang, K.~Duffy, A.~R. Michaelis, and R.~R. Nemani,
  ``Spectral synthesis for geostationary satellite-to-satellite translation,''
  \emph{{IEEE} Trans. Geosci. Remote. Sens.}, vol.~60, pp. 1--11, 2022.

\bibitem{schmitt2018colorizing}
M.~Schmitt, L.~Hughes, M.~K{\"o}rner, and X.~X. Zhu, ``Colorizing sentinel-1
  {SAR} images using a variational autoencoder conditioned on sentinel-2
  imagery,'' \emph{The International Archives of the Photogrammetry, Remote
  Sensing and Spatial Information Sciences (ISSN: 1682-1750)}, vol.~42, pp.
  1045--1051, 2018.

\bibitem{tasar2019}
O.~Tasar, S.~L. Happy, Y.~Tarabalka, and P.~Alliez, ``Colormapgan: Unsupervised
  domain adaptation for semantic segmentation using color mapping generative
  adversarial networks,'' \emph{{IEEE} Trans. Geosci. Remote. Sens.}, vol.~58,
  no.~10, pp. 7178--7193, 2020.

\bibitem{enomoto2017filmy}
K.~Enomoto, K.~Sakurada, W.~Wang, H.~Fukui, M.~Matsuoka, R.~Nakamura, and
  N.~Kawaguchi, ``Filmy cloud removal on satellite imagery with multispectral
  conditional generative adversarial nets,'' in \emph{Proceedings of the IEEE
  Conference on Computer Vision and Pattern Recognition Workshops}, 2017, pp.
  48--56.

\bibitem{singh2018}
P.~Singh and N.~Komodakis, ``Cloud-gan: Cloud removal for sentinel-2 imagery
  using a cyclic consistent generative adversarial networks,'' in \emph{IGARSS
  2018 - 2018 IEEE International Geoscience and Remote Sensing Symposium},
  2018, pp. 1772--1775.

\bibitem{grohnfeldt2018_conditional}
C.~Grohnfeldt, M.~Schmitt, and X.~Zhu, ``A conditional generative adversarial
  network to fuse sar and multispectral optical data for cloud removal from
  sentinel-2 images,'' in \emph{IGARSS 2018-2018 IEEE International Geoscience
  and Remote Sensing Symposium}.\hskip 1em plus 0.5em minus 0.4em\relax IEEE,
  2018, pp. 1726--1729.

\bibitem{zotov2019}
M.~Zotov and J.~Gamper, ``Conditional denoising of remote sensing imagery using
  cycle-consistent deep generative models,'' \emph{CoRR}, vol. abs/1910.14567,
  2019.

\bibitem{ebel2020}
P.~Ebel, M.~Schmitt, and X.~X. Zhu, ``Cloud removal in unpaired sentinel-2
  imagery using cycle-consistent gan and sar-optical data fusion,'' in
  \emph{IGARSS 2020 - 2020 IEEE International Geoscience and Remote Sensing
  Symposium}, 2020, pp. 2065--2068.

\bibitem{gao2020cloud}
J.~Gao, Q.~Yuan, J.~Li, H.~Zhang, and X.~Su, ``Cloud removal with fusion of
  high resolution optical and sar images using generative adversarial
  networks,'' \emph{Remote Sensing}, vol.~12, no.~1, p. 191, 2020.

\bibitem{sarukkai2020cloud}
V.~Sarukkai, A.~Jain, B.~Uzkent, and S.~Ermon, ``Cloud removal from satellite
  images using spatiotemporal generator networks,'' in \emph{Proceedings of the
  IEEE/CVF Winter Conference on Applications of Computer Vision}, 2020, pp.
  1796--1805.

\bibitem{wen2021}
\BIBentryALTinterwordspacing
X.~Wen, Z.~Pan, Y.~Hu, and J.~Liu, ``Generative adversarial learning in yuv
  color space for thin cloud removal on satellite imagery,'' \emph{Remote
  Sensing}, vol.~13, no.~6, 2021. [Online]. Available:
  \url{https://www.mdpi.com/2072-4292/13/6/1079}
\BIBentrySTDinterwordspacing

\bibitem{He2016resnet}
K.~He, X.~Zhang, S.~Ren, and J.~Sun, ``Deep residual learning for image
  recognition,'' in \emph{IEEE Conference on Computer Vision and Pattern
  Recognition (CVPR)}, 2016.

\bibitem{mao2017_lsgan}
X.~Mao, Q.~Li, H.~Xie, R.~Y.~K. Lau, Z.~Wang, and S.~P. Smolley, ``Least
  squares generative adversarial networks,'' in \emph{{IEEE} International
  Conference on Computer Vision, {ICCV} 2017, Venice, Italy, October 22-29,
  2017}.\hskip 1em plus 0.5em minus 0.4em\relax {IEEE} Computer Society, 2017,
  pp. 2813--2821.

\bibitem{baier2022synthesis}
G.~Baier, A.~Deschemps, M.~Schmitt, and N.~Yokoya, ``Synthesizing optical and
  sar imagery from land cover maps and auxiliary raster data,'' \emph{IEEE
  Transactions on Geoscience and Remote Sensing}, vol.~60, pp. 1--12, 2022.

\bibitem{spade_2019}
T.~Park, M.~Liu, T.~Wang, and J.~Zhu, ``Semantic image synthesis with
  spatially-adaptive normalization,'' in \emph{{IEEE} Conference on Computer
  Vision and Pattern Recognition, {CVPR} 2019, Long Beach, CA, USA, June 16-20,
  2019}.\hskip 1em plus 0.5em minus 0.4em\relax Computer Vision Foundation /
  {IEEE}, 2019, pp. 2337--2346.

\bibitem{geonrw_2020}
\BIBentryALTinterwordspacing
G.~Baier, A.~Deschemps, M.~Schmitt, and N.~Yokoya, ``Geonrw,'' 2020. [Online].
  Available: \url{https://dx.doi.org/10.21227/s5xq-b822}
\BIBentrySTDinterwordspacing

\bibitem{DFC_2020}
\BIBentryALTinterwordspacing
M.~S. L. H. P. G. N. Y.~R. Hansch, ``2020 ieee grss data fusion contest,''
  2019. [Online]. Available: \url{https://dx.doi.org/10.21227/rha7-m332}
\BIBentrySTDinterwordspacing

\bibitem{dem_2021}
P.~L. Guth, A.~V. Niekerk, C.~H. Grohmann, J.~Muller, L.~Hawker, I.~V.
  Florinsky, D.~Gesch, H.~I. Reuter, V.~Herrera{-}Cruz, S.~Riazanoff,
  C.~L{\'{o}}pez{-}V{\'{a}}zquez, C.~C. Carabajal, C.~Albinet, and P.~Strobl,
  ``Digital elevation models: Terminology and definitions,'' \emph{Remote.
  Sens.}, vol.~13, no.~18, p. 3581, 2021.

\bibitem{liu2017_unsupvaegan}
M.~Liu, T.~M. Breuel, and J.~Kautz, ``Unsupervised image-to-image translation
  networks,'' in \emph{Advances in Neural Information Processing Systems 30:
  Annual Conference on Neural Information Processing Systems 2017, December
  4-9, 2017,Long Beach, CA, {USA}}, I.~Guyon, U.~von Luxburg, S.~Bengio, H.~M.
  Wallach, R.~Fergus, S.~V.~N. Vishwanathan, and R.~Garnett, Eds., 2017, pp.
  700--708.

\bibitem{Larsen_2016}
A.~B.~L. Larsen, S.~K. S{\o}nderby, H.~Larochelle, and O.~Winther,
  ``Autoencoding beyond pixels using a learned similarity metric,'' in
  \emph{Proceedings of the 33nd International Conference on Machine Learning,
  {ICML} 2016, New York City, NY, USA, June 19-24, 2016}, ser. {JMLR} Workshop
  and Conference Proceedings, M.~Balcan and K.~Q. Weinberger, Eds.,
  vol.~48.\hskip 1em plus 0.5em minus 0.4em\relax JMLR.org, 2016, pp.
  1558--1566.

\bibitem{unit_2017}
M.~Liu, T.~M. Breuel, and J.~Kautz, ``Unsupervised image-to-image translation
  networks,'' in \emph{Advances in Neural Information Processing Systems 30:
  Annual Conference on Neural Information Processing Systems 2017, December
  4-9, 2017, Long Beach, CA, {USA}}, I.~Guyon, U.~von Luxburg, S.~Bengio, H.~M.
  Wallach, R.~Fergus, S.~V.~N. Vishwanathan, and R.~Garnett, Eds., 2017, pp.
  700--708.

\bibitem{deshpande2017_imgclr}
A.~Deshpande, J.~Lu, M.~Yeh, M.~J. Chong, and D.~A. Forsyth, ``Learning diverse
  image colorization,'' in \emph{2017 {IEEE} Conference on Computer Vision and
  Pattern Recognition, {CVPR} 2017, Honolulu, HI, USA, July 21-26, 2017}.\hskip
  1em plus 0.5em minus 0.4em\relax {IEEE} Computer Society, 2017, pp.
  2877--2885.

\bibitem{bishop94_mixturedensity}
C.~M. Bishop, ``Mixture density networks,'' Tech. Rep., 1994.

\bibitem{pauli_77}
H.~Pauli, ``Cie recommendations on uniform color spaces, color-difference
  equations, and metric color terms,'' \emph{Color Research \& Application},
  vol.~2, no.~1, pp. 5--6, 1977.

\bibitem{perlin2002_noise}
K.~Perlin, ``Improving noise,'' \emph{ACM Trans. Graph.}, vol.~21, no.~3, p.
  681–682, jul 2002.

\bibitem{wang2019}
X.~Wang, G.~Xu, Y.~Wang, D.~Lin, P.~Li, and X.~Lin, ``Thin and thick cloud
  removal on remote sensing image by conditional generative adversarial
  network,'' in \emph{IGARSS 2019 - 2019 IEEE International Geoscience and
  Remote Sensing Symposium}, 2019, pp. 1426--1429.

\bibitem{maliciousOverheadAI}
P.~Tucker, \emph{The Newest AI-Enabled Weapon: ‘Deep-Faking’ Photos of the
  Earth}, 31 March 2019 (accessed August 8, 2022),
  \emph{https://www.defenseone.com/technology/2019/03/next-phase-ai-deep-faking-whole-world-and-china-ahead/155944/}.

\bibitem{fawcett_2006}
T.~Fawcett, ``An introduction to {ROC} analysis,'' \emph{Pattern Recognit.
  Lett.}, vol.~27, no.~8, pp. 861--874, 2006.

\bibitem{taha_2015}
A.~A. Taha and A.~Hanbury, ``Metrics for evaluating 3d medical image
  segmentation: analysis, selection, and tool,'' \emph{{BMC} Medical Imaging},
  vol.~15, p.~29, 2015.

\bibitem{he_2013}
H.~He and Y.~Ma, \emph{Imbalanced Learning: Foundations, Algorithms, and
  Applications}, 1st~ed.\hskip 1em plus 0.5em minus 0.4em\relax Wiley-IEEE
  Press, 2013.

\bibitem{manning_2001}
C.~D. Manning and H.~Sch{\"{u}}tze, \emph{Foundations of statistical natural
  language processing}.\hskip 1em plus 0.5em minus 0.4em\relax {MIT} Press,
  2001.

\bibitem{yarlagadda2018satellite}
S.~Yarlagadda, D.~G\"uera, P.~Bestagini, F.~Zhu, S.~Tubaro, and E.~Delp,
  ``Satellite image forgery detection and localization using {GAN} and
  one-class classifier,'' in \emph{Electronic Imaging (EI)}, 2018.

\bibitem{bartusiak2019splicing}
E.~R. Bartusiak, S.~K. Yarlagadda, D.~G\"uera, P.~Bestagini, S.~Tubaro, F.~M.
  Zhu, and E.~J. Delp, ``Splicing detection and localization in satellite
  imagery using conditional {GAN}s,'' in \emph{IEEE Conference on Multimedia
  Information Processing and Retrieval (MIPR)}, 2019.

\bibitem{horvath2019anomaly}
J.~Horv\`ath, D.~G\"uera, S.~K. Yarlagadda, P.~Bestagini, F.~M. Zhu, S.~Tubaro,
  and E.~J. Delp, ``Anomaly-based manipulation detection in satellite images,''
  in \emph{IEEE Conference on Computer Vision and Pattern Recognition Workshop
  (CVPRW)}, 2019.

\bibitem{masmontserrat2020generative}
D.~Mas~Montserrat, J.~Horv\`ath, S.~K. Yarlagadda, F.~M. Zhu, and E.~J. Delp,
  ``Generative autoregressive ensembles for satellite imagery manipulation
  detection,'' in \emph{IEEE International Workshop on Information Forensics
  and Security (WIFS)}, 2020.

\bibitem{horvath2020manipulation}
J.~Horv\`ath, D.~Mas~Montserrat, H.~Hao, and E.~J. Delp, ``Manipulation
  detection in satellite images using deep belief networks,'' in \emph{IEEE
  Conference on Computer Vision and Pattern Recognition Workshop (CVPRW)},
  2020.

\bibitem{horvath2021_visiontransformer4detection}
J.~Horváth, S.~Baireddy, H.~Hao, D.~M. Montserrat, and E.~J. Delp,
  ``Manipulation detection in satellite images using vision transformer,'' in
  \emph{2021 IEEE/CVF Conference on Computer Vision and Pattern Recognition
  Workshops (CVPRW)}, 2021, pp. 1032--1041.

\bibitem{horvath2021_attentionunet}
J.~Horv{\'a}th, D.~M. Montserrat, E.~J. Delp, and J.~Horv{\'a}th, ``Nested
  attention u-net: A splicing detection method for satellite images,'' in
  \emph{Pattern Recognition. ICPR International Workshops and Challenges},
  A.~Del~Bimbo, R.~Cucchiara, S.~Sclaroff, G.~M. Farinella, T.~Mei, M.~Bertini,
  H.~J. Escalante, and R.~Vezzani, Eds.\hskip 1em plus 0.5em minus 0.4em\relax
  Cham: Springer International Publishing, 2021, pp. 516--529.

\bibitem{chen2021_geodefakehop}
\BIBentryALTinterwordspacing
H.~Chen, K.~Zhang, S.~Hu, S.~You, and C.~J. Kuo, ``Geo-defakehop:
  High-performance geographic fake image detection,'' \emph{CoRR}, vol.
  abs/2110.09795, 2021. [Online]. Available:
  \url{https://arxiv.org/abs/2110.09795}
\BIBentrySTDinterwordspacing

\bibitem{cannas2022amplitude}
E.~D. Cannas, N.~Bonettini, S.~Mandelli, P.~Bestagini, and S.~Tubaro,
  ``Amplitude {SAR} imagery splicing localization,'' \emph{IEEE Access}, 2022.

\bibitem{cannas2022panchromatic}
E.~D. Cannas, J.~Horv\'{a}th, S.~Baireddy, P.~Bestagini, E.~J. Delp, and
  S.~Tubaro, ``Panchromatic imagery copy-paste localization through data-driven
  sensor attribution,'' in \emph{IEEE International Conference on Acoustics,
  Speech and Signal Processing (ICASSP)}, 2022.

\bibitem{chen2016xgboost}
T.~Chen and C.~Guestrin, ``Xgboost: A scalable tree boosting system,'' in
  \emph{Proceedings of the 22nd ACM SIGKDD International Conference on
  Knowledge Discovery and Data Mining}, 2016.

\bibitem{Sundararajan2017axiomatic}
M.~Sundararajan, A.~Taly, and Q.~Yan, ``Axiomatic attribution for deep
  networks,'' in \emph{Proceedings of the International Conference on Machine
  Learning}, 2017.

\bibitem{Verdoliva2018deep}
L.~Verdoliva, ``Deep learning in multimedia forensics,'' in \emph{Proceedings
  of the 6th ACM Workshop on Information Hiding and Multimedia Security}, 2018.

\bibitem{Ruff2018deep}
L.~Ruff, R.~A. Vandermeulen, N.~G{\"o}rnitz, L.~Deecke, S.~A. Siddiqui,
  A.~Binder, E.~M{\"u}ller, and M.~Kloft, ``Deep one-class classification,'' in
  \emph{Proceedings of the 35th International Conference on Machine Learning},
  vol.~80, 2018, pp. 4393--4402.

\bibitem{Hinton2006reducing}
G.~E. Hinton and R.~R. Salakhutdinov, ``Reducing the dimensionality of data
  with neural networks,'' \emph{Science}, 2006.

\bibitem{Ackley1985learning}
D.~H. Ackley, G.~E. Hinton, and T.~J. Sejnowski, ``A learning algorithm for
  boltzmann machines,'' \emph{Cognitive Science}, 1985.

\bibitem{Oord2016pixel}
A.~V. Oord, N.~Kalchbrenner, and K.~Kavukcuoglu, ``Pixel recurrent neural
  networks,'' in \emph{Proceedings of The 33rd International Conference on
  Machine Learning}, 2016.

\bibitem{Oord2016conditional}
A.~van~den Oord, N.~Kalchbrenner, L.~Espeholt, k.~kavukcuoglu, O.~Vinyals, and
  A.~Graves, ``Conditional image generation with pixelcnn decoders,'' in
  \emph{Advances in Neural Information Processing Systems}, 2016.

\bibitem{Kolesnikov2021transformers}
A.~Kolesnikov, A.~Dosovitskiy, D.~Weissenborn, G.~Heigold, J.~Uszkoreit,
  L.~Beyer, M.~Minderer, M.~Dehghani, N.~Houlsby, S.~Gelly, T.~Unterthiner, and
  X.~Zhai, ``An image is worth 16x16 words: Transformers for image recognition
  at scale,'' 2021.

\bibitem{Kirchner2015forensic}
M.~Kirchner and T.~Gloe, ``{Forensic Camera Model Identification},'' in
  \emph{Handbook of Digital Forensics of Multimedia Data and Devices}.\hskip
  1em plus 0.5em minus 0.4em\relax Chichester, UK: John Wiley \& Sons, Ltd,
  2015, pp. 329--374.

\bibitem{cannas2021open}
E.~D. Cannas, S.~Baireddy, E.~R. Bartusiak, S.~K. Yarlagadda,
  D.~Mas~Montserrat, P.~Bestagini, S.~Tubaro, and E.~J. Delp, ``Open-set source
  attribution for panchromatic satellite imagery,'' in \emph{IEEE International
  Conference on Image Processing (ICIP)}, 2021.

\end{thebibliography}
